\documentclass[10pt,journal,letterpaper,compsoc]{IEEEtran}
\ifCLASSOPTIONcompsoc
\else
\fi
\ifCLASSINFOpdf
\else
\fi

\usepackage{epsfig}
\usepackage{amsmath}
\usepackage{amssymb}
\usepackage{latexsym}
\usepackage{graphicx}
\usepackage{mathrsfs}
\usepackage{dsfont}
\usepackage{epstopdf}
\usepackage{algorithm}
\usepackage{algorithmic}
\numberwithin{equation}{section}

\begin{document}
%
\title{Learning Discriminative Bayesian Networks from High-dimensional Continuous Neuroimaging Data}

%
%
%
%

\author{Luping~Zhou,~
        Lei~Wang,~
        Lingqiao~Liu,
	  Philip~Ogunbona,~
	  Dinggang~Shen
\IEEEcompsocitemizethanks{\IEEEcompsocthanksitem L. Zhou, L. Wang and P. Ogunbona are with the School of Computing and Information Technology, University of Wollongong, NSW 2500, Australia. E-mail: {lupingz, leiw, philipo}@uow.edu.au. L. Liu is with Shool of Computer Science, University of Adelaide, Australia. D. Shen is with Department of Radiology, University of North Carolina at Chapel Hill, USA.}}

\IEEEcompsoctitleabstractindextext{%

\begin{abstract}
Due to its causal semantics, Bayesian networks (BN) have been widely employed to discover the underlying data relationship in exploratory studies, such as brain research. Despite its success in modeling the probability distribution of variables, BN is naturally a generative model, which is not necessarily discriminative. This may cause the ignorance of subtle but critical network changes that are of investigation values across populations. In this paper, we propose to improve the discriminative power of BN models for continuous variables from two different perspectives. This brings two general discriminative learning frameworks for Gaussian Bayesian networks (GBN). In the first framework, we employ Fisher kernel to bridge the generative models of GBN and the discriminative classifiers of SVMs, and convert the GBN parameter learning to Fisher kernel learning via minimizing a generalization error bound of SVMs.  In the second framework, we employ the max-margin criterion and build it directly upon GBN models to explicitly optimize the classification performance of the GBNs. The advantages and disadvantages of the two frameworks are discussed and experimentally compared. Both of them demonstrate strong power in learning discriminative parameters of GBNs for neuroimaging based brain network analysis, as well as maintaining reasonable representation capacity. The contributions of this paper also include a new Directed Acyclic Graph (DAG) constraint with theoretical guarantee to ensure the graph validity of GBN.

\end{abstract}


}

\maketitle

\IEEEdisplaynotcompsoctitleabstractindextext

%
\IEEEpeerreviewmaketitle

\section{Introduction}\label{sec:introduction}
\IEEEPARstart{A}{s} an important probabilistic graphical model, Bayesian network (BN) has been used to model the probability distribution of a set of random variables for a wide spectrum of applications, e.g., diagnosis, troubleshooting, web mining, meteorology and bioinformatics. It combines graph representation with Bayesian analysis, providing an effective way to model and infer the conditional dependency of the variables. A BN has to be a directed acyclic graph (DAG). Two factors characterize a BN, i.e., the structure of the network (the presence / absence of edges in the graph) and the parameters of the probability distribution. Recent research of BN focuses on how to learn the structure and the parameters of BN directly from the data.   

The approaches of learning BN structures can be roughly categorized into the constraint-based,  the score-based, and the hybrid approaches. The constraint-based approaches use a serie of conditional independence testing to ensure the model structure is consistent with the conditional independency entailed by the observations.  Methods in this class include the IC algorithm~\cite{Verma-UAI-1991}, PC algorithm~\cite{Spirtes-SSCR-1991}, and more recent methods~\cite{Fast-thesis-2010,Scutari-arXiv-2014}. Score-based approaches define a scoring function over the space of candidate DAGs and optimize this function through certain search strategies. Methods in this class vary with scoring criteria, e.g., the posterior probability~\cite{Friedman-ML-2003, Koivisto-JMLR-2004, Geiger-arXiv-2013} and the minimum description length~\cite{Suzuki-UAI-1993}, or vary with search strategies, e.g., the heuristic search~\cite{Acid-JAIR-2003} and the Monte Carlo methods~\cite{Friedman-ML-2003}.  Hybrid approaches usually employ constraint-based methods to prune the search space of DAG structures and consequently restrict a subsequent score-based search~\cite{Tsamardinos-ML-2006,Jose-DMKD-2011}. Many existing BN learning methods, such as  LIMB-DAG~\cite{Schmidt-AAAI-2007}, MMHC~\cite{Tsamardinos-ML-2006}, TC and TC-bw~\cite{Pellet-JMLR-2008},  comprise of two stages: the identification of candidate parent sets in the first stage and the further pruning of them based on certain criteria in the second stage. Despite the mitigation of computational complexity, a drawback arises that if a parent node is missed in the first stage, it will never be recovered in the second stage~\cite{Huang-TPAMI-2012}. To address this issue,  one-stage learning process has been preferred in recent research work~\cite{Huang-TPAMI-2012,Xiang-NIPS-2013}. In these studies, based on Gaussian Bayesian network (GBN), the parent sets of all variables are learned together to optimize a LASSO-based score function in a single stage. The related optimization problems are solved either approximately~\cite{Huang-TPAMI-2012} or exactly~\cite{Xiang-NIPS-2013}. They have demonstrated improved reliability of BN edge identification over traditional two stage methods.

Although BN is naturally a generative method, it has also been used in classification tasks for diagnostic or predictive purposes. A straightforward usage is to train each class a BN and classify a new sample into the class with the highest likelihood value~\cite{Huang-TPAMI-2012}. Another kind of approaches trains ``Bayesian network classifiers" with discriminative objective functions~\cite{Pernkopf-JMLR-2010,Pernkopf-TPAMI-2012,Guo-UAI-2005}. In these approaches, usually a single BN is learned to optimize the discrimination performance. Either the structure or the parameters of the BN are adjusted to reflect the class difference for better classification. Therefore, the resulting BN 
does not model the distribution of any individual class. The ``Bayesian network classifiers"  in~\cite{Pernkopf-JMLR-2010,Pernkopf-TPAMI-2012,Guo-UAI-2005} are designed for discrete variables of multinomial distribution. They still inherit the two-stage learning process, i.e., have to predefine candidate parent sets as mentioned above.

Learning BN from the data faces new challenges in exploratory domains, such as brain research, where the mechanism of brain and mental diseases remain unclear and need to be explored. These domains usually cater for both interpretation and discrimination. ``Interpretation" requires interpretable models of the data and the findings explained by domain language rather than mathematical terms. This requirement comes from the demand of understanding the domain problems. ``Discrimination" requires the models to have sufficient discriminative power to distinguish groups of interest (such as identifying the diseased from the healthy), for the purpose of prediction.  To some extent, a high accuracy of the predictive model also provides a measure of the amount of information captured by that model. 

Being a generative method, BN represents the distribution of the data and is naturally amenable for interpretation. However, it is known that generative methods are not necessarily discriminative. They are prone to emphasizing major structures that are shared within each group, and neglecting the subtle but critical changes across groups. The latter, unfortunately, often happens, for example, in disease-induced brain changes across clinical groups. Consequently, generative methods are usually inferior in prediction compared with the discriminative methods that target only the boundary of classes (such as Support Vector Machines (SVMs)). On the other hand, discriminative methods often encounter the difficulty of interpretation, which is critical in exploratory research aimed at both the understanding and the prediction. Thus, this paper is motivated by the advantages that can be gained by learning BNs that are both representative and discriminative. 
Different from the Bayesian network classifiers in~\cite{Pernkopf-JMLR-2010,Pernkopf-TPAMI-2012,Guo-UAI-2005} that address discrete variables, we learn discriminative BNs for continuous variables, which is often needed in many domains including neuroimaging-based brain research. Moreover,  we learn for each class a BN with enhanced discrimination and maintain the BN representation of each individual class for interpretation\footnote{In this paper, we deal with the scenario that maintaining the BN representation of individual class is critical for the understanding of domain problems, such as the brain network models for the healthy and the diseased groups. However, it is not difficulty to see our discriminative learning frameworks could be slightly modified to learn only a single BN as the existing ``Bayesian network classifiers" for continuous variables. However, this deviates from our motivation and therefore is not unfolded in this paper.}.  
To achieve our goal, we propose two discriminative learning frameworks based on sparse Gaussian Bayesian network (SGBN). 

In the first framework (termed KL-SGBN), we employ Fisher kernel~\cite{Jaakkola-NIPS-1998} to link the generative models of SGBN to the discriminative classifiers of SVMs, and convert the SGBN parameter learning to Fisher kernel learning via maximizing a generalization bound of SVMs. The contributions of this framework include the following. i) By inducing Fisher kernel on SGBN models, we provide a way to obtain sample-specific SGBN-induced feature vectors that can be used by the discriminative classifiers such as SVMs. Through this, we bridge the generative models and the discriminative classifiers. ii) We propose a kernel learning approach to discriminatively learn the parameters of SGBNs by optimizing the performance of SVM. iii) As a by-product, the manipulation of Fisher kernel on SGBN provides a new way of variable selection for SGBNs. This framework has a computational advantage: through the mapping of Fisher kernel, the SGBN-induced feature vectors become linear functions of the SGBN parameters, which significantly simplifies the optimization problem in the learning process.

Unlike KL-SGBN where the discrimination is obtained by optimizing the classification performance of SVMs, in the second learning framework (termed MM-SGBN),  we propose to optimize a criterion directly built upon the classification performance of SGBNs. The motivation is that optimizing the performance of SVMs may not necessarily guarantee an equivalent improvement on SGBNs when SGBNs are the goal of applications. The contribution of this framework is a max-margin based method to jointly learn SGBNs, one for each class, for both representation and discrimination. 

In addition to the two discriminative SGBN learning frameworks, our contributions in this paper also include a new DAG constraint of SGBN based on topological ordering to ensure the validity of the graph. This new DAG constraint circumvents the awkward hard binarization of SGBN parameters in the process of optimization in~\cite{Huang-TPAMI-2012}, and simplifies the related optimization problems. This consequently makes it possible to optimize all the SGBN parameters together to avoid the influence of feature ordering encountered in the Block Coordinate Descent (BCD) optimization in~\cite{Huang-TPAMI-2012}. Moreover,  this new DAG constraint also circumvents the need for presetting candidate parent sets as in~\cite{Pernkopf-TPAMI-2012}.

Although the discriminative learning frameworks proposed in this paper are general methods, we focus on their applications in neuroimaging analysis for the early diagnosis of mental diseases.  A newly emerging field in the imaging-based neuroscience, called brain network analysis, attempts to model the brain as a complex network and study the interactions of brain regions via imaging-based features~\cite{Bullmore-NRN-2009}. Such research is important because brain network change is often found to be a response of the brain to damages. Due to its causal semantics, BN has been employed to model the ``effective connectivity" of the brain~\cite{Huang-TPAMI-2012,Smith-Neuroimage-2011,RuiLi-PLoSOne-2013}. The directionality of the connections may disclose the pathways of how one brain region affects another. The discoveries may lend further credence to the evidence of causal relationship changes found in many mental diseases, such as the Alzheimer's disease (AD)~\cite{Li-Neuroradio-2012,Huang-TPAMI-2012,Li-Neuroradiology-2011,RuiLi-PLoSOne-2013}, and uncover novel connectivity-based biomarkers for disease diagnosis. The proposed learning frameworks has been tested on multiple neuroimaging data sets. As demonstrated, our methods can significantly improve the discriminative power of the obtained SGBNs, as well as maintaining their representation capacity.

Early conference versions of this work were published in~\cite{Zhou-CVPR-2013,Zhou-MICCAI-2014}. In this paper, a significant extension has been made on the following aspects.  First, we analyze the problems of the DAG constraint used in~\cite{Zhou-CVPR-2013,Zhou-MICCAI-2014,Huang-TPAMI-2012}, and propose a new constraint with theoretically guaranteed DAG property to overcome those drawbacks. Second, we experimentally verify the new DAG constraint on benchmark Bayesian network data sets for network structure learning, and compare our method with another eight competing methods in the literature. Third, we update our two discriminative learning frameworks with the new DAG constraint and redo all the experiments in our early work~\cite{Zhou-CVPR-2013,Zhou-MICCAI-2014}. Fourth, we analyze the connections and differences between the two proposed discriminative learning frameworks, and conduct more comprehensive experiments to explore the characteristics of our frameworks with varied parameters, which has not been done in~\cite{Zhou-CVPR-2013,Zhou-MICCAI-2014}.

The rest of the paper is organized as follows. Section~\ref{sec:background} reviews SGBN and introduces the background of brain network analysis. Sections~\ref{sec:general-DL} elaborates two frameworks to learn discriminative and representative SGBNs from continuous data. Section~\ref{sec:DAG-constraint} revisits the problem of the existing DAG constraint of SGBN, and proposes a new one based on topological ordering. The proposed two learning frameworks with the new DAG constraint are experimentally tested in Section~\ref{sec:experiment}. This paper is concluded in Section~\ref{sec:conclusion}.  The notations of symbols frequently occurring in this paper are summarized in Table~\ref{Table:notations}.

\begin{table}[!ht]
\caption{\textbf{\textit{Notation}}}\label{Table:notations}
\vspace{-3mm}
\begin{center}
\renewcommand{\arraystretch}{1.2}
{\small
\begin{tabular}{l||l}
\hline
$x_i$ & a random variable\\
${\mathbf x}$ & a sample of $m$ variables: ${\mathbf x}=[x_1, x_2,\cdots,x_m]^{\top}$\\
${\mathbf X}$ & the data matrix of $n$ samples, ${\mathbf X}\in {\mathbb R}^{n\times m}$ \\
${\mathbf x}_{i, :}$ & the $i$-th row of ${\mathbf X}$,  representing a sample\\
${\mathbf x}_{:, i}$ & the $i$-th column of ${\mathbf X}$,  representing the realization\\
& ~~~~~~~~of the random variable $x_i$ on $n$ samples\\
${\boldsymbol \Theta}$ & the parameters of a Gaussian Bayesian Network\\
&~~~~~~~~${\boldsymbol \theta}=[{\boldsymbol \theta}_1,\cdots,{\boldsymbol \theta}_m]$, ${\boldsymbol \Theta}\in {\mathbb R}^{m\times m}$\\
${\mathbf {Pa}}_i$ & a vector containing the parents of  $x_i$\\
${\mathbf {PA}}_i$ & a matrix whose $j$-th column represents a\\
& ~~~~~~~~realization of ${\mathbf {Pa}}_i$ on the $j$-th sample.\\
${\mathbf G}$& an $m\times m$ matrix for BN: if there is a directed \textit{edge}\\
&~~~~~~~~from $x_i$  to $x_j$, ${\mathbf G}_{ij}=1$, otherwise ${\mathbf G}_{ij}=0$\\
${\mathbf P}$& an $m\times m$ matrix for BN: if there is a directed \textit{path}\\
&~~~~~~~~ from $x_i$ to $x_j$, ${\mathbf P}_{ij}=1$, otherwise ${\mathbf P}_{ij}=0$\\\hline
\end{tabular}
}
\end{center}
\end{table}
\section{Background}\label{sec:background}
To make this paper self-contained, we introduce the background for both the methodology and its application to brain network analysis. Please note that the methodology could be generalized to applications beyond the example given in this paper.

\subsection{Sparse Gaussian Bayesian Network (SGBN)}\label{subsec:orig-SGBN}
Because this paper is based on SGBN model, in the following, we review the fundamentals of SGBN in~\cite{Huang-TPAMI-2012}. All the symbols are defined in Table~\ref{Table:notations}. 

A Bayesian network (BN) ${\mathcal G}$ is a directed  acyclic graph (DAG), i.e. there is no closed path within the graph. It expresses the factorization property of a joint distribution $p({\mathbf x})=\underset{i=1,\cdots,m}\prod p(x_i | {\mathbf {Pa}}_i)$. The conditional probability $p(x_i | {\mathbf {Pa}}_i)$ is assumed to follow a Gaussian distribution in Gaussian Bayesian Network (GBN). Each node $x_i$ is regressed over its parent nodes ${\mathbf {Pa}}_i$: $x_i = {\boldsymbol \theta}_{i}^{\top}{\mathbf {Pa}}_i+\varepsilon_i$, where the vector ${\boldsymbol \theta}_i$ is the regression coefficients, and $\varepsilon_i \thicksim {\mathcal N}(0, \sigma_i^2)$. 
The structure of BN could be characterized by the $m\times m$ matrix ${\mathbf G}$ or ${\mathbf P}$ (defined in Table~\ref{Table:notations}), representing the \textit{edges} / \textit{paths} in the graph, respectively. 

Identifying parent sets is critical for BN learning. Traditional methods often consist of two stages: the candidate parent sets are initially identified in the first stage and further pruned by some criteria in the second stage. A drawback arises that when a true parent is missing in the first stage, it will never be recovered in the second stage. The work in~\cite{Huang-TPAMI-2012} proposed a different approach based on sparse GBN (SGBN), denoted as H-SGBN in this paper. In H-SGBN, each node $x_i$ is regressed over all the other nodes, and its parent set is implicitly selected by the regression coefficients ${\boldsymbol \theta}_{i}$ that are estimated through a constrained LASSO regression.
The following optimization is solved in~\cite{Huang-TPAMI-2012}:
\begin{align}\label{eqn:Huang_origin}
&\underset{\boldsymbol \Theta}\min\sum_{i=1}^m\|{\mathbf x}_{:, i} - {\mathbf {PA}}_i^{\top}{\boldsymbol \theta}_i\|_2^2+\lambda_1\|{\boldsymbol \theta}_i\|_1\\\nonumber
&s.t.~~{\boldsymbol \Theta}_{ji}\times{{\mathbf P}_{ij}}=0, \forall i, j=1,\cdots,m,~~i \neq j.
\end{align}
A challenge for BN learning is how to enforce the DAG property, i.e., avoiding directed cycles in the graph.  A sufficient and necessary condition for being a DAG is proposed in~\cite{Huang-TPAMI-2012},  which requires ${\boldsymbol \Theta}_{ji}\times{{\mathbf P}_{ij}}=0$ for all $i$ and $j$.  Note that ${\mathbf P}_{ij}$ is an implicit function of ${\boldsymbol \Theta}_{ji}$ (i.e., ${\mathbf P}=\mbox{expm}({\boldsymbol \Theta})$,  the matrix exponential function of ${\boldsymbol \Theta}$, as in~\cite{Huang-TPAMI-2012}).  Eqn.~(\ref{eqn:Huang_origin}) is difficult to solve. In~\cite{Huang-TPAMI-2012}, a block coordinate descent (BCD) method is employed to solve a LASSO-like problem efficiently. The whole ${\boldsymbol \Theta}$ is optimized column-wisely and iteratively. 
In each iteration $t$, only one column of ${\boldsymbol \Theta}$, say ${\boldsymbol \Theta}_{:, j}$, is optimized with ${\mathbf P}$ fixed as ${\mathbf P}^{(t-1)}$ in the last iteration. Then ${\boldsymbol \Theta}^{(t)}$, with the updated column ${\boldsymbol \Theta}_{:, j}$, is binarized to obtain ${\mathbf G}^{(t)}$, based on which, ${\mathbf P}^{(t)}$ is recalculated by a Breadth-first search with $x_i$ being the root node. The process is repeated until convergence. H-SGBN simultaneously obtains the structure and the parameters of an SGBN via learning ${\boldsymbol \Theta}$, e.g., there is no edge $i\rightarrow j$ if ${\boldsymbol \Theta}_{ij}$ is zero. It has been demonstrated to outperform the conventional two-stage methods in network edge recovery.

\subsection{Brain Network Analysis}\label{subsec:BrainNetwork}
Neuroimaging modalities and analysis techniques can provide more sensitive and consistent measurements than traditional cognitive assessment for the early diagnosis
of disease. Many mental disorders are found associated with subtle abnormalities distributed over the entire brain, rather than an individual brain region. The ``distributive" nature of mental disorders suggests the alteration of interactions between brain regions (neuronal systems) and thus the necessity of studying the brain as a complex network.  Brain networks are mathematically represented by graphical models, which can be constructed from neuroimaging data as follows. The brain images belonging to different subjects are first  spatially aligned to a common stereotaxic space by affine or deformable transformation, and then partitioned into regions of interest (ROI), i.e., clusters of imaging voxels, using either data-driven methods or predefined brain atlas. A brain network is then modeled by a graph with each node corresponding to a brain region and each edge corresponding to the connectivity between regions. Brain network analysis studies three kinds of brain connectivity. In this paper, we focus on the ``effective connectivity" that describes the influence one brain region exert upon another. Some early works in this field require a prior model of brain connectivity and most have only considered a small number ($ \le 10$) of brain regions using techniques such as structural equation
modeling ~\cite{Kim-HBM-2007} and dynamic causal modeling~\cite{Friston-Neuroimage-2003}.
More recently, models such as BN and Granger Causality have also been introduced into this field. It is suggested that BN may have advantages over those lag-based methods for brain network analysis by an experimental fMRI study~\cite{Smith-Neuroimage-2011}. Among BN-related methods, it is worth noting that the work in~\cite{Huang-TPAMI-2012} is completely data-driven, which recovers SGBN from more than 40 brain regions in fluorodeoxyglucose PET (FDG-PET) images. The method employs the strategy of sparsity constraint to handle relatively larger scale BN construction, and circumvents the traditional two-stage procedure for identifying parent sets in many sparse BN learning methods~\cite{Schmidt-AAAI-2007,Tsamardinos-ML-2006}. 
\section{Proposed Discriminative Learning of Generative SGBN}\label{sec:general-DL}
BN models are by definition generative models, focusing on how the data could be generated through an underlying process. In the context of neuroimage analysis, these models represent the effective brain connectivity of the given population. When used for classification, e.g., identifying AD patients from the healthy,  the SGBN models are trained for each class separately. A new sample ${\mathbf x}_i$ is then assigned to the class with the higher likelihood of SGBN. This may ignore some subtle but critical network differences that distinguish the classes. Therefore, we argue that the parameters of the generative model should be learned from the two classes jointly to keep the essential discrimination. 

Integrating generative and discriminative models is an important research topic in machine learning. In~\cite{Perina-TPAMI-2012}, the related approaches are roughly divided into three categories: blending, staging and iterative methods. In blending methods, both the discriminative and the generative terms are incorporated into the same objective function. In staging methods, the discriminative model is trained on features provided by the generative model. In iterative methods,  the generative and the discriminative models are trained iteratively to influence each other. In this paper, we propose two kinds of discriminative learning frameworks to achieve our goal. One is a staging method, called Fisher-kernel-induced discriminative learning (KL-SGBN). It extracts sample-based features from SGBN by Fisher kernel to optimize the classification performance of SVM. The other is a blending method, called max-margin-based discriminative learning (MM-SGBN). It directly optimizes the classification performance of SGBNs subject to maintaining SGBN's representation capacity. The two frameworks are elaborated in the following sections, respectively.


\subsection{Proposed Fisher-kernel-induced Discriminative Learning (KL-SGBN)}\label{sec:KL-SGBN}
\begin{centering}
\begin{figure}[ht]
\begin{center}
\includegraphics[width=0.45\textwidth]{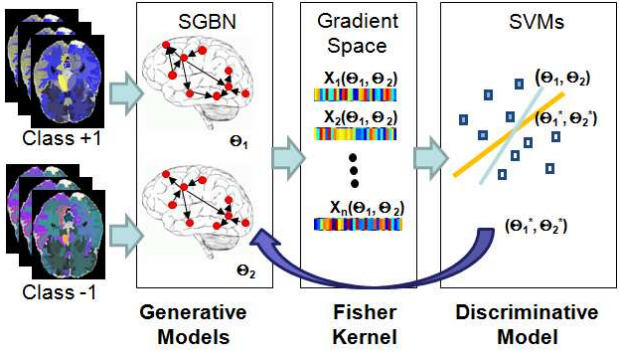}
\caption{\em Illustration of Fisher-kernel-induced Discriminative Learning.}\label{fig:KL-SGBN}
\end{center}
\vspace{-2mm}
\end{figure}
\end{centering}

We first introduce the Fisher-kernel-induced discriminative learning of SGBN, i.e., KL-SGBN. The algorithm is illustrated in Fig.~\ref{fig:KL-SGBN} and overviewed as follows. Given two classes in comparison, two SGBN models (with the parameters of ${\boldsymbol \Theta}_1$ and ${\boldsymbol \Theta}_2$) are learned, one for each individual class. The original samples are then mapped into the gradient space of the SGBN parameters ${\boldsymbol \Theta}_1$ and ${\boldsymbol \Theta}_2$ by Fisher kernel (Section~\ref{subsec:fisher-kernel}). Through this mapping, each sample is represented by a new feature vector (called Fisher vector~\cite{Jaakkola-NIPS-1998}) that is a function of ${\boldsymbol \Theta}=[{\boldsymbol \Theta}_1, {\boldsymbol \Theta}_2]$. These sample-specific feature vectors are then fed into an SVM classifier to minimize its generalization errors by adjusting ${\boldsymbol \Theta}$ (Section~\ref{subsec:DL-fisher-kernel}). The obtained optimal ${\boldsymbol \Theta}_1^{\star}$ and  ${\boldsymbol \Theta}_2^{\star}$ encode the discriminative information and therefore improve the original SGBNs. In this way, we convert the discriminative learning of SGBN parameters to the discriminative learning of Fisher kernels.  

\subsubsection{\textbf{Induction of Fisher vectors from SGBN}}\label{subsec:fisher-kernel}
Below we introduce how to use Fisher kernel on SGBNs to obtain feature vectors required for kernel learning. 
 
Fisher kernel~\cite{Jaakkola-NIPS-1998} provides a way to compare samples induced by a generative model. It maps a sample to a feature vector in the gradient space of the model parameters. The intuition is that similar objects induce similar log-likelihood gradients of the model parameters. Fisher kernel is computed as $K({\mathbf x}, {\mathbf x}') = {\mathbf g}_{\mathbf x}^{\top}{\mathbf U}^{-1}{\mathbf g}_{{\mathbf x}'}$, where the Fisher vector ${\mathbf g}_{\mathbf x} = \nabla_{\boldsymbol \theta} \log(p({\mathbf x}|{\boldsymbol \theta}))$ describes the changing direction of parameters to better fit the model. The Fisher information metric ${\mathbf U}$ weights the similarity measure, but is often set as an identity matrix in practice~\cite{Jaakkola-NIPS-1998}. 
 
Fisher kernel has recently witnessed successful applications in image categorization~\cite{Perronnin-cvpr-2007,Krapac-iccv-2011} for inducing feature vectors from Gaussian Mixture Model (GMM) of a visual vocabulary.  Despite its success, in the applications above, Fisher kernel is mainly used as a feature extractor\footnote{An exception~\cite{Sydorov-CVPR-2014} is discussed in ``Generalization" in Section~\ref{subsec:discussion}, which is published after our work~\cite{Zhou-CVPR-2013}.}. It has not been applied to learning the parameters of probability distributions before the early work of this paper in~\cite{Zhou-CVPR-2013}. The advantage of learning discriminative Fisher kernel has also been confirmed by a recent study that maximizes the class separability~\cite{Maaten-ICML-2011} of samples based on Fisher kernel, which is developed with different context and different criteria from ours. 

Following~\cite{Huang-TPAMI-2012},  we only consider ${\boldsymbol \Theta}$ as parameters and predefine $\sigma$.  Let ${\mathcal L}({\mathbf x} | {\boldsymbol \Theta})=\log(p({\mathbf x}| {\boldsymbol \Theta}))$  denote the log-likelihood. Our Fisher vector for each sample ${\mathbf x}$ is 
\[\Phi_{\boldsymbol \Theta}({\mathbf x}) = [\nabla_{{\boldsymbol \Theta}_1} {\mathcal L}({\mathbf x} | {\boldsymbol \Theta}_1)^{\top},~ \nabla_{{\boldsymbol \Theta}_2}  {\mathcal L}({\mathbf x} | {\boldsymbol \Theta}_2)^{\top}]^{\top},\] where ${\boldsymbol \Theta}_1$ and ${\boldsymbol \Theta}_2$ are the parameters of the SGBNs for the two classes ($y=1,2)$, respectively. Recall that, using a BN, the probability $p({\mathbf x}| {\boldsymbol \Theta})$ can be factorized as
$p({\mathbf x}| {\boldsymbol \Theta}) = \underset{i=1,\cdots,m}\prod p(x_i | {\mathbf {Pa}}_i, {\boldsymbol \theta}_i)$. Therefore, for GBN it can be shown that
\begin{align}\label{eqn:log-likelihood}
{\mathcal L}({\mathbf x}| {\boldsymbol \Theta})  &=\sum_{i=1}^m \log p(x_i | {\mathbf {Pa}}_i, {\boldsymbol \theta}_i)\\\nonumber
&= \sum_{i=1}^m \frac{-(x_i - {\mathbf {Pa}}_i^{\top}{\boldsymbol \theta}_i)^2}{2\sigma_i^2} - \log(2\pi\sqrt{\sigma_i}).
\end{align}
Taking partial derivative over  ${\boldsymbol \theta}_i$, we have
\begin{align}\label{eqn:fisher-kernel}
\frac{\partial {\mathcal L}({\mathbf x}| {\boldsymbol \Theta})}{\partial {\boldsymbol \theta}_i} &= -\frac{{\mathbf {Pa}}_i{\mathbf {Pa}}_i^{\top}}{\sigma_i^2}{\boldsymbol \theta}_i -\frac{x_i {\mathbf {Pa}}_i}{\sigma_i^2}\\\nonumber
&\triangleq{\mathbf S}(x_i){\boldsymbol \theta}_i+{\mathbf s}_{0}(x_i),
\end{align}
where ${\mathbf S}(x_i)$ is a squared matrix and ${\mathbf s}_{0}(x_i)$ is a vector. As shown, both ${\mathbf S}(x_i)$ and ${\mathbf s}_{0}(x_i)$ are constant with respect to ${\boldsymbol \Theta}$. Therefore, the Fisher vector $\Phi_{\boldsymbol \Theta}({\mathbf x})$ is a \textit{linear} function of ${\boldsymbol \Theta}$. This simple form of $\Phi_{\boldsymbol \Theta}({\mathbf x})$ significantly facilitates our further kernel learning.

\subsubsection{\textbf{Discriminative Fisher kernel learning via SVM}}\label{subsec:DL-fisher-kernel}
As each Fisher vector is a function of the SGBN parameters, discriminatively learning these parameters  can thus be converted to learning discriminative Fisher kernels.  We require that the learned SGBN models possess the following properties. Firstly, the Fisher vectors induced by the learned SGBN model should be well separated between classes. Secondly, the learned SGBN models should maintain reasonable capacity of representation. Thirdly, the learned SGBN models should not violate DAG.

We use the following strategies to achieve our goal. Firstly, to obtain a discriminative Fisher kernel, we jointly learn the parameters of SGBN and the separating hyper-plane of SVMs with Fisher kernel. Radius-margin bound, the upper bound of the Leave-One-Out error, is minimized to keep good generalization of the SVMs. Secondly, to maintain reasonable representation, we explicitly control the fitting (regression) errors of the learned model during optimization. Recall that GBN learns the network by minimizing the regression errors of each node over its parent nodes. Thirdly, we enforce the DAG constraint to ensure the validity of the graph. Our method is developed as follows.

In order to use radius-margin bound, ${\mathcal L}_2$-SVM with soft margin is be employed~\cite{Chapelle-ML-2002}\footnote{Radius-margin bound is rooted in hard-margin SVM.  ${\mathcal L}_2$-SVM with soft-margin can be rewritten as SVM with hard margin.}, which optimizes
\begin{align}\label{eqn:L2-svm}
&\underset{{\mathbf w},  {\boldsymbol \xi}}\min~~\frac{1}{2}\|{\mathbf w}\|_2^2 + C{\boldsymbol \xi}^{\top}{\boldsymbol \xi}\\\nonumber
s.t.~~&y_i({\mathbf w}^{\top}\Phi({\mathbf x}_i)+b) \geq 1-\xi_i, ~~  \xi_i\geq 0,~~\forall i.
\end{align}
Following the convention in SVMs, ${\mathbf w}$ is the normal of the separating plane, $b$ the bias term, ${\boldsymbol \xi}$ the slack variables and $C$ the regularization parameter.  Here $y_i$ is the class label of the $i$-th sample. ${\mathcal L}_2$-SVM can be rewritten as SVM with hard margin by slightly modifying the kernel ${\mathbf K} := {\mathbf K} + {\mathbf I}/C$, where ${\mathbf I}$ is identity matrix. For convenience, in the following, we redefine ${\mathbf w} := [{\mathbf w}^{\top}~\sqrt{C}{\boldsymbol \xi}^{\top}]^{\top}$ and $\Phi({\mathbf x}_i) := [\Phi^{\top}({\mathbf x}_i)~~{\mathbf e}^{\top}_i y_i/\sqrt{C}]^{\top}$. The vector ${\mathbf e}_i$ has the value of $1$ at the $i$-th element, and $0$ elsewhere.


Incorporating radius information leads to solving  
\begin{align}\label{eqn:R2-svm}
&\underset{{\mathbf w}}\min~~\frac{1}{2}R^2\|{\mathbf w}\|_2^2\\\nonumber
s.t.~~&y_i({\mathbf w}^{\top}\Phi({\mathbf x}_i)+b) \geq 1, ~~\forall i,
\end{align}
where $R^2$ denotes the radius of Minimal Enclosing Ball (MEB). It has been observed that when the sample size is small, the estimation of $R^2$ may become noisy and unstable~\cite{Wang-TPAMI-2008}. Therefore, it has been proposed to use the trace of the total scatter matrix instead for such cases~\cite{Wang-TPAMI-2008,Liu-TSMCB-2012}. We finally solve the following optimization problem:
\begin{align}\label{eqn:fisher_original}
&\underset{{\boldsymbol \theta}, {\mathbf w}}\min~~\frac{1}{2}{\mathrm {tr}}({\mathbf S}_T)\|{\mathbf w}\|_2^2 \\\nonumber
s.t.~~&y_i({\mathbf w}^{\top}\Phi_{\boldsymbol \Theta}({\mathbf x}_i)+ b) \geq 1, ~~\forall i\\\nonumber
&h({\mathbf X}_1, {\boldsymbol \Theta_1})\leq T_1,~~h({\mathbf X}_2, {\boldsymbol \Theta_2})\leq T_2,\\\nonumber
& {\boldsymbol \Theta}_1 \in DAG, ~~{\boldsymbol \Theta}_2 \in DAG.
\end{align}
Here ${\mathrm {tr}}({\mathbf S}_T)$ is the trace of the total scatter matrix ${\mathbf S}_T$, where ${\mathbf S}_T=\sum_{i=1}^n(\Phi({\mathbf x}_i)-{\mathbf m})(\Phi({\mathbf x}_i)-{\mathbf m})^{\top}$, and ${\mathbf m}$ is the mean of total $n$ samples in the kernel-induced space. It can be shown that ${\mathrm {tr}}({\mathbf S}_T)={\mathrm {tr}}({\mathbf K}) - {\mathbf 1}^{\top}{\mathbf K}{\mathbf 1}/n$, where ${\mathbf 1}$ denotes a vector whose elements are all $1$, and ${\mathbf K}$ the kernel matrix. Fisher vector $\Phi_{\boldsymbol \Theta}({\mathbf x}_i)$ is obtained as in Section~\ref{subsec:fisher-kernel}. The function $h(\cdot)$ measures the squared fitting errors of the corresponding SGBNs for the data ${\mathbf X}_1$ and ${\mathbf X}_2$ from the two classes. It is defined as
\begin{align}\label{eqn:fitting-constraint}
h({\mathbf X}, {\mathbf \Theta}) = \sum_{i=1}^m\|{\mathbf x}_{:, i} - {\mathbf {PA}}_i^{\top}{\boldsymbol \theta}_i\|^2_2.
\end{align}
The two user-defined parameters $T_1$ and $T_2$ explicitly control the degree of fitting during the learning. Adding these constraints also avoids the scaling problem of ${\boldsymbol \Theta}$.

The DAG constraint in H-SGBN could be employed to enforce the validity of the graph. However, here we adopt a new DAG constraint proposed in Section~\ref{sec:DAG-constraint} due to its advantages over that of H-SGBN. The new DAG constraint employs a set of topological ordering variables $({\mathbf o}, {\boldsymbol \Upsilon})$ to guarantee DAG. It is a bilinear function of the ordering variables $({\mathbf o}, {\boldsymbol \Upsilon})$ and the SGBN parameters ${\boldsymbol \Theta}$.  An elaboration is given in Section~\ref{sec:DAG-constraint}. At the moment, let us temporarily skip the details of this DAG constraint and concentrate on the discriminative learning. 

One possible approach for solving Eqn.~(\ref{eqn:fisher_original}) is to alternately optimize the separating hyperplane ${\mathbf w}$ and the parameter ${\boldsymbol \Theta}$.  That is,
\begin{align}\label{eqn:fisher-prime}
&\underset{{\boldsymbol \Theta}, {\mathbf o}, {\boldsymbol \Upsilon}}\min~~J({\boldsymbol \Theta})\\\nonumber
s.t.&~~h({\mathbf X}_1, {\boldsymbol \Theta_1})\leq T_1,~~h({\mathbf X}_2, {\boldsymbol \Theta_2})\leq T_2,\\\nonumber
&~{\boldsymbol \Theta}_1 \in DAG({\mathbf o}_1, {\boldsymbol \Upsilon}_1),~~{\boldsymbol \Theta}_2 \in DAG({\mathbf o}_2, {\boldsymbol \Upsilon}_2).
\end{align}
where
\begin{align}\label{eqn:svm_original}
&J({\boldsymbol \Theta})=\underset{{\mathbf w}}\min~~\frac{1}{2}{\mathrm {tr}}({\mathbf S}_T)\|{\mathbf w}\|_2^2 \\\nonumber
s.t.~~&y_i({\mathbf w}^{\top}\Phi_{\boldsymbol \Theta}({\mathbf x}_i)+b) \geq 1, ~~\forall i.
\end{align}
Note that for a given ${\boldsymbol \Theta}$, the term ${\mathrm {tr}}({\mathbf S}_T)$ is constant in Eqn.~(\ref{eqn:svm_original}). Due to the strong duality in SVM optimization, we solve the term $\|{\mathbf w}\|_2^2$ by 
\begin{align}\label{eqn:svm-dual}
J_0({\boldsymbol \Theta}) &=\underset{\boldsymbol \alpha}\max \sum_{i=1}^n \alpha_i - \frac{1}{2}\sum_{i=1}^n\sum_{j=1}^n y_i y_j \alpha_i \alpha_j K_{\boldsymbol \Theta}({\mathbf x}_i, {\mathbf x}_j)\\\nonumber
s.t. &\sum_{i=1}^n \alpha_i y_i = 0,~~\alpha_i \geq 0 ~~\forall i,
\end{align}
where $\alpha_i$ is the Lagrangian multiplier and $K_{\boldsymbol \Theta}({\mathbf x}_i, {\mathbf x}_j)=\langle\Phi_{\boldsymbol \Theta}({\mathbf x}_i), \Phi_{\boldsymbol \Theta}({\mathbf x}_j) \rangle$. 

As mentioned above, the DAG constraint is a bilinear function of $({\mathbf o}, {\boldsymbol \Upsilon})$ and ${\boldsymbol \Theta}$. Many quadratic programming packages could be used to solve Eqn.~(\ref{eqn:fisher-prime}). We use fmincon-SQP (sequential quadratic programming) in Matlab. Gradient information is required by many optimization algorithms (including fmincon-SQP) to speed up the line search. It is not difficult to find that the gradient of $K_{\boldsymbol \Theta}({\mathbf x}_i, {\mathbf x}_j)$  is just a linear function of ${\boldsymbol \Theta}$, making the evaluation of gradient $\nabla_{\boldsymbol \Theta} J$ easy. Our learning process is summarized in Algorithm~\ref{alg:KL-SGBN}.

\begin{algorithm}[tb]
   \caption{~~KL-SGBN: Discriminative Learning}
   \label{alg:KL-SGBN}
\begin{algorithmic}
   \STATE {\bfseries Input:} data ${\mathbf X}_1, {\mathbf X}_2 \in {\mathbb R}^{n\times m}$, label ${\mathbf y} \in {\mathbb R}^{n\times 1}$
\STATE~~~~~~~~~Denote ${\boldsymbol \Theta}=[{\boldsymbol \Theta}_1, {\boldsymbol \Theta}_2]$
\vspace{3mm}
\STATE  Initialize ${\boldsymbol \Theta}^{(0)}, {\mathbf o}^{(0)}, {\boldsymbol \Upsilon}^{(0)}$ by Algorithm~\ref{alg:OR-SGBN} for each class.
  \STATE Let ${\boldsymbol \Theta}^{(t-1)} = {\boldsymbol \Theta}^{(0)}$, ${\mathbf o}^{(t-1)} = {\mathbf o}^{(0)}$, ${\boldsymbol \Upsilon}^{(t-1)} = {\boldsymbol \Upsilon}^{(0)}$
 \REPEAT
  \STATE 1. Compute $\Phi_{\boldsymbol \Theta}^{(t-1)}$ and ${\mathbf K}_{\boldsymbol \Theta}^{(t-1)}$ by Eqn.~(\ref{eqn:fisher-kernel})
  \STATE 2. Compute ${\mathrm {tr}}({\mathbf S}_T)^{(t-1)}={\mathrm {tr}}({\mathbf K}_{\boldsymbol \Theta}^{(t-1)}) - {\mathbf 1}^{\top}{\mathbf K}_{\boldsymbol \Theta}^{(t-1)}{\mathbf 1}/n$
  \STATE 3. Solve $J_0({\boldsymbol \Theta}^{(t-1)})$ and ${\boldsymbol \alpha}^{\star}$ by Eqn.~(\ref{eqn:svm-dual})
  \STATE 4. $J({\boldsymbol \Theta}^{(t-1)})=J_0({\boldsymbol \Theta}^{(t-1)}) \times {\mathrm {tr}}({\mathbf S}_T)^{(t-1)}$
  \STATE 6. Minimize Eqn.~(\ref{eqn:fisher-prime}) with ${\boldsymbol \alpha}^{\star}$ and obtain ${\boldsymbol \Theta}^{(t)}$:
\STATE~~~~\textit{6.1} Let ${\mathbf o}={\mathbf o}^{(t-1)}, {\boldsymbol \Upsilon}={\boldsymbol \Upsilon}^{(t-1)}$,  solve ${\boldsymbol \Theta}^{(t)}$ by Eqn.~(\ref{eqn:fisher-prime});
\STATE~~~~~\textit{6.2} Let ${\boldsymbol \Theta}={\boldsymbol \Theta}^{(t)}$, solve ${\mathbf o}^{(t)}, {\boldsymbol \Upsilon}^{(t)}$ by 
Eqn.~(\ref{Eqn:single-class}).
  \STATE 7. Let ${\boldsymbol \Theta}^{(t-1)}={\boldsymbol \Theta}^{(t)}$, ${\mathbf o}^{(t-1)}={\mathbf o}^{(t)}$, ${\boldsymbol \Upsilon}^{(t-1)}={\boldsymbol \Upsilon}^{(t)}$
  \UNTIL{convergence/max number of iterations}
\vspace{3mm}
 \STATE {\bfseries Output:} ~${\boldsymbol \Theta}^{\star}={\boldsymbol \Theta}^{(t)}$
\end{algorithmic}
\end{algorithm}

\subsection{Proposed Max-margin-based Discriminative Learning (MM-SGBN)}\label{sec:MM-SGBN}
KL-SGBN introduces group discrimination into SGBNs by optimizing the performance of SVM classifiers with SGBN-induced features. Although this leads to a relatively simple optimization problem, optimizing the performance of SVMs does not necessarily imply optimizing the discrimination of SGBNs. We believe that, the discrimination of SGBNs can be further improved if we \textit{directly} optimize their (instead of SVMs') classification performance. Therefore we propose a new learning framework based on max-margin formulation directly built on SGBNs.  We call this method MM-SGBN.

%

For binary classification, maximizing the minimum margin between two classes can be obtained by maximizing the minimum conditional likelihood ratio (MCLR)~\cite{Guo-UAI-2005}:
\begin{align}
&\mbox{MCLR}({\boldsymbol \Theta}) = \min_{i=1}^n \frac{P(y_i|{\mathbf x}_i, {\boldsymbol \Theta}_{y_i})}{P(\bar{y}_i|{\mathbf x}_i, {\boldsymbol \Theta}_{\bar{y}_i})}, \nonumber
\end{align}
Without loss of generality, $y_i$ and $\bar{y}_i \in \{-1, 1\}$, representing the true and false labels for the $i$-th sample, respectively. The parameter ${\boldsymbol \Theta}_{y_i}={\boldsymbol \Theta}_1$ if $y_i=1$, or ${\boldsymbol \Theta}_{y_i}={\boldsymbol \Theta}_2$ if $y_i=-1$.  We can see that MCLR identifies the most confusing sample whose probability of the true class assignment is close to or even less than that of the false class assignment. Hence, maximizing MCLR targets the maximal separation of the most confusing samples in the two classes. It is not difficult to see that MCLR can naturally handle multi-class case when replacing the denominator by the maximal probability induced by all false class assignments. Let ${\boldsymbol \Theta}=[{\boldsymbol \Theta}_1, {\boldsymbol \Theta}_2]$. Taking log-likelihood of MCLR, we have
\begin{align}\label{Eqn:log-MCLR}
&\log~\mbox{MCLR}({\boldsymbol \Theta}) \nonumber \\
&= \min_{i=1}^n \left(\log~p({\mathbf x}_i|y_i, {\boldsymbol \Theta}_{y_i}) - \log~p({\mathbf x}_i| \bar{y}_i, {\boldsymbol \Theta}_{\bar{y}_i}) \right) + const,
\end{align}
where the prior probabilities of $P(y_i)$ and $P(\bar{y}_i)$ that are irrelevant to ${\boldsymbol \Theta}$ are absorbed into the constant term. Eqn.~(\ref{Eqn:log-MCLR}) can be shown to be a quadratic function of ${\boldsymbol \Theta}$ in the case of SGBN. In order to maximize MCLR, we require the difference of log-likelihood function in Eqn.~(\ref{Eqn:log-MCLR}) be larger than a margin for all samples, $r$, and maximize the margin $r$. To deal with hard separations, we employ a soft margin formulation as follows.
\begin{align}\label{Eqn:max-margin-opt}
 &\underset{{\boldsymbol \Theta}_1,{\boldsymbol \Theta}_2, \xi_i, r, {\mathbf o}, {\boldsymbol \Upsilon}}\min~~\lambda\sum_{i=1}^n \xi_i - r\\
&s.t.~~y_i\left({\mathcal L}({\boldsymbol \Theta}_1, {\mathbf x}_i) - {\mathcal L}({\boldsymbol \Theta}_2, {\mathbf x}_i)\right) \geq r-\xi_i, ~~\forall i \nonumber \tag{\ref{Eqn:max-margin-opt}a}\\
&~~~~~~\xi_i \geq 0, ~~r \geq 0, \nonumber \tag{\ref{Eqn:max-margin-opt}b}\\
&~~~~~~h({\mathbf X}_1, {\boldsymbol \Theta}_1) \leq T_1, ~~h({\mathbf X}_2, {\boldsymbol \Theta}_2) \leq T_2 \nonumber \tag{\ref{Eqn:max-margin-opt}c}\\
&~~~~~~{\boldsymbol \Theta}_1 \in DAG({\mathbf o}_1, {\boldsymbol \Upsilon}_1), ~~ {\boldsymbol \Theta}_2 \in DAG({\mathbf o}_2, {\boldsymbol \Upsilon}_2) \nonumber \tag{\ref{Eqn:max-margin-opt}d}
\end{align}

The constraints in (\ref{Eqn:max-margin-opt}a) enforce the likelihood of ${\mathbf x}_i$ to its true class larger than
that to its false class by a margin $r$. The variables $\xi_i$ are slack variables indicating the intrusion of the margin. The function ${\mathcal L}(\cdot)$ denotes the log-likelihood, defined in Eqn.~(\ref{eqn:log-likelihood}). We require ${\mathcal L}({\boldsymbol \Theta}_1, {\mathbf x}_i)$ larger than ${\mathcal L}({\boldsymbol \Theta}_2, {\mathbf x}_i)$ when $y_i=1$, and ${\mathcal L}({\boldsymbol \Theta}_2, {\mathbf x}_i)$ larger than ${\mathcal L}({\boldsymbol \Theta}_1, {\mathbf x}_i)$ when $y_i=-1$.

The constraints in (\ref{Eqn:max-margin-opt}c) control the fitting errors, same to that used in KL-SGBN, and the function $h(\cdot)$ is defined in Eqn.~(\ref{eqn:fitting-constraint}).

The constraints in (\ref{Eqn:max-margin-opt}d) are the DAG constraint proposed in Section~\ref{sec:DAG-constraint}, Eqn.~(\ref{eqn:DAG_constraint}). To enforce the validity of DAG on both graphs, we introduce a set of order variables ${\mathbf o} = \{o_1, o_2, \cdots, o_m\}$ and ${\boldsymbol \Upsilon}$  for each class separately, and employ the constraints stated in Eqn.~(\ref{eqn:DAG_constraint}). Please refer to Eqn.~(\ref{eqn:DAG_constraint}) for details.

The optimization in Eqn.~(\ref{Eqn:max-margin-opt}) can be solved iteratively by optimizing $({\boldsymbol \Theta}, \xi_i, r)$ and $({\mathbf o}, {\boldsymbol \Upsilon})$ alternately, as summarized in Algorithm~\ref{alg:MM-SGBN}. In Step 1, we solve a linear objective function with $n$ non-convex and two convex quadratic constraints by fmincon-SQP (sequential quadratic programming) in Matlab. In Step 2, we solve the linear programming by the package of CVX\footnote{http://cvxr.com/cvx/}.

\begin{algorithm}[tb]
   \caption{~~MM-SGBN: Discriminative Learning}
   \label{alg:MM-SGBN}
\begin{algorithmic}
 \STATE {\bfseries Input:} data ${\mathbf X}_1, {\mathbf X}_2 \in {\mathbb R}^{n\times m}$, label ${\mathbf y} \in {\mathbb R}^{n\times 1}$ 
\STATE~~~~~~~~~Denote ${\boldsymbol \Theta}=[{\boldsymbol \Theta}_1, {\boldsymbol \Theta}_2]$
\vspace{3mm}
\STATE  Initialize ${\boldsymbol \Theta}^{(0)}, {\mathbf o}^{(0)}, {\boldsymbol \Upsilon}^{(0)}$ by Algorithm~\ref{alg:OR-SGBN} for each class.
 \STATE Fix ${\boldsymbol \Theta}={\boldsymbol \Theta}^{(0)}$ and estimate $r^{(0)}$ and $\xi_i^{(0)}$ by Eqn.~(\ref{Eqn:max-margin-opt}) 
\STATE~~~~~only with the two constraints (\ref{Eqn:max-margin-opt}a) and (\ref{Eqn:max-margin-opt}b).
\STATE Initialize $t = 1$.
 \REPEAT
  \STATE \textit{Step 1}: Fixing ${\mathbf o}={\mathbf o}^{(t-1)}$ and ${\mathbf \Upsilon} = {\mathbf \Upsilon}^{(t-1)}$, optimize Eqn.~(\ref{Eqn:max-margin-opt}) with the constraints (\ref{Eqn:max-margin-opt}a $\sim$ \ref{Eqn:max-margin-opt}c) to update ${\boldsymbol \Theta}^{(t)}$, $r^{(t)}$ and $\epsilon_i^{(t)}$;
  \STATE \textit{Step 2}: Fixing ${\boldsymbol \Theta}^{(t)}$, optimize 
Eqn.~(\ref{Eqn:single-class}) to update ${\mathbf o}^{(t)}$ and ${\mathbf \Upsilon}^{(t)}$ to enforce DAG.
\STATE Let $t = t+1$
  \UNTIL{convergence/max number of iterations}
\vspace{3mm}
 \STATE {\bfseries Output:} ~${\boldsymbol \Theta}^{\star}={\boldsymbol \Theta}^{(t)}$
\end{algorithmic}
\end{algorithm}

It is worthy noting that, we learn an SGBN model for each individual class in order to meet the requirement of both interpretation and discrimination in exploratory research. For example, each SGBN may model the brain network of the healthy or the diseased class, as well as carrying the essential class discrimination. Both the network modelling and the discrimination are of interest in such cases. Our method is different from the conventional BN classifers~\cite{Pernkopf-JMLR-2010,Pernkopf-TPAMI-2012,Guo-UAI-2005} that solely focus on classification. In those methods, only a single BN is learned to reflect the ``difference" of the two classes. It does not model any individual class as our method does, and hence deviates from our purpose of both representing and discriminating brain networks. Moreover, the works in~\cite{Pernkopf-JMLR-2010,Pernkopf-TPAMI-2012,Guo-UAI-2005} cannot handle the continuous variables of brain imaging measures, and inherit the drawbacks of the traditional two-stage methods.

\subsection{Discussion and Analysis}\label{subsec:discussion}
In the following, some issues regarding the two proposed discriminative learning frameworks are discussed.

\textbf{Classifiers.} The proposed discriminative learning frameworks produce a set of jointly learned SGBN models, one for each class. Based on these SGBN models, two kinds of classifiers can be constructed, i.e., the SGBN classifier and the SVM classifier.  The SGBN classifier categorizes a sample by comparing the sample's likelihood according to each SGBN model. The SVM classifier is trained by the sample-specific Fisher vectors induced from the SGBN models. These two classifiers are tightly coupled by the underlying SGBN models. Specifically, more discriminative SGBN models directly lead to a better SGBN classifier, and can provide discriminative Fisher vectors to SVM for better classification. Rooted in this relationship, both the KL-SGBN and the MM-SGBN can improve the classification performance of these two classifiers simultaneously. Put simply, KL-SGBN explicitly optimizes the SVM classifier and in turn implicitly improves the SGBN classifier; while MM-SGBN explicitly optimizes the SGBN classifier, bringing an implicit improvement of the SVM classifier as well.  When evaluating the discriminative power of the learned SGBN models by the SGBN classifier (a direct measurement), it is therefore expected that MM-SGBN can outperform KL-SGBN. However, KL-SGBN has some computational advantages and provides a new perspective to manipulate BN models, analyzed as follows. 

\textbf{Computational Issues.}~~Compared with KL-SGBN, MM-SGBN requires to solve more complicated optimization problems, which may become problematic when the number of training samples increase. Let us compare Eqn.~(\ref{eqn:fisher-prime}) for KL-SGBN and Eqn.~(\ref{Eqn:max-margin-opt}) for MM-SGBN. For KL-SGBN, Eqn.~(\ref{eqn:fisher-prime}) optimizes $J({\boldsymbol \Theta})$ with two convex quadratic constraints of data fitting and two DAG constraints, which are independent of the number of training samples $n$. The evaluation of $J({\boldsymbol \Theta})$ needs to solve an SVM-like problem in Eqn.~(\ref{eqn:svm_original}), taking just $n$ linear constraints of ${\boldsymbol \Theta}$, which could be efficiently solved by off-the-shelf SVM packages. For MM-SGBN, in addition to the data fitting and DAG constraints as in Eqn.~(\ref{eqn:fisher-prime}), the optimization problem in Eqn.~(\ref{Eqn:max-margin-opt}) also has to satisfy $n$ non-convex quadratic constraints. When $n$ increases to a medium or large value, the optimization problem could be quite hard to solve.

\textbf{Edge Selection.}~~In addition to the discriminative learning of SGBN, the employment of Fisher kernel in KL-SGBN also provides a new perspective of edge selection for GBN. As introduced in Section~\ref{subsec:fisher-kernel}, applying Fisher kernel on GBN produces sample-specific feature vectors whose component is the gradient of the log likelihood, i.e., $\frac{\partial {\mathcal L}({\mathbf x}| {\boldsymbol \Theta})}{\partial {\boldsymbol \Theta}_{ij}}$.  In other words, each feature now corresponds to an edge ${\boldsymbol \Theta}_{ij}$ in the SGBN.  This makes it possible to convert the SGBN edge selection to a more traditional feature selection problem that has been well studied and has a large body of options in the literature. 
Edge selection has been employed in our work to deal with the ``small sample size" problem that is often encountered in medical applications. For example, it is common to have only $100$ training samples but $3200$ parameters (for SGBNs of 40 nodes from two classes) to learn in brain network analysis. To handle this issue, we keep using the whole ${\boldsymbol \Theta}$ for computing ${\mathbf K}_{\boldsymbol \Theta}$, but only optimize a selected subset ${\boldsymbol \Theta}_{s}$. There are many options to determine ${\boldsymbol \Theta}_{s}$. We just compute the Fisher vector $\Phi_{\boldsymbol \Theta}$ for each sample, calculate the Pearson correlation between each component of $\Phi_{\boldsymbol \Theta}$ and the class labels on the training data, and select the top ${\boldsymbol \theta}_i$ with the highest correlations. To keep our problem simple, only the parameters associated with edges present in the graph are optimized to avoid the violation of DAG. It is remarkable that even this simple selection process has significantly improved the discrimination for both KL-SGBN and MM-SGBN. Note that this edge selection step is essentially different from that of the traditional two-stage methods. It is just an empirical method to handle the small sample size problem and will become \textit{unnecessary} when sufficient training data are available. In contrast, identifying the candidate-parent sets is an indispensable step in two-stage methods to obtain computationally tractable solutions. 

\textbf{Generalization.}~~We would like to point out that our learning framework of KL-SGBN could be easily generalized. It could be used to discriminatively learn the parameters of distributions other than that represented by GBN by just simply switching GBN to the target distribution, such as Gaussian Mixture Model (GMM). Indeed, this has been seen in~\cite{Sydorov-CVPR-2014}, after our work~\cite{Zhou-CVPR-2013}. However, as shown in this paper, the Fisher vector of GBN is a linear function of the model parameters, which significantly simplifies the learning problem. This favorable property may not be guaranteed with other distributions, including GMM. 
\section{Proposed DAG Constraint}\label{sec:DAG-constraint}
In this section,  we revisit H-SGBN and propose a new DAG constraint that could simplify the optimization problems in SGBN and its discriminative learning process as introduced in Sections~\ref{sec:KL-SGBN} and \ref{sec:MM-SGBN}. 
\subsection{H-SGBN Revisited}\label{subsec:revisit-DAG}
Recall that, the DAG constraint in H-SGBN (Section~\ref{subsec:orig-SGBN}) utilizes the matrix ${\mathbf P}$, an implicit function of ${\boldsymbol \Theta}$, which significantly complicates the optimization problem in Eqn.~(\ref{eqn:Huang_origin}). In~\cite{Huang-TPAMI-2012}, for simplicity, in each optimization iteration, ${\mathbf P}$ is first treated as a constant while optimizing ${\boldsymbol \Theta}$, and then recalculated by searching on the binarized new ${\boldsymbol \Theta}$. This hard binarization could introduce high discontinuity of ${\boldsymbol \Theta}$ into the optimization. Solving ${\boldsymbol \Theta}$ column-wisely by BCD may mitigate this problem since only one column of ${\boldsymbol \Theta}$ is changed in each iteration, inducing less discontinuity. However, we observe that the solution of BCD depends on which column of ${\boldsymbol \Theta}$ to be optimized first. In other words, if we randomly permute the ordering of features (the columns in ${\mathbf X}$), we will obtain different SGBNs, which impairs the interpretability of the SGBN model. The optimization ordering matters because the matrix ${\mathbf P}$ used in the DAG constraint changes with the ordering. This problem has been demonstrated in our experiment. Moreover, we find experimentally that if ${\mathbf P}$ is solved as a whole instead of BCD, the optimization in Eqn.~(\ref{eqn:Huang_origin}) will not converge but oscillate between some \textit{non-DAG} solutions, possibly due to the high discontinuity mentioned above~\footnote{Please note that, solving ${\boldsymbol \Theta}$ column-wisely without updating ${\mathbf P}$ in each iteration will only lead to non-DAG solutions}. Early stop cannot help because no premature
solution satisfies DAG. These optimization difficulties motivate our work of proposing a new DAG constraint that is much simpler for SGBN, as described below.

\subsection{Proposed DAG constraint }\label{subsec:topological-ordering}
It is known that, a BN is equivalent to a topological ordering (Page 362 in~\cite{Bishop-2007}). Therefore, we propose a new DAG constraint applicable to continuous variables with GBN based on this equivalence. With a few linear inequalities and variables separable from ${\boldsymbol \Theta}$, the new DAG constraint significantly simplifies that used in~\cite{Huang-TPAMI-2012}. Specifically, given a directed graph $\mathcal G$ and the parameters ${\boldsymbol \Theta}$, a real-valued order variable $o_i$ is assigned to each node $i$, where $0 \leq {o}_i \leq \Delta$, and $\Delta$ is a predefined arbitrary positive number. We propose a sufficient and necessary condition for $\mathcal G$ to be DAG as in Proposition 1.

\textbf{Proposition 1.} Given a sparse Gaussian Bayesian Network parameterized by ${\boldsymbol \Theta}$ and its associated directed graph $\mathcal G$ with $m$ nodes,  the graph $\mathcal G$ is DAG if and only if there exist some $o_i$ ($i=1,\cdots,m$) and ${\boldsymbol \Upsilon} \in {\mathbb R}^{m \times m}$, such that for arbitrary $\Delta > 0$, the following constraints are satisfied:
\begin{align}\label{eqn:DAG_constraint}
&\\
& o_j - o_i \geq \frac{\Delta}{m} - {\boldsymbol \Upsilon}_{ij},~~ \forall i, j \in \{1,\cdots,m\},~~i\neq j  \tag{\ref{eqn:DAG_constraint}a}\\
& {\boldsymbol \Upsilon}_{ij} \geq 0, \nonumber \tag{\ref{eqn:DAG_constraint}b}\\
&{\boldsymbol \Upsilon}_{ij}\times {\boldsymbol \Theta}_{ij} = 0,\nonumber \tag{\ref{eqn:DAG_constraint}c}\\
&\Delta \geq o_i \geq 0. \nonumber \tag{\ref{eqn:DAG_constraint}d}
\end{align}
Eqn.(\ref{eqn:DAG_constraint}) leads to a topological ordering equivalent to DAG. The topological ordering means that if node $j$ comes after node $i$ in the ordering ($o_j>o_i$), there cannot be a link from node $j$ to node $i$, which guarantees the acyclicity. The proof of Proposition 1 is given in Appendix.

By Proposition 1, we remove the awkward hard binarization for computing P in~\cite{Huang-TPAMI-2012}. The inequalities of (\ref{eqn:DAG_constraint}a, \ref{eqn:DAG_constraint}b, \ref{eqn:DAG_constraint}d) are linear to the ordering variables $o_i$ and ${\boldsymbol \Upsilon}$.  The equation~(\ref{eqn:DAG_constraint}c) differs from the equation ${\boldsymbol \Theta}_{ji}\times{{\mathbf P}_{ij}}=0$ in~\cite{Huang-TPAMI-2012} in that the variable ${\boldsymbol \Upsilon}_{ij}$ is now separable from ${\boldsymbol \Theta}_{ij}$ (while ${\mathbf P}_{ij}$ is not) and does not require the binarization of ${\boldsymbol \Theta}$. This makes it tractable to solve ${\boldsymbol \Theta}$ as a whole instead of BCD (to avoid the feature ordering problem).

It is worth noting that, provided ${\boldsymbol \Theta}$ is sparse, the number of constraints in Eqn.~(\ref{eqn:DAG_constraint}) could be significantly reduced. As can be seen, for any ${\boldsymbol \Theta}_{ij}=0$, as long as we set the corresponding ${\boldsymbol \Upsilon}_{ij}$ an arbitary value greater than $(\frac{1}{m}+1)\Delta$, all the conditions in Eqn.~(\ref{eqn:DAG_constraint}) will be automatically satisfied. Therefore, we only need to consider the constraints related to ${\boldsymbol \Theta}_{ij}\neq 0$.

The idea of topological ordering is also used to design DAG constraint for the discrete variables in~\cite{Pernkopf-ICML-2012}. However, the work in~\cite{Pernkopf-ICML-2012} addresses the multinominal distribution of discrete variables, while here we target the Gaussian distribution of continuous variables.  It is worthy noting that the constraint in~\cite{Pernkopf-ICML-2012} has to predefine candidate parent-node sets. Therefore, it inherits the drawbacks of the two-stage methods as pointed out in Section~\ref{sec:introduction}. This has been circumvented in our proposed DAG constraint for SGBN. 


\subsection{Estimation of SGBN from A Single Class}\label{subsec:single_class}
 With our DAG constraint proposed in Eqn.~(\ref{eqn:DAG_constraint}), we could estimate SGBN from a single class as the initial solution to our discriminative learning of KL-SGBN or MM-SGBN. In particular, we optimize
\begin{align}\label{Eqn:single-class}
&\underset{{\boldsymbol \Theta}, {\mathbf o}, {\boldsymbol \Upsilon}}\min\sum_{i=1}^m\|{\mathbf x}_{:, i} - {\mathbf {PA}}_i^{\top}{\boldsymbol \theta}_i\|_2^2+\lambda_1\|{\boldsymbol \theta}_i\|_1+\lambda_{dag}{{\boldsymbol \epsilon}_i}^{\top}|{\boldsymbol \theta}_i|\\
&s.t. ~~  o_j - o_i \geq \frac{\Delta}{m} - {\boldsymbol \Upsilon}_{ij},\forall i, j \in \{1,\cdots,m\},~~i\neq j\nonumber\\
&~~~~~~~0 \leq o_i \leq \Delta, ~~ {\boldsymbol \Upsilon}_{ij} \geq 0, \nonumber
\end{align}
where ${\boldsymbol \epsilon}_i$ is the $i$-th column of the matrix ${\boldsymbol \Upsilon}$, and $|{\boldsymbol \theta}_i|$ the component-wise absolute value of ${\boldsymbol \theta}_i$. This optimization problem is solved in an iterative way with two alternate steps in each iteration:  i) optimize ${\mathbf o}$ and ${\boldsymbol \Upsilon}$ (with ${\boldsymbol \Theta}$ fixed) and ii) optimize ${\boldsymbol \Theta}$ (with ${\mathbf o}$ and ${\boldsymbol \Upsilon}$ fixed). This process is repeated until convergence.  We call this proposed method OR-SGBN  (Algorithm~\ref{alg:OR-SGBN}). 

When the coefficient $\lambda_{dag}$ is sufficiently large, the alternate optimization strategy of Eqn.~(\ref{Eqn:single-class}) will converge to a DAG solution, as shown in Proposition 2 in Appendix. In practice, for numerical stability, we adopt a ``warm start" strategy as in~\cite{Huang-TPAMI-2012}, that is, to gradually increase the values of $\lambda_{dag}$ until the resulting ${\mathcal G}$ becomes DAG. Specifically, we use a set of values of $\lambda_{dag}$: $\lambda_{dag}^{(1)} < \lambda_{dag}^{(2)} <\cdots <\lambda_{dag}^{(M)}$ to solve Eqn.~(\ref{Eqn:single-class})  (Algorithm~\ref{alg:OR-SGBN}).
\begin{algorithm}[tb]
   \caption{~~OR-SGBN: SGBN from a single class}
\label{alg:OR-SGBN}
\begin{algorithmic}
\STATE {\bfseries Input:} data ${\mathbf X} \in {\mathbb R}^{n\times m}$
\vspace{3mm}
\STATE Initialize ${\boldsymbol \Theta}^{(0)}$ by least square fitting.
\STATE Initialize ${\mathbf o}^{(0)}$ and ${\mathbf \Upsilon}^{(0)}$ by solving Eqn.~(\ref{Eqn:single-class}) with ${\boldsymbol \Theta}={\boldsymbol \Theta}^{(0)}$.

\STATE Let $T = 1$.
 \REPEAT
\STATE Fixing ${\mathbf \Upsilon} = {\mathbf \Upsilon}^{(T-1)}$ and ${\mathbf o} = {\mathbf o}^{(T-1)}$.
\STATE Let $t=1$,~~${\boldsymbol \Theta}^{(T-1,  t=0)}={\boldsymbol \Theta}^{(T-1)}$.
 \FOR{$\lambda_{dag}=\lambda_{dag}^{(1)}$  {\bfseries to} $\lambda_{dag}^{(M)}$ }
\STATE  Optimize Eqn.~(\ref{Eqn:single-class}) with the initial solution ${\boldsymbol \Theta}^{(T-1,  t-1)}$ to obtain ${\boldsymbol \Theta}^{(T-1, t)}$.
\STATE Let t=t+1.
\ENDFOR
\STATE Let ${\boldsymbol \Theta}^{(T)}={\boldsymbol \Theta}^{(T-1, M)}$.
  \STATE  Fixing ${\boldsymbol \Theta}^{(T)}$, optimize Eqn.~(\ref{Eqn:single-class}) to update ${\mathbf o}^{(T)}$ and ${\mathbf \Upsilon}^{(T)}$ to enforce DAG.
\STATE Let $T = T+1$.
 \UNTIL{convergence/max number of iterations}
\vspace{3mm}
\STATE {\bfseries Output:} ~${\boldsymbol \Theta}^{\star}={\boldsymbol \Theta}^{(T)}$
\end{algorithmic}
\end{algorithm}

 We use a bias variable $x_0=1$ in the regression model to improve data fitting, thus $x_i = [{\boldsymbol \theta}_i~~ {\boldsymbol \theta}_0]^{\top}[{\mathbf {Pa}}_i~~1]+\varepsilon_i$ ($i>1$). In the following part,  we denote ${\boldsymbol \theta}_i \triangleq [{\boldsymbol \theta}_i~~ {\boldsymbol \theta}_0]$ and ${\mathbf {Pa}}_i \triangleq [{\mathbf {Pa}}_i~~1]$. The bias term ${\boldsymbol \theta}_0$ is learned together with other ${\boldsymbol \theta}_i$. This equals to introducing a bias node into the graph. It has no parent but is the parent of all the other
nodes. If the original graph is a DAG, this does not cause the violation of DAG. 

It is interesting yet challenging to analyze the network consistency of OR-SGBN. It is noted that Eqn.~(\ref{Eqn:single-class}) can be reorganized into a weighted LASSO problem, which can be conceptually linked to ``adaptive LASSO" in the literature~\cite{Zou-JASA-2006,Shojaie-Biometrika-2010,Fu-JASA-2013}. The analysis framework provided by these works is suggestive of promising strategies to analyze the network consistency for L1-penalized Gaussian networks. However, a complete treatment of this analysis for OR-SGBN requires a deep investigation. Considering the significant 
amount of the required workload and its importance, we will explore this
problem in a separate paper in our future work.

\section{Experiment}\label{sec:experiment}
In this section, we investigate the properties of our proposed methods from three aspects: the DAG constraint, the discriminative learning process, and the resulting connectivity for brain network analysis. Four experiments are conducted, summarized in Table~\ref{Table:experiments}. The data sets and the experiments are elaborated as follows.

\begin{table*}[!h!t]
\caption{\textbf{\textit{Summary of Experiment Purpose}}}\label{Table:experiments}
\vspace{-2mm}
\begin{center}
\renewcommand{\arraystretch}{1.5}
\begin{tabular}{c|c|l}
\hline
Experiment & Test Subject & ~~~~~~~~~~~~~~~~~~~~~~~~~~~~Purpose\\
\hline
Exp-I (Sec.~\ref{subsec:compare_DAG}) & DAG constraint & Test the invariance of solution to feature ordering\\
Exp-II (Sec.~\ref{subsec:compare_DAG}) & DAG constraint & Test the ability of network structure recovery\\
Exp-III (Sec.~\ref{subsec:compare_discrimination}) & discriminative learning & Test the improvement of discriminative power of SGBN models\\
Exp-IV (Sec.~\ref{subsec:compare_connectivity})& brain network analysis & Investigate the learned brain connectivity patterns\\
\hline
\end{tabular}
\end{center}
\vspace{-4mm}
\end{table*}
\subsection{Neuroimaging Data Sets}\label{subsec:determine-node}
We conduct our experiment on the publicly accessible ADNI~\cite{ADNI} database to analyze brain effective connectivity for the Alzheimer's disease. Three data sets are used from two imaging modalities of MRI and FDG-PET downloaded from ADNI. They are elaborated as follows.

\noindent \textbf{MRI} data set includes 120 T1-weighted MR images belonging to 50 mild cognitive impairment (MCI) patients and 70 normal controls (NC). These images are preprocessed by the typical procedure of intensity correction, skull stripping, and cerebellum removal. We segment the images into gray matter (GM), white matter (WM), and cerebrospinal fluid (CSF) using the standard FSL\footnote{http://fsl.fmrib.ox.ac.uk/fsl/fslwiki/} package, and parcellate them into 93 Region of Interest (ROI) based on an ROI atlas~\cite{Kabani-Neuroimage-1998} after spatial normalization. The GM volume of each ROI is used as the imaging feature to characterize each network node. Forty ROIs are included in this study, following~\cite{Huang-TPAMI-2012}. They  have higher correlation with the disease and are mainly located in the temporal lobe and subcortical region.  Studying brain morphology as a network can take the advantage of statistical tools from graph theory. Moreover, it has been reported that the covariation of gray matter morphology might be related to the anatomical connectivity~\cite{Tijms-cereb-2012}.

\noindent \textbf{PET} data set includes 103 FDG-PET images (and their corresponding MR images) of 51 AD patients and 52 NC.  The MR images belonging to different subjects are co-registered and partitioned into ROIs as before. The ROI partitions are copied onto their corresponding PET images by a rigid transformation. The average tracer uptakes within each ROI is used as the imaging feature to characterize each network node.  Forty ROIs discriminative to the disease are used in the study. The retention of tracer in FDG-PET is analogous to the glucose uptake,  thus reflecting the tissue metabolic activity. 

\noindent \textbf{MRI-II} data set is similar to the MRI data set but using 40 different ROIs covering the typical brain regions spread over the  frontal, parietal, occipital and temporal lobes.

We randomly partition each data set into 30 groups of training-test pairs.  Each group includes 80 training and 40 test samples in MRI and MRI-II, or 60 training and 43 test samples in PET.
\subsection{DAG Constraint}\label{subsec:compare_DAG}
With our proposed DAG constraint,  the SGBN model for an individual class can be learned with all the parameters ${\boldsymbol \Theta}$ optimized together (OR-SGBN), instead of column-wisely as did in~\cite{Huang-TPAMI-2012,Zhou-CVPR-2013,Zhou-MICCAI-2014}. To explore the properties of our DAG constraint, we test three experimental configurations, namely, OR-SGBN (WHOLE), H-SGBN (BCD) and H-SGBN (WHOLE). The word in the parenthesis is used to explicitly indicate whether the parameters ${\boldsymbol \Theta}$ are optimized together (WHOLE) or column-wisely (BCD).  OR-SGBN (WHOLE) is our SGBN learning method for a single class in Algorithm~\ref{alg:OR-SGBN}, implemented with the package of CVX. H-SGBN (BCD) is the column-wise method in~\cite{Huang-TPAMI-2012} and implemented with the code downloaded from the authors' website. H-SGBN (WHOLE) is our attempt to optimize ${\boldsymbol \Theta}$ together for the objective function of H-SGBN in~\cite{Huang-TPAMI-2012}, which is implemented with the package of CVX\footnote{The optimization problem is solved by a series of convex sub-problems.}. The same ${\boldsymbol \Theta}$ that is computed by a sparse least square fitting of the training set is provided to all the methods to initialize the optimizations. The ``warm-start" strategy is applied wherever applicable in all methods.

It is found that when solving all ${\boldsymbol \Theta}$ as a whole, H-SGBN (WHOLE) that uses the DAG constraint in~\cite{Huang-TPAMI-2012} does not converge: the optimization is trapped to oscillate between a few solutions that are not DAG. Therefore,  from now on, we only consider  H-SGBN (BCD) and OR-SGBN (WHOLE).

\begin{centering}
\begin{figure}[ht]
\begin{center}
\begin{tabular}{cc}
Original & After Permutation\\
\includegraphics[width=0.22\textwidth]{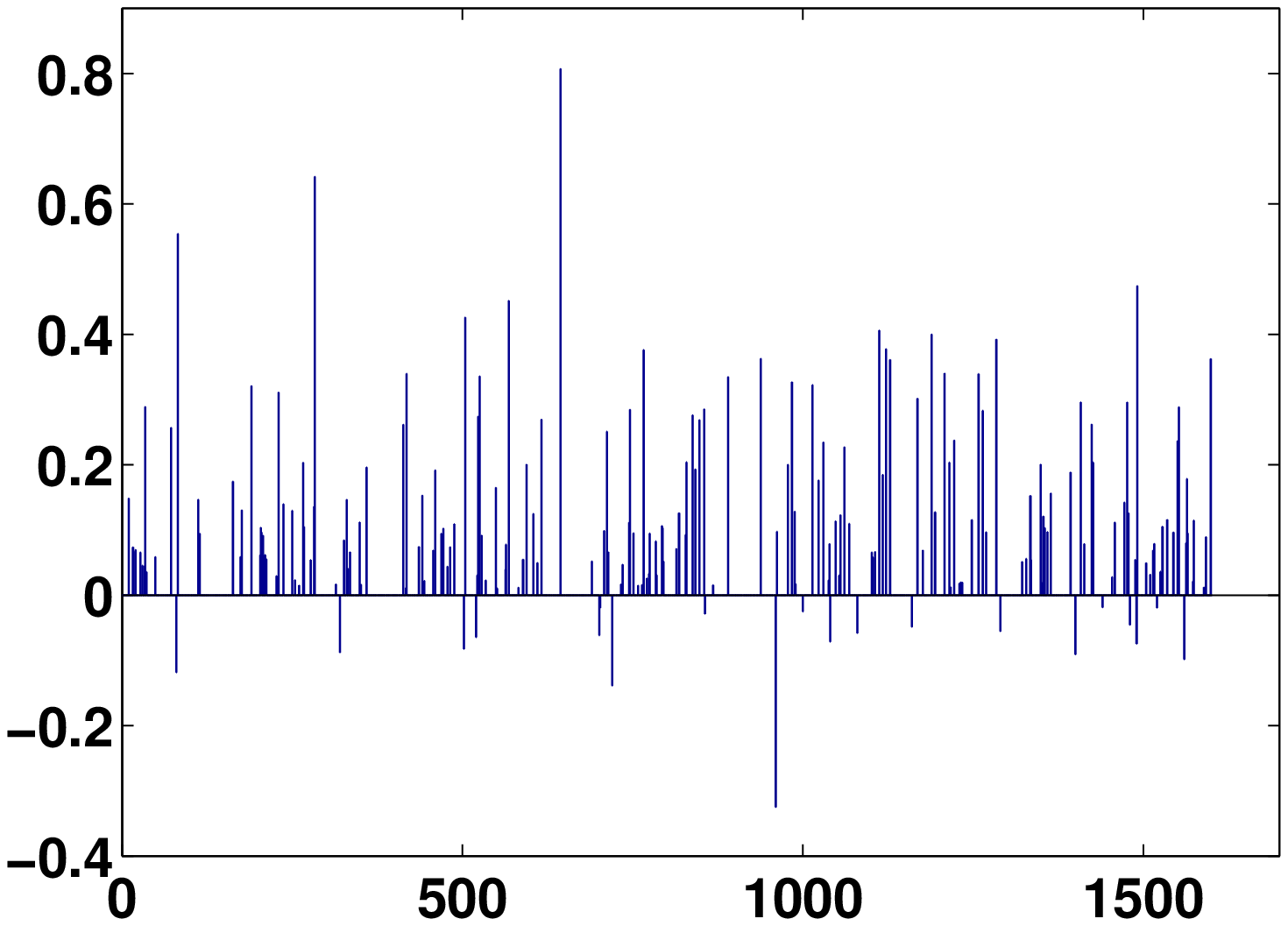}
&\includegraphics[width=0.22\textwidth]{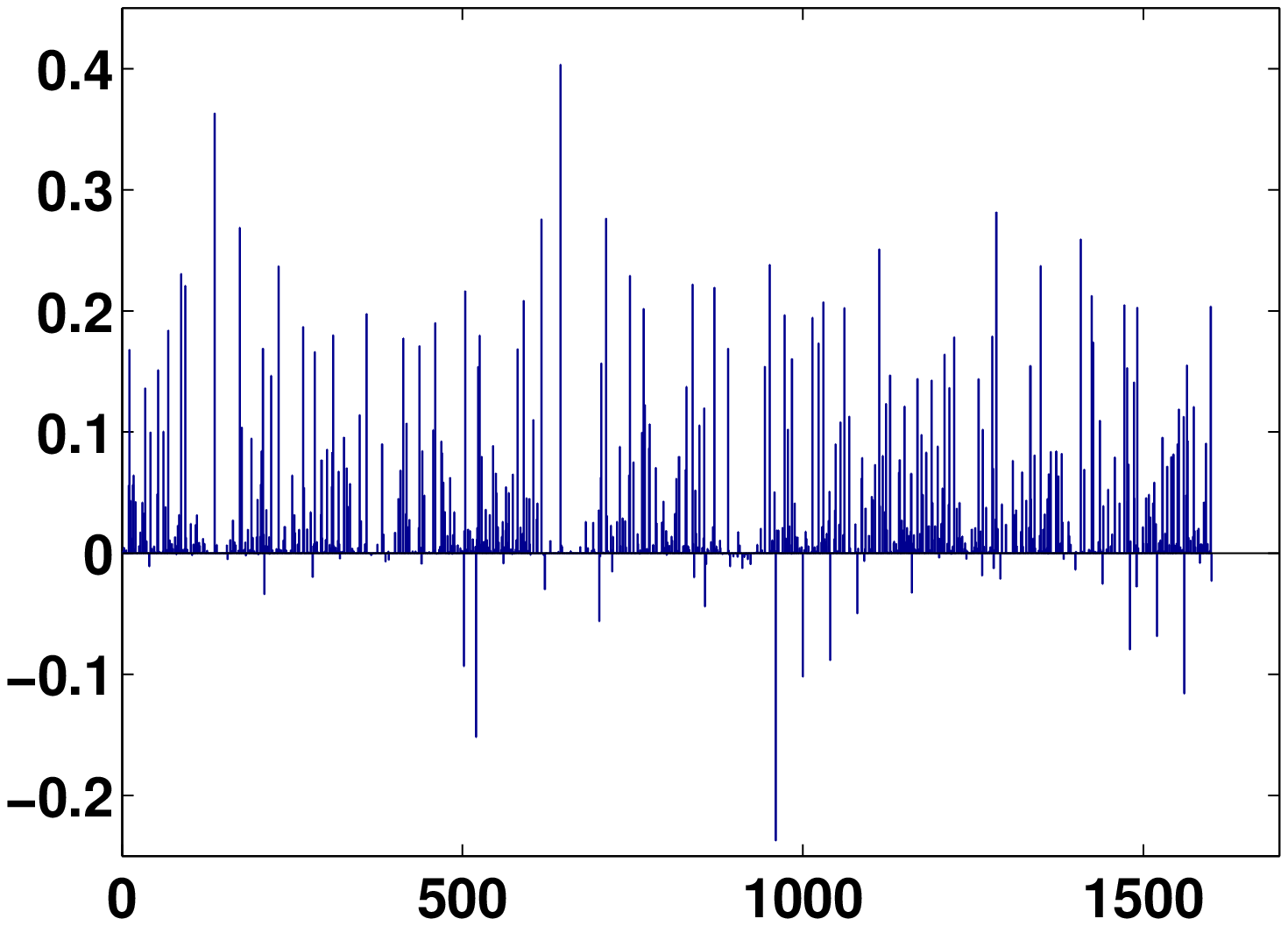}\\
\small{(a) H-SGBN (BCD)} & \small{(b) H-SGBN (BCD)}  \vspace{5mm}\\
Original & After Permutation\\
\includegraphics[width=0.22\textwidth]{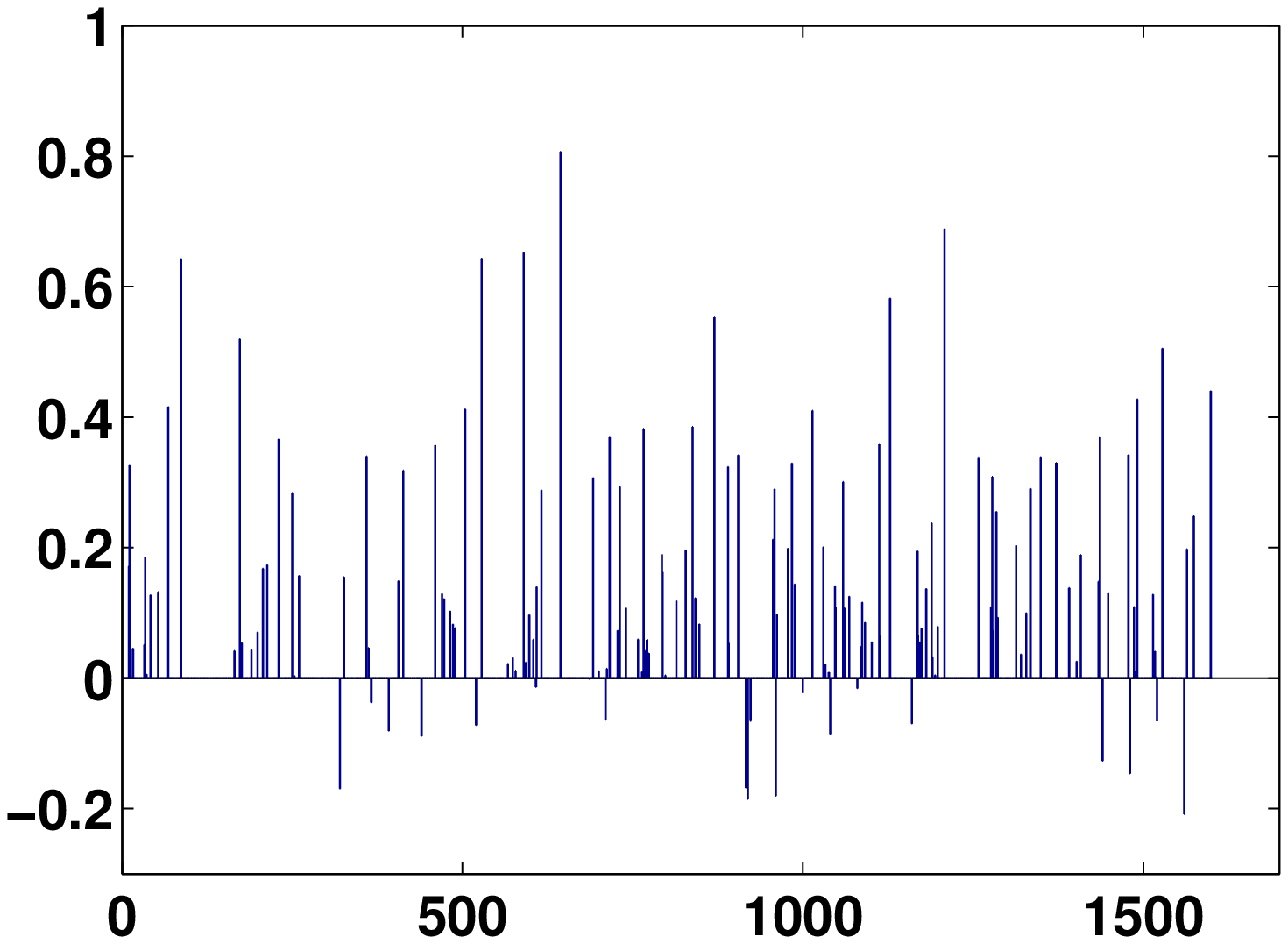}
&\includegraphics[width=0.22\textwidth]{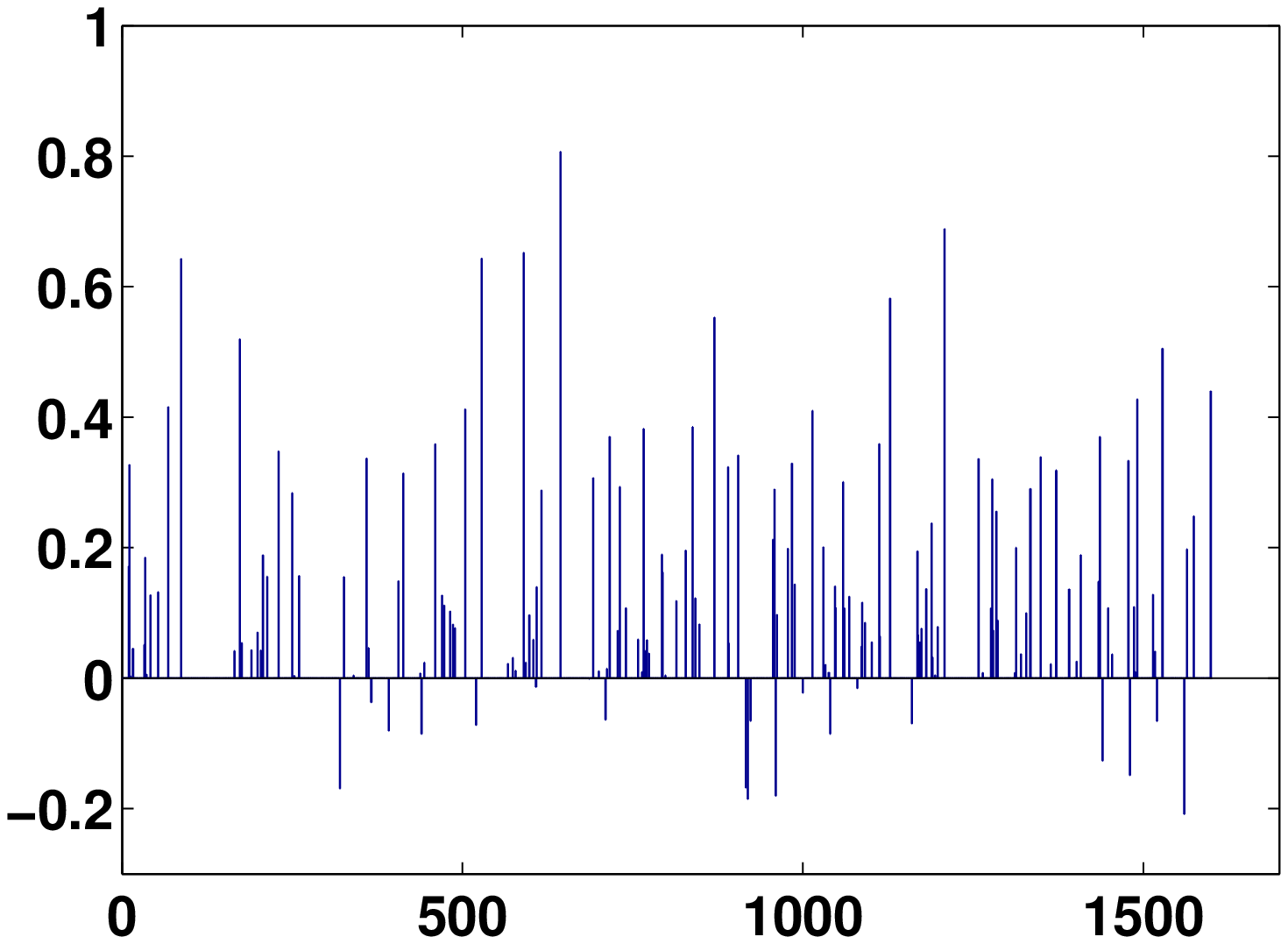}\\
\small{(c) OR-SGBN (WHOLE)} & \small{(d) OR-SGBN (WHOLE)}
\end{tabular}
\caption{\em One example of the estimated parameter ${\boldsymbol \Theta}$ for the MCI class (reshaped as a long vector) with regard to the random permutation of the feature ordering. Quantitative measurements of the changes are given in Table~\ref{Table:diff-randperm}.
}
\label{fig:DAG-randperm}
\end{center}
\vspace{-4mm}
\end{figure}
\end{centering}

\textbf{Exp-I.}~~In this experiment, we compare the solutions of OR-SGBN (WHOLE) and H-SGBN (BCD) with respect to the change of feature ordering. To do that, for the neuroimaging data sets, we randomly permute the feature ordering for 100 times. The estimated ${\boldsymbol \Theta}$ of the resulting 100 SGBNs are re-arranged according to the initial feature ordering and then averaged as in Fig.~\ref{fig:DAG-randperm}. As shown, the averaged result from OR-SGBN (WHOLE) (Fig.~\ref{fig:DAG-randperm} (d)) is almost identical to the result using the original feature ordering (Fig.~\ref{fig:DAG-randperm} (c)), reflecting its robustness to feature ordering.  In contrast, H-SGBN (BCD) generates SGBNs with large variations when the feature ordering changes ((Fig.~\ref{fig:DAG-randperm} (a) versus (b)). To give a quantitative evaluation, the Euclidean distance and the correlation between the averaged ${\boldsymbol \Theta}$ and the original ${\boldsymbol \Theta}$ are presented in Table~\ref{Table:diff-randperm}. Consistently, the solutions from OR-SGBN (WHOLE) are much less affected by the ordering permutation, indicating the advantage of solving ${\boldsymbol \Theta}$ as a whole via the proposed DAG constraint. 
\begin{table}[!h]
\caption{\textbf{\textit{Quantitative Analysis of ${\boldsymbol \Theta}$ for the random permutation of feature ordering (between the original and the averaged ${\boldsymbol \Theta}$)}}}\label{Table:diff-randperm}
\vspace{-2mm}
\begin{center}
\renewcommand{\arraystretch}{1.5}
\begin{tabular}{|c|c|c|c|}
\hline \multicolumn{2}{|c|}{}& Distance & Correlation (R)\\\hline
\hline
OR-SGBN & ${\boldsymbol \Theta}_1$ & 0.08 & 0.9996\\
\cline{2-4} (WHOLE) & ${\boldsymbol \Theta}_2$ & 0.18 & 0.9981\\\hline
\hline H-SGBN & ${\boldsymbol \Theta}_1$ & 1.91 & 0.6828\\
\cline{2-4} (BCD) & ${\boldsymbol \Theta}_2$ & 2.06 & 0.6396\\
\hline
\end{tabular}
\end{center}
\end{table}

\textbf{Exp-II.}~~In this experiment, we test the ability of OR-SGBN (WHOLE) at identifying network structures from data. Since no ground-truth is available for the three neuroimaging data sets due to the unknown mechanism of the disease, we conduct experiments on nine benchmark network data sets mostly coming from the Bayesian Network Repository~\cite{BNR} as was done in the literature~\cite{Schmidt-AAAI-2007,Mackey-2003}. The nine benchmark data sets are: Factors (27 nodes, 68 arcs), Alarm (37 nodes, 46 arcs), Barley (48 nodes, 84 arcs), Carpo (61 nodes, 74 arcs), Chain (7 nodes, 6 arcs), Hailfinder (56 nodes, 66 arcs), Insurance (27 nodes, 52 arcs), Mildew (35 nodes, 46 arcs) and Water (32 nodes, 66 arcs). We compare the OR-SGBN (WHOLE) with another eight BN learning methods, including L1MB~\cite{Schmidt-AAAI-2007}, GS~\cite{Margaritis-NIPS-1999}, TC and its variant TC-bw~\cite{Pellet-JMLR-2008} and three variants of IAMB~\cite{Tsamardinos-IWAIS-2003}. The experiment is repeated for 50 simulations. In each simulation, for each network, we randomly sample 1000 samples from $\pm\mbox{Uniform}(0.5, 1)$ for the regression coefficients of each variable on its parents. The parameters of the eight methods to be compared are set according to~\cite{Huang-TPAMI-2012}. A predefined $\lambda$ that controls the sparsity of OR-SGBN is uniformly applied to all the nine data sets, which simply brings the number of the resulting edges to a reasonable range~\footnote{The Bayesian Information Criterion is used to select $\lambda$ in ~\cite{Huang-TPAMI-2012}. However, it did not behave well in our experiment.}. We use the first stage estimate of L1MB as the initial solution of OR-SGBN. Table~\ref{Table:benchmark-errs} shows the total numbers of mis-identified edges (including both the false and the missing edges), while Table~\ref{Table:benchmark-FPs} shows the numbers of falsely identified edges (false positive). In addition, Table~\ref{Table:benchmark-PDAGs} lists the numbers of falsely identified PDAG structures. PDAG structures are statistically indistinguishable structures, i.e., representing the same statistical dependency. The PDAG of BN is obtained by the method in~\cite{Spirtes-1993}.  From Tables~\ref{Table:benchmark-errs} $\sim$ \ref{Table:benchmark-PDAGs}, it can be seen that OR-SGBN shows significantly smaller errors on six data sets (Factors, Alarm, Barley, Carpo, Hailfinder and Insurance) in identifying both edges and PDAG structures. For the data sets of Mildew and Water, OR-SGBN performs similarly to the other methods. It only performs relatively inferior on Chain. This experiment demonstrates that the proposed DAG constraint for SGBN can perform effectively for BN structure identification. Its relatively low risk of mis-edge identification is a favorable property for exploratory research.


\begin{table*}[!th]
\caption{\textbf{\textit{Total errors (number of both false and missing edges, averaged on 50 simulations) on benchmark networks}}}\label{Table:benchmark-errs}
\vspace{-3mm}
\begin{center}
\renewcommand{\arraystretch}{1.0}
\begin{tabular}{|c||c|c|c|c|c|c|c|c|c|}
\hline
&L1MB & GS & TC-bw & TC & IAMB & IAMB1 &IAMB2 &IAMB3&OR-SGBN\\\hline\hline
Factors & 101.48 &104.50 & 102.90 & 103.02 & 103.14 & 103.30 & 103.14 & 103.14 &  \textbf{54.82}\\\hline
Alarm & 56.58 & 59.30 &  57.76 &  60.60  &  61.76 & 59.16 &  61.76 & 61.76 & \textbf{44.40}\\\hline
Barley & 113.24& 114.70&  114.38 & 122.78 & 123.80 & 109.92 & 123.80 & 123.80 & \textbf{99.26}\\\hline
Carpo &125.74& 131.72&  131.18 &  133.16 & 132.76 & 132.90 & 132.76 & 132.76 & \textbf{25.58}\\\hline
Chain & 5.32& 4.88 & 5.50& \textbf{4.42} & 4.70 & 5.00 &  4.70 &  4.70  &  7.04\\\hline
Hailfinder & 91.50& 94.94 & 96.18 & 99.02 & 103.10 & 92.74 & 103.10 & 103.10&   \textbf{57.04}\\\hline
Insurance & 74.78 &74.64 &73.74 &76.30 &78.78& 73.04& 78.78& 78.78& \textbf{59.04}\\\hline
Mildew & 60.86  & 60.74 & 59.66 & 63.80 & 68.46 &92.74 &103.10 & 103.10 & \textbf{57.04}\\\hline
Water & 92.86 & 94.04 &  90.24 &  97.16 & 102.70 & \textbf{90.06} & 102.70 & 102.70 & 93.08\\\hline
\end{tabular}
\end{center}
\end{table*}

\begin{table*}[!th]
\caption{\textbf{\textit{Number of falsely identified edges (averaged on 50 simulations) on benchmark networks}}}\label{Table:benchmark-FPs}
\vspace{-3mm}
\begin{center}
\renewcommand{\arraystretch}{1.0}
\begin{tabular}{|c||c|c|c|c|c|c|c|c|c|}
\hline
&L1MB & GS & TC-bw & TC & IAMB & IAMB1 &IAMB2 &IAMB3&OR-SGBN\\\hline\hline
Factors & 47.66  & 50.74 & 49.40 & 49.74 & 50.28 & 49.70 &  50.28 & 50.28 & \textbf{17.70}\\\hline
Alarm & 36.04 & 37.72 & 36.86 & 39.24 & 40.96 & 37.30 & 40.96 & 40.96 & \textbf{23.14} \\\hline
Barley & 71.70 & 72.30 & 72.60 &  80.76 &  82.96 &  69.76 & 82.96 & 82.96 & \textbf{48.70}\\\hline
Carpo &71.96 &  76.30 &  75.14 & 77.38 & 77.18 & 77.36 & 77.18 & 77.18 & \textbf{14.56}\\\hline
Chain & 2.66 & 2.44 & 2.76& \textbf{2.22} & 2.36 & 2.50 & 2.36 & 2.36 & 3.52\\\hline
Hailfinder & 60.00 & 62.04 & 63.16 & 65.42 & 66.40 & 60.90 & 66.40 & 66.40 & \textbf{28.66} \\\hline
Insurance & 42.80 & 42.16 & 41.72 & 44.06 & 48.08 & 41.42 & 48.08 & 48.08 & \textbf{31.20}\\\hline
Mildew & 46.22 & 46.46 & 45.46 & 49.82 & 52.48 & 44.82 & 52.48 & 52.48 & \textbf{33.86}\\\hline
Water & 64.52 & 65.02 & 63.70 & 68.06 & 74.22 & 63.48 & 74.22 & 74.22 & \textbf{46.74}\\\hline
\end{tabular}
\end{center}
\end{table*}

\begin{table*}[!th]
\caption{\textbf{\textit{Number of falsely identified P-DAG structures (averaged on 50 simulations) on benchmark networks}}}\label{Table:benchmark-PDAGs}
\vspace{-3mm}
\begin{center}
\renewcommand{\arraystretch}{1.0}
\begin{tabular}{|c||c|c|c|c|c|c|c|c|c|}
\hline
&L1MB & GS & TC-bw & TC & IAMB & IAMB1 &IAMB2 &IAMB3&OR-SGBN\\\hline\hline
Factors &107.20 & 109.54 & 108.96 & 108.84 & 109.22 &108.84 & 109.22 & 109.22 &  \textbf{63.40}\\\hline
Alarm & 61.74 & 64.08 & 62.54 & 65.34 & 66.82 & 63.98 & 66.82 & 66.82 & \textbf{51.02}\\\hline
Barley & 120.54 & 122.26 & 121.38 & 130.04 & 131.24 & 116.92 & 131.24 & 131.24 & \textbf{105.50}\\\hline
Carpo &129.02 & 135.34 & 134.78 & 136.92 & 136.74 &136.22 & 136.74 & 136.74 & \textbf{33.74}\\\hline
Chain & 5.96 & 5.54 & 6.12 & \textbf{5.16} & 5.30 & 5.66 & 5.30 & 5.30 & 7.42\\\hline
Hailfinder & 103.72 & 106.08 & 107.56 & 110.04 & 113.44 & 104.86 & 113.44 & 113.44 &\textbf{63.76}\\\hline
Insurance &81.58 & 81.68 & 81.44 & 83.70 & 86.60 & 80.66 & 86.60 & 86.60 & 68.26\\\hline
Mildew & 61.68 &  61.32 & 60.34 & 64.48 & 69.30 & \textbf{58.08} & 69.30 & 69.30 & 67.24\\\hline
Water & 96.34  & 97.46 & 93.80 & 100.38  & 106.14 & \textbf{93.60} & 106.14 & 106.14 & 94.52\\\hline
\end{tabular}
\end{center}
\end{table*}
\subsection{Comparison of Discrimination}\label{subsec:compare_discrimination}
After testing the effectiveness of the proposed DAG constraint, we now investigate the theme of this paper: the discriminative learning frameworks. We consider two kinds of classifiers: i) the SGBN classifier (with two SGBN models, one for each class), and ii) the SVM classifier learned by the Fisher vectors induced from the SGBN models.  

\textbf{Exp-III.}~~In this experiment, we test whether our learning methods (KL-SGBN and MM-SGBN) can improve the discriminative power on both kinds of classifiers for the real neuroimaging data sets. The initial SGBN models are obtained by our proposed OR-SGBN (WHOLE), since it has been shown more robust to feature ordering than H-SGBN as above. For the SGBN classifier, assuming equal prior, we assign a test sample to the class with a higher likelihood. For the SVM classifier, we use ${\mathcal L}_2$-SVM with Fisher kernels. In order to maintain representation capability, we allow maximal $1\%$ additional squared fitting errors (that is, $T_i=1.01\times T_{i0},~(i={1, 2})$, where $T_{i0}$ is the squared fitting error of the initial solution) to be introduced during the learning process of KL-SGBN or MM-SGBN.

\begin{centering}
\begin{figure*}[!h!t]
\begin{center}
\begin{tabular}{ccc}
\includegraphics[width=0.33\textwidth]{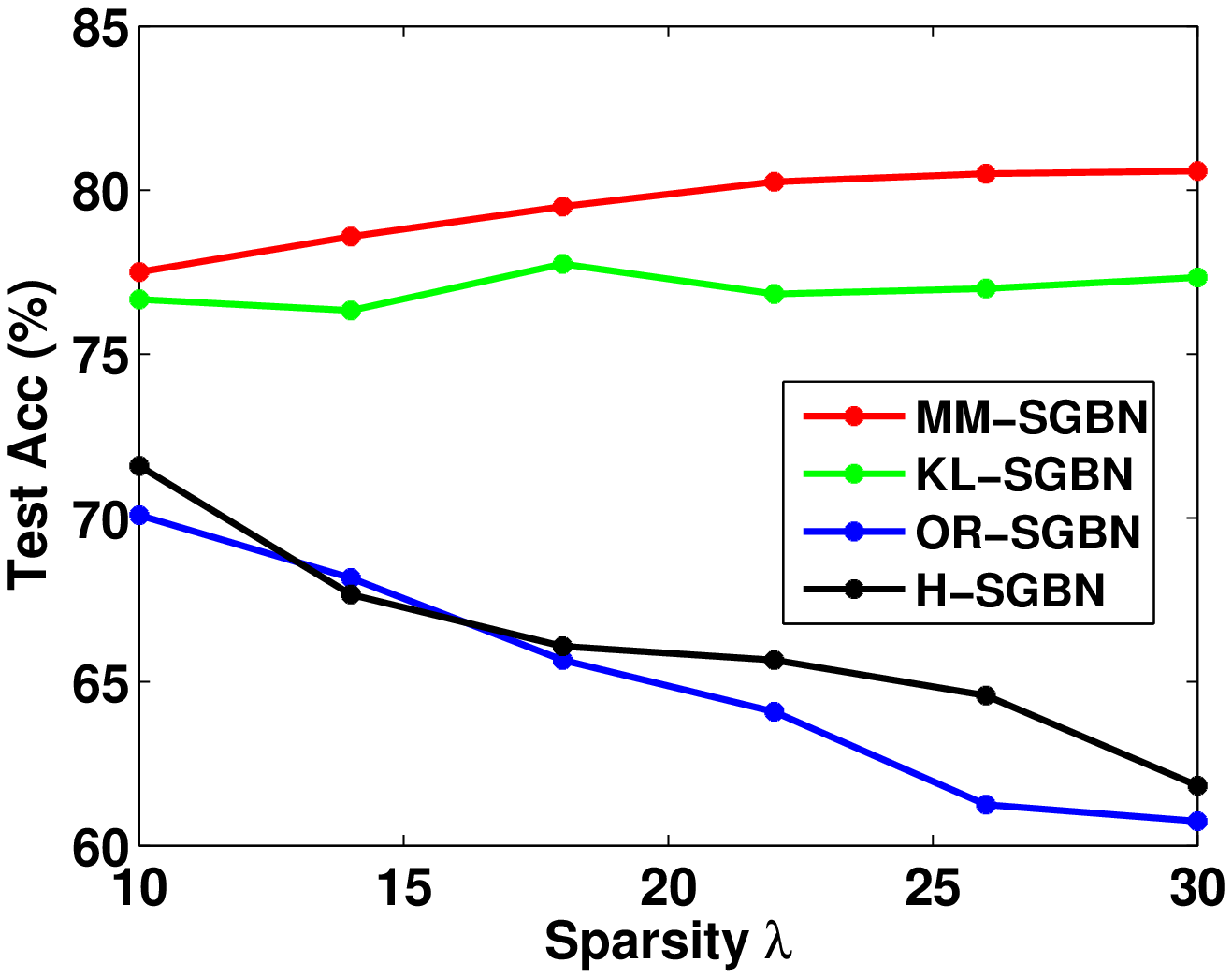}
&\includegraphics[width=0.33\textwidth]{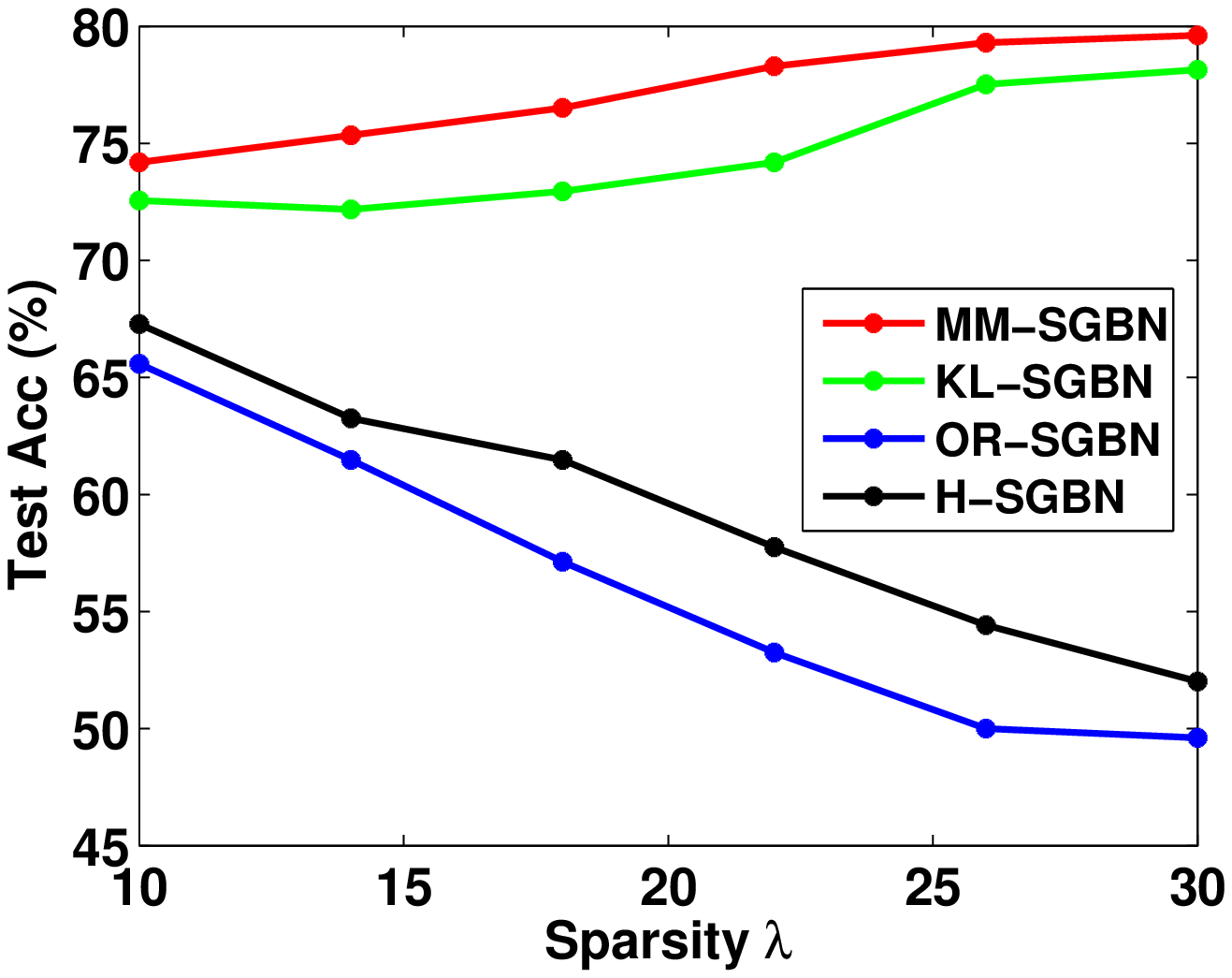}
&\includegraphics[width=0.33\textwidth]{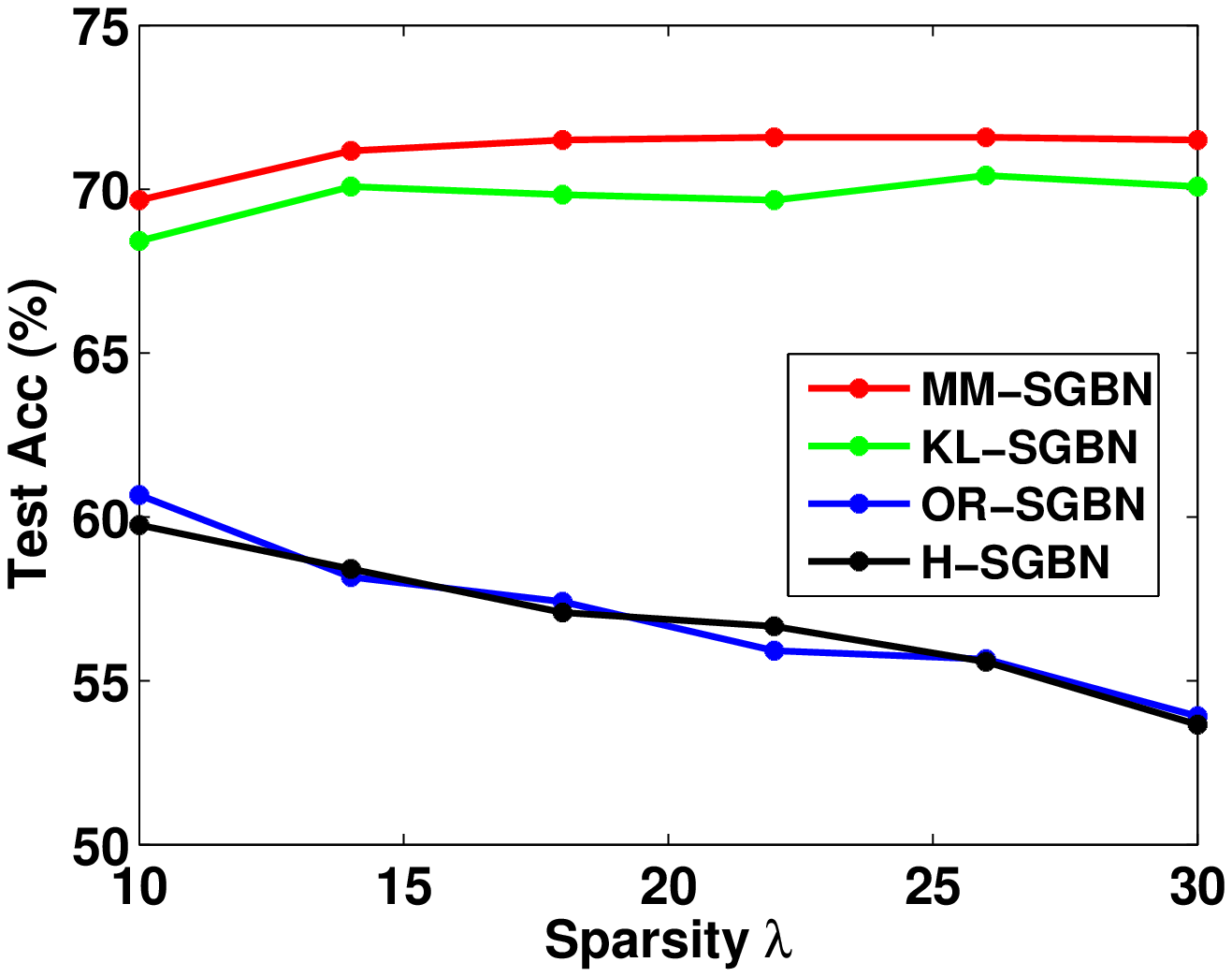}\\
\includegraphics[width=0.33\textwidth]{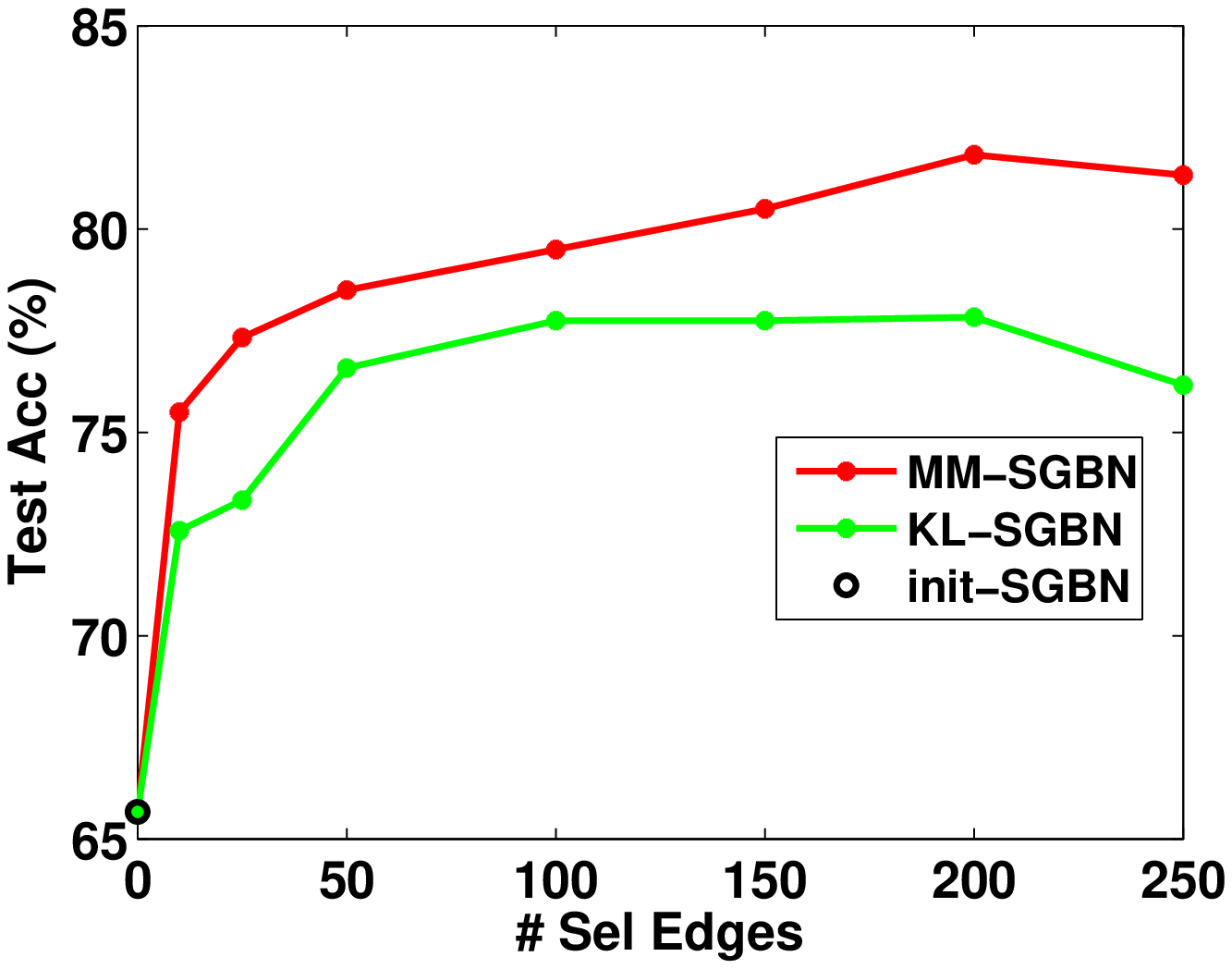}
&\includegraphics[width=0.33\textwidth]{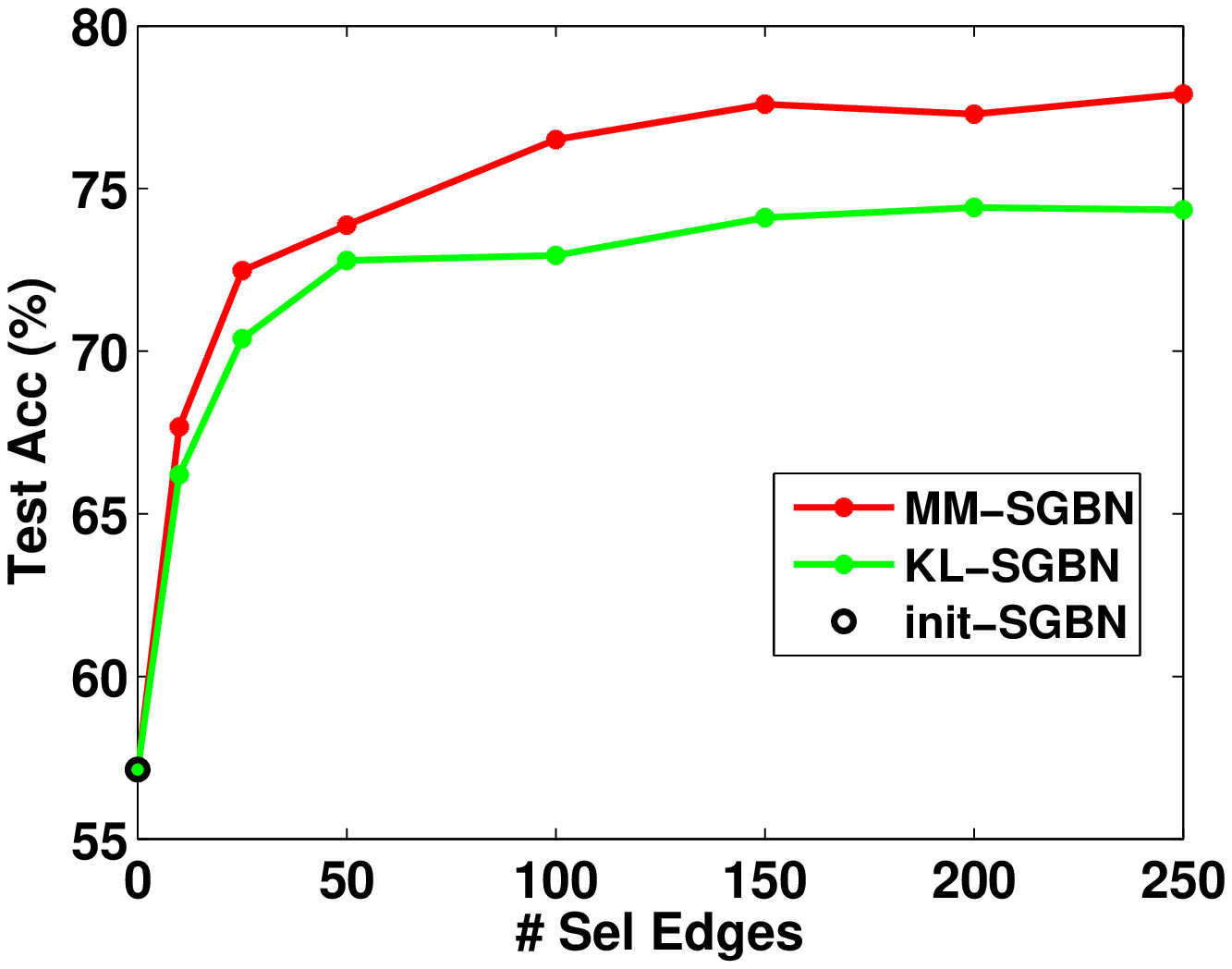}
&\includegraphics[width=0.33\textwidth]{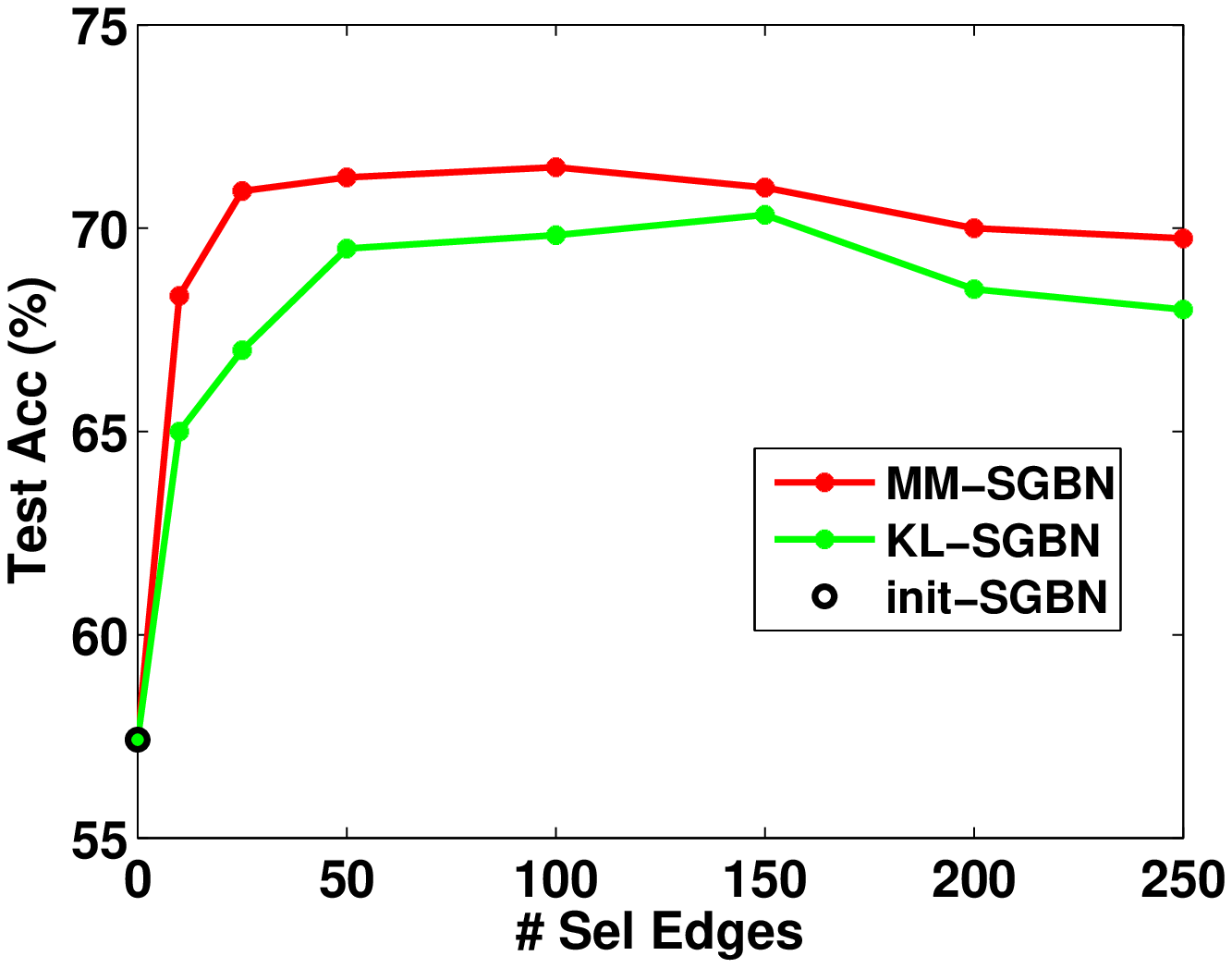}\\
\includegraphics[width=0.33\textwidth]{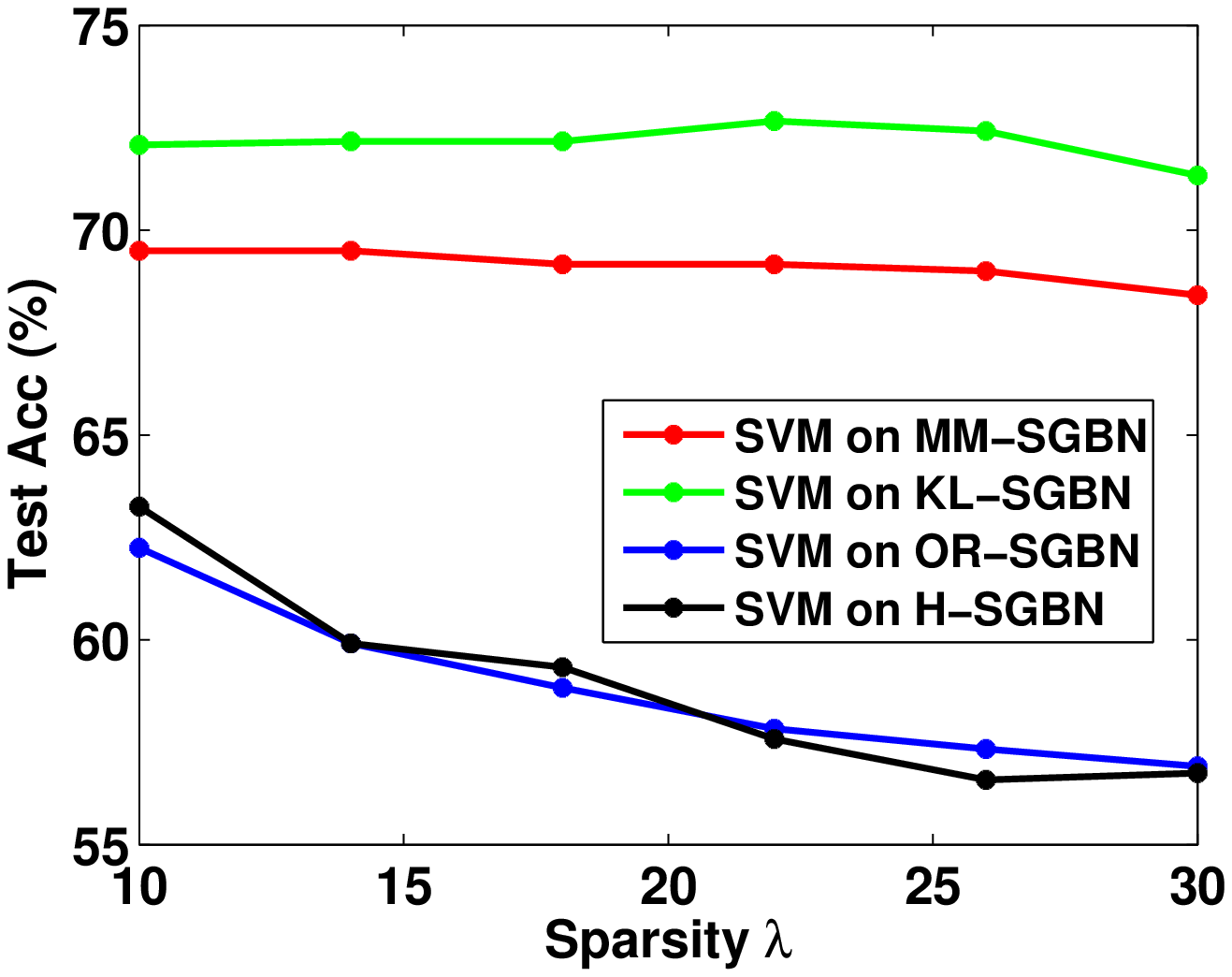}
&\includegraphics[width=0.33\textwidth]{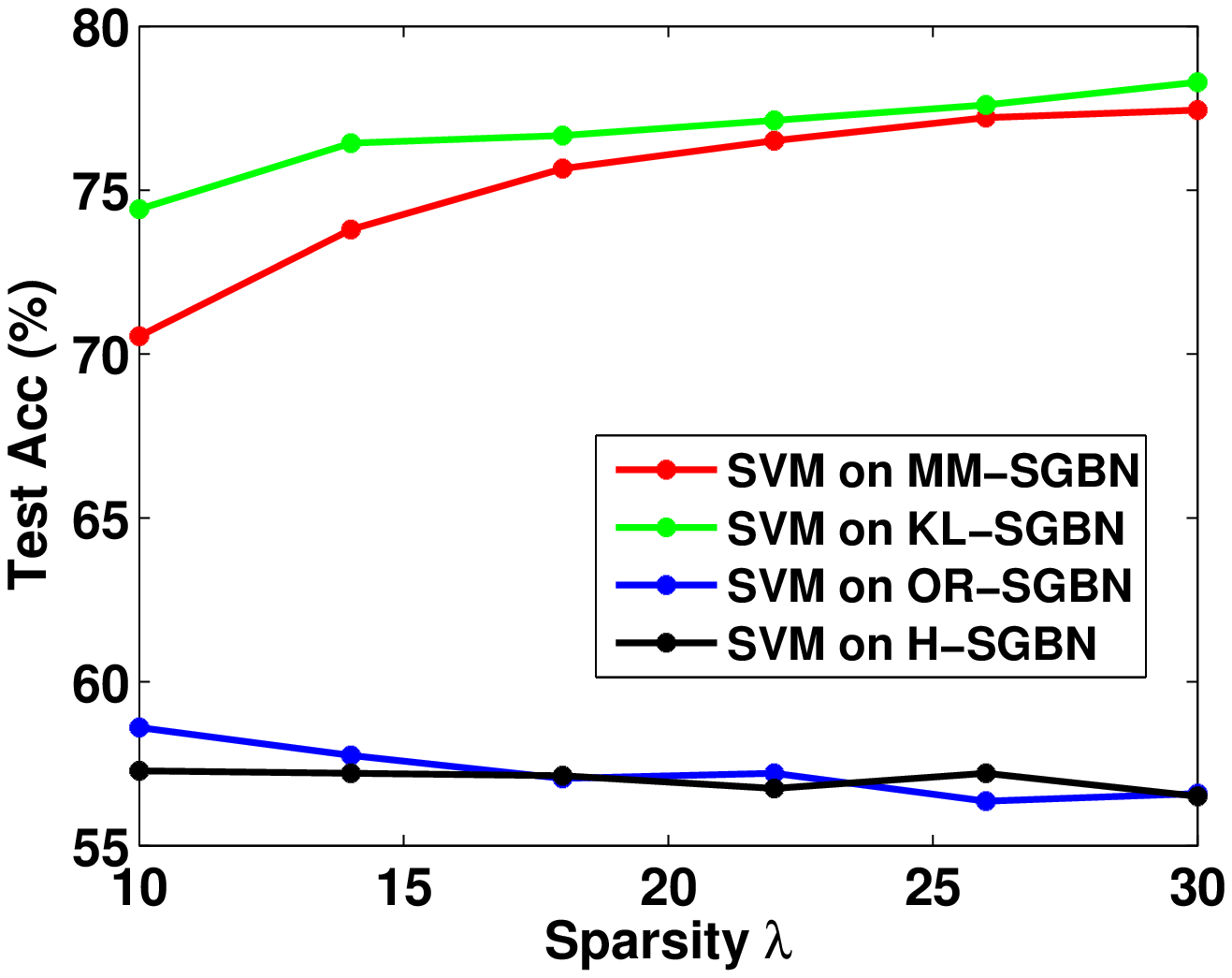}
&\includegraphics[width=0.33\textwidth]{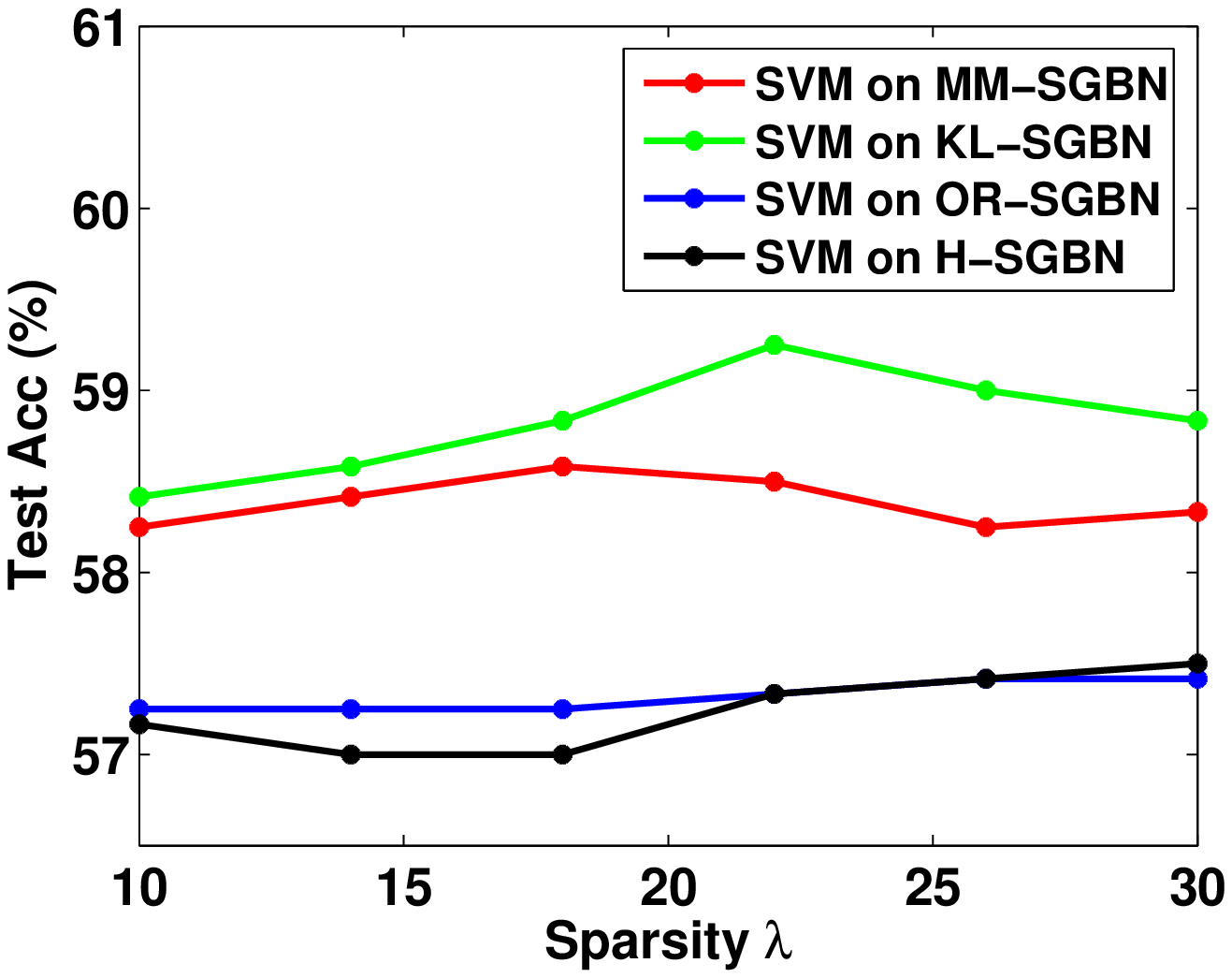}\\
\includegraphics[width=0.33\textwidth]{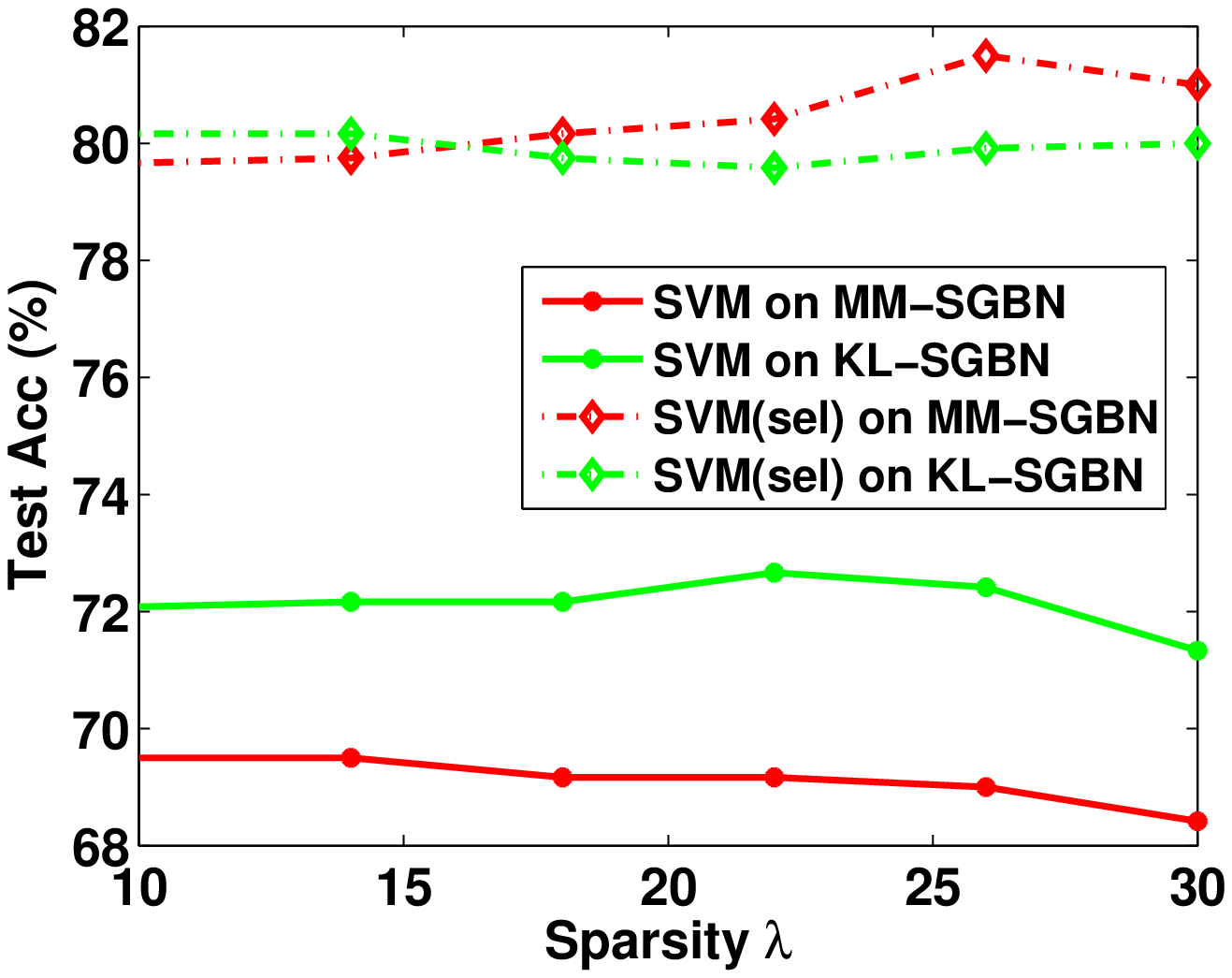}
&\includegraphics[width=0.33\textwidth]{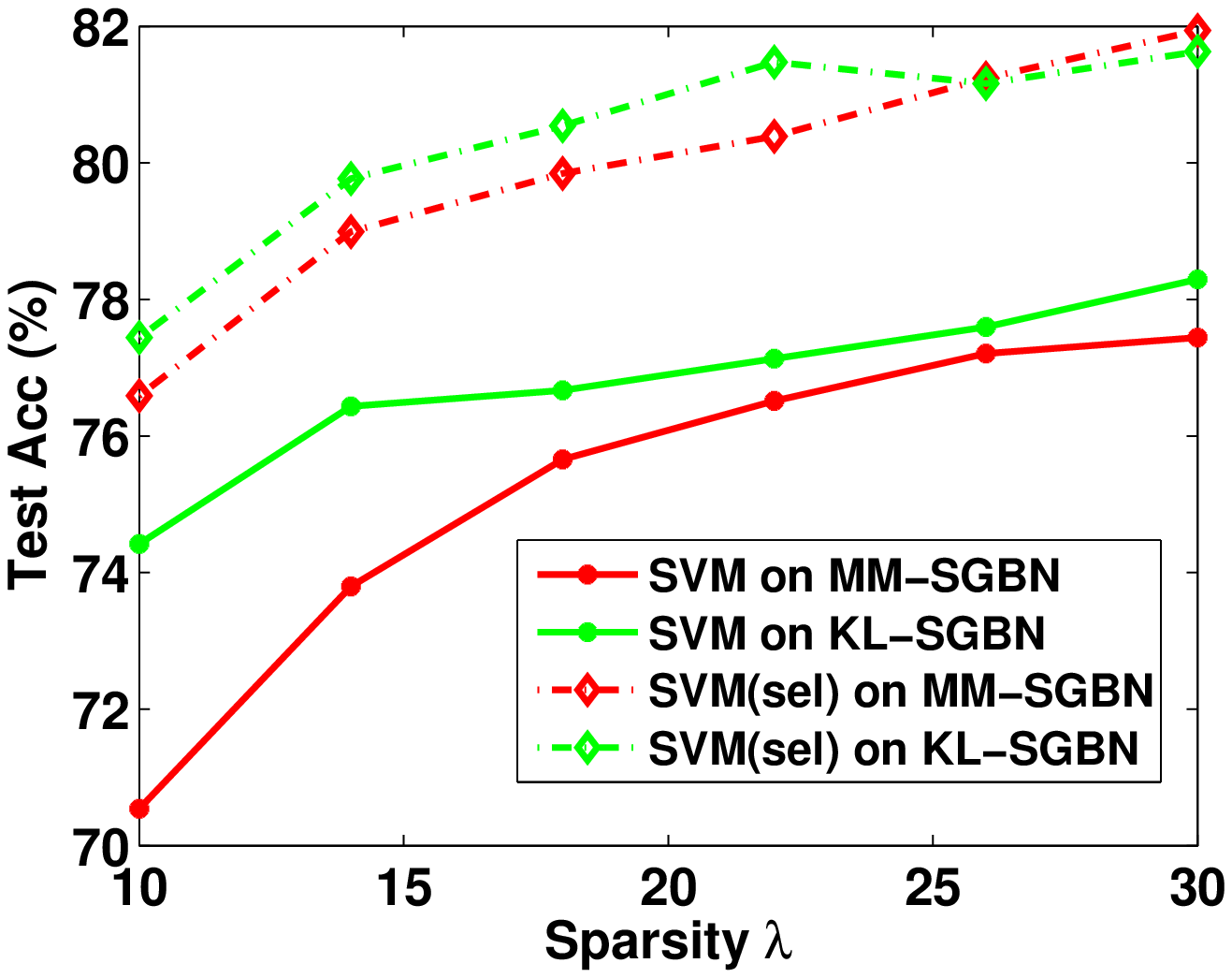}
&\includegraphics[width=0.33\textwidth]{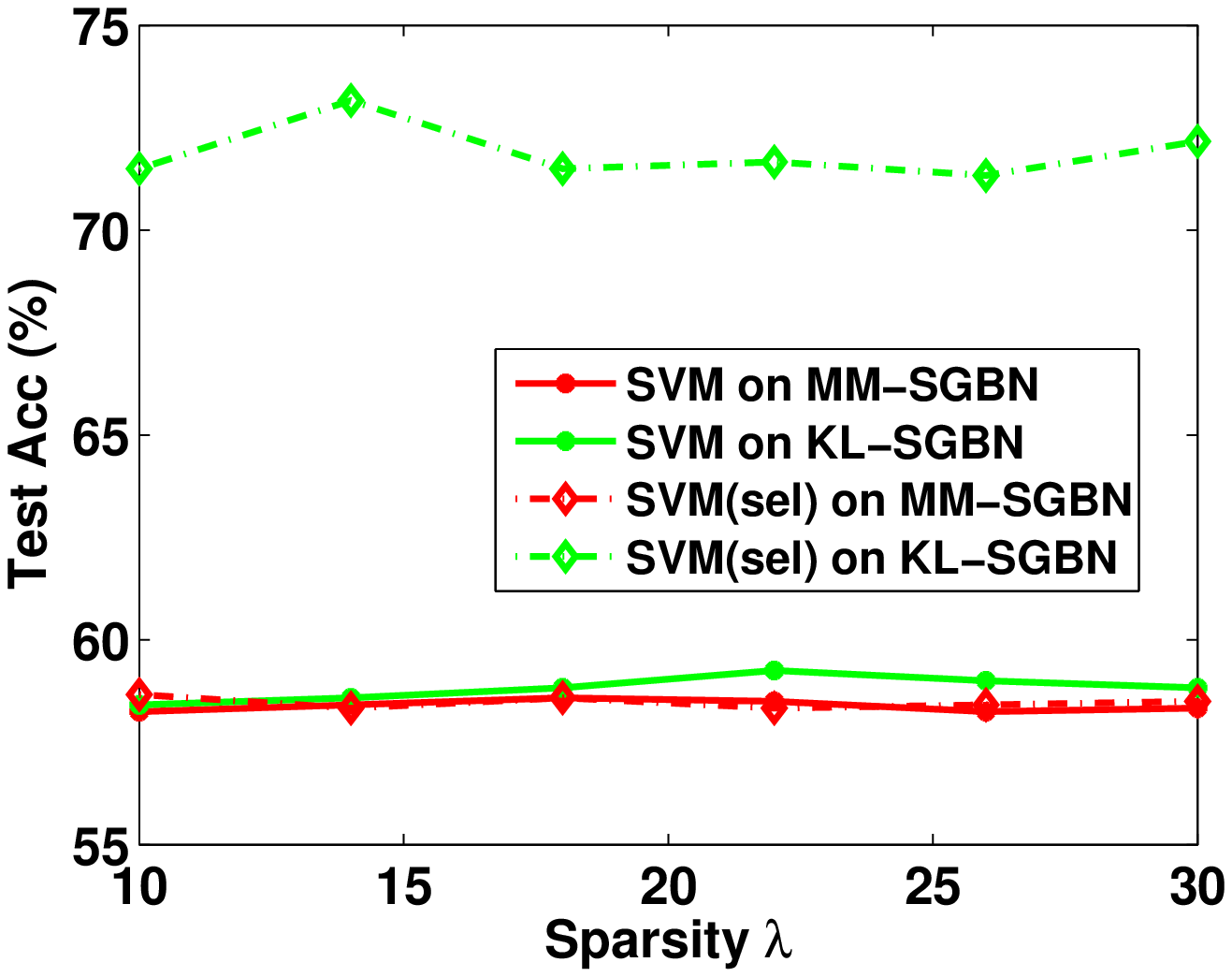}\\
MRI & PET & MRI-II\\
\end{tabular}
\caption{\em Comparison of classification accuracies on data sets of MRI (the left column), PET (the middle column) and MRI-II (the right column). The top two rows correspond to the test accuracies obtained by the learned SGBNs. The first row shows the test accuracies varied with the sparsity levels (i.e., the parameter $\lambda$). The  second row shows the test accuracies varied with the number of edges (denoted as ``\#Sel Edges" in the figure) optimized in discriminative learning. The bottom two rows correspond to the test accuracies obtained by SVMs using the SGBN-induced Fisher vectors either in full length (the third row) or with (100) selected components (the fourth row). }\label{fig:test_accs}
\end{center}
\vspace{-3mm}
\end{figure*}
\end{centering}

The test accuracies are averaged over the 30 randomly partitioned training-test groups. The classification performances of SGBN and SVM classifiers are evaluated with the varied parameter $\lambda$ that controls the sparsity level and the number of edges optimized in the learning process in Fig.~\ref{fig:test_accs}.  The results of our proposed KL-SGBN and MM-SGBN are plotted by the green and the red lines, respectively. The results of the individually learned OR-SGBN and H-SGBN are plotted by the blue and the black lines, respectively. 
The top two rows in Fig.~\ref{fig:test_accs} correspond to the results from the SGBN classifiers, while the bottom two rows correspond to those from the SVM classifiers. From Fig.~\ref{fig:test_accs}, we have the following observations.

i) Both KL-SGBN  and MM-SGBN can significantly improve the discriminative power of the individually learned SGBNs (Fig.~\ref{fig:test_accs}, the top two rows), as well as their associated SVM classifiers (Fig.~\ref{fig:test_accs}, the third row).  Such improvements are consistent over the three neuroimaging data sets and different parameter settings, and could reach the significant increases of $10\% \sim 20\%$ on most occasions. When the network becomes more sparse, the classification performance of the individually learned SGBNs (H-SGBN and OR-SGBN) drops significantly possibly due to the insufficient modeling of data.  However, under such circumstances,  KL-SGBN and MM-SGBN can still maintain high classification accuracies, which may indicate the necessity and effectiveness of the discriminative learning in classification scenarios.

ii) When using SGBN classifiers, for all the three data sets, MM-SGBN consistently achieves higher test accuracies at all sparsity levels (Fig.~\ref{fig:test_accs}, the first row) with different numbers of optimized edges than KL-SGBN (Fig.~\ref{fig:test_accs}, the second row). The advantage of MM-SGBN over KL-SGBN comes from explicitly optimizing the discriminative power of SGBNs, instead of getting help from optimizing the performance of SVM on SGBN-induced features.

iii) When using SVM classifiers, the SVM built upon KL-SGBN-induced features performs slightly better than that built upon MM-SGBN-induced features at all sparsity levels (Fig.~\ref{fig:test_accs}, the third row). This is expected since KL-SGBN optimizes the performance of its associated SVM classifier. 

iv) When cross-referencing the first and the third rows in Fig.~\ref{fig:test_accs}, it is noticed that SVM classifiers in general perform worse than the discriminatively learned SGBN classifiers. These may be because our Fisher vectors have very high dimensionality, which causes serious overfitting of data in SVM classifiers. Such situation might be somewhat improved for SGBN-classifiers since the simple Gaussian model may ``regularize" the model fitting. Based on this assumption, we further select a number of leading features from Fisher vectors by computing the Pearson correlation of the features and the labels, and use the selected features to construct the Fisher kernel for the SVM classifiers. As shown in the fourth row of Fig.~\ref{fig:test_accs}, the simple feature selection step can further significantly improve the classification performance of the Fisher-kernel based SVM.

v) The individually learned OR-SGBN and H-SGBN perform similarly for classification. However, as mentioned above, OR-SGBN has an additional advantage of being invariant to the feature ordering.

vi) Recall that these improvement on discrimination are achieved with no more than $1\%$ increase of squared fitting errors, which is explicitly controlled through the user defined parameters $T_1$ and $T_2$. Note that the rate of $1\%$ is application dependent. More tolerance of fitting errors can potentially bring more discrimination.

%

\subsection{Comparison of Connectivity}\label{subsec:compare_connectivity}
We also conduct an investigation to gain some insights into the learned brain networks for the diseased and the healthy populations, respectively.

\textbf{Exp-IV.}~~In this experiment, we visualize the learned brain networks and compare the connectivity patterns obtained by different methods and from different populations. MRI-II data set is used for this study since it covers regions spread over the four lobes of brain. 

\begin{centering}
\begin{figure}[!ht]
\begin{center}
\begin{tabular}{cc}
\hspace{-4mm}\includegraphics[width=0.23\textwidth]{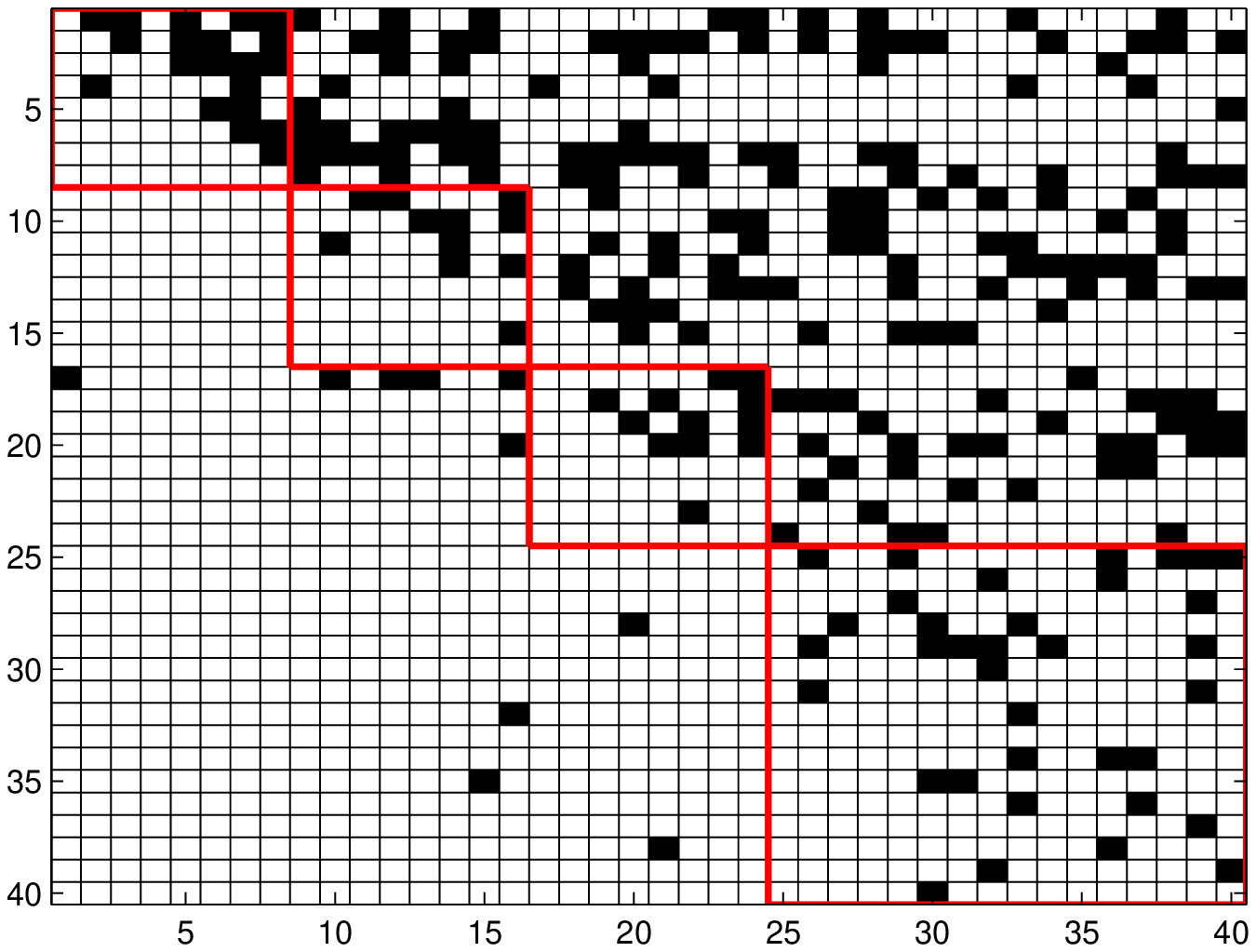}
&\hspace{-4mm}\includegraphics[width=0.23\textwidth]{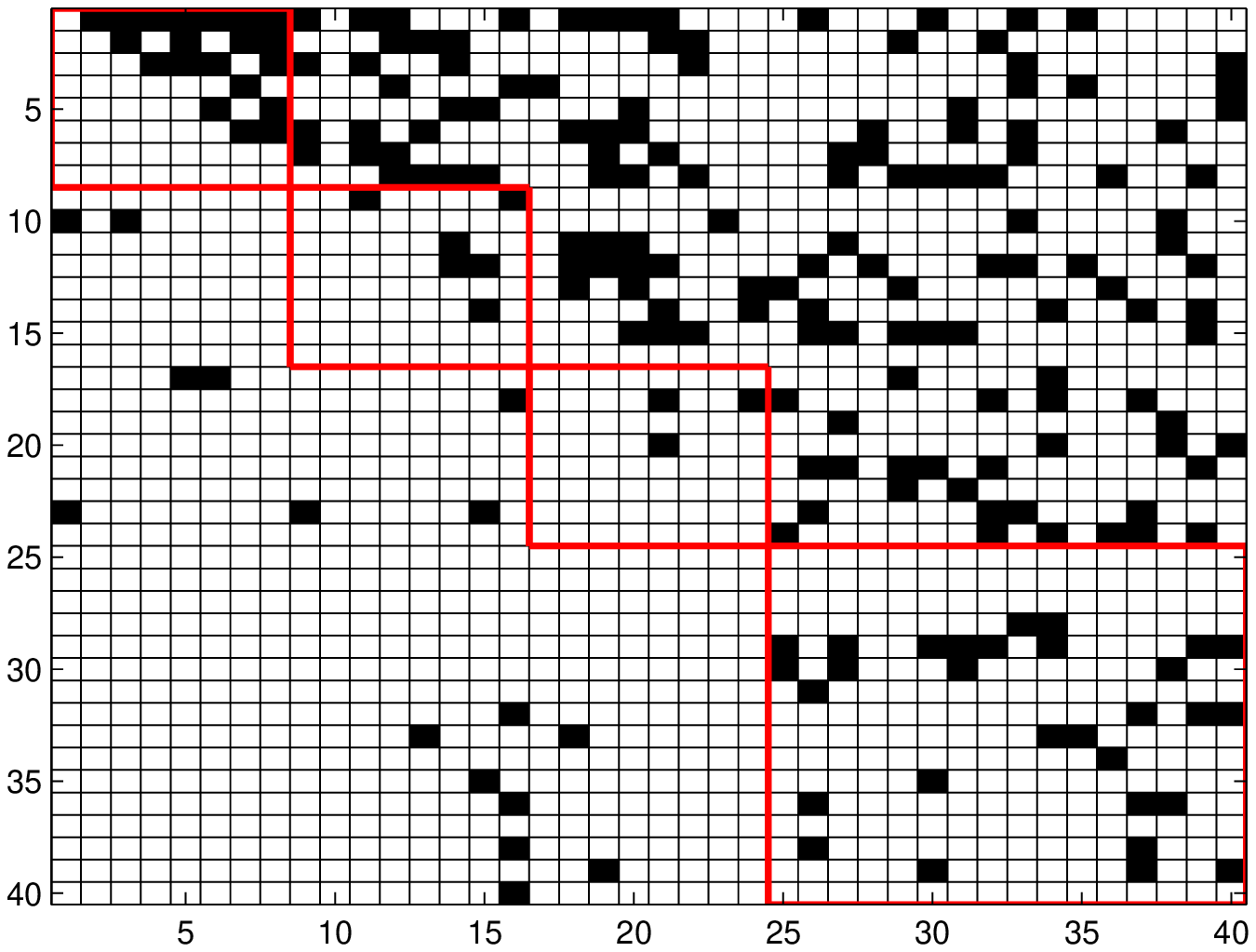}\\
H-SGBN (BCD): NC & H-SGBN (BCD): MCI\vspace{3mm}\\
\hspace{-4mm}\includegraphics[width=0.23\textwidth]{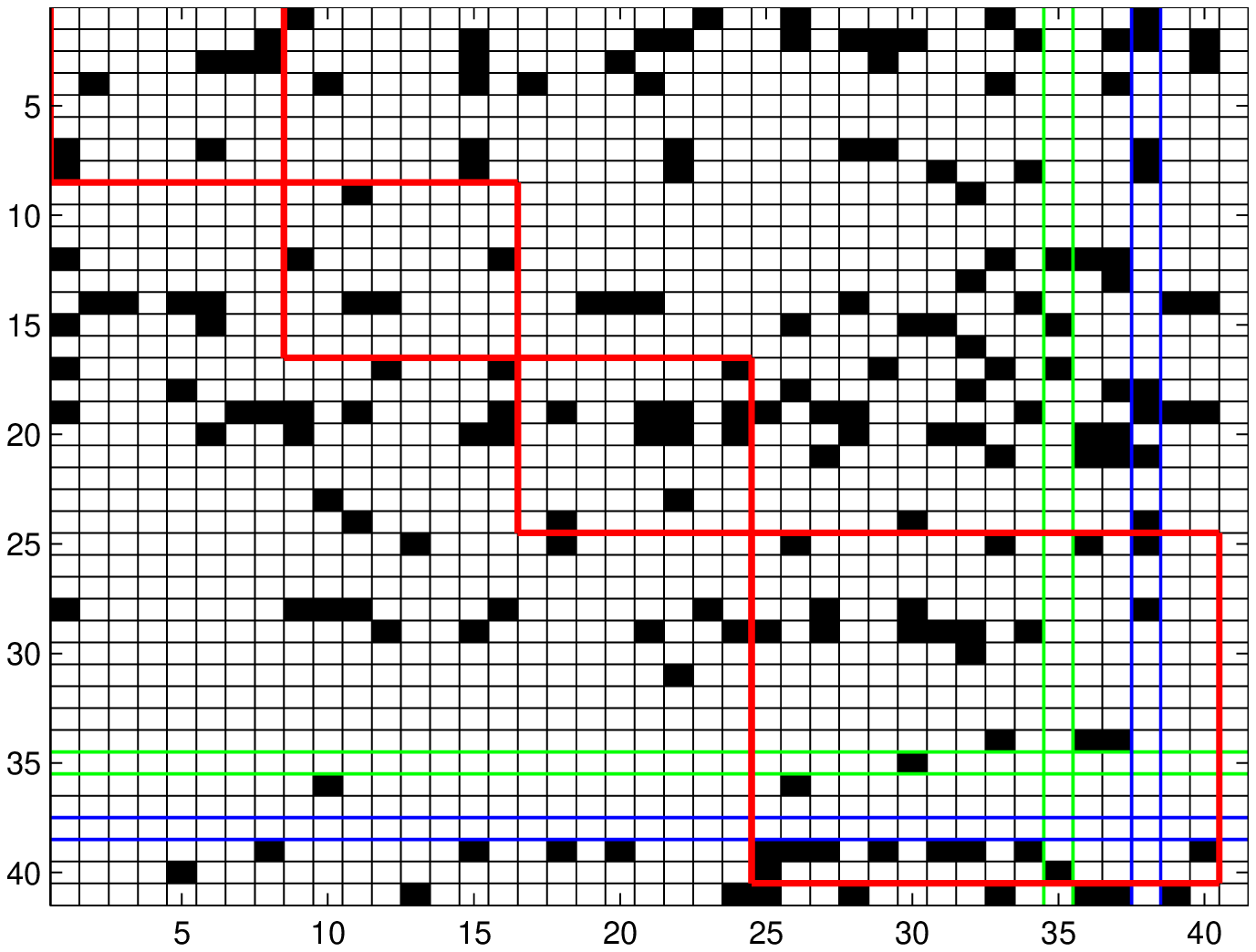}
&\hspace{-4mm}\includegraphics[width=0.23\textwidth]{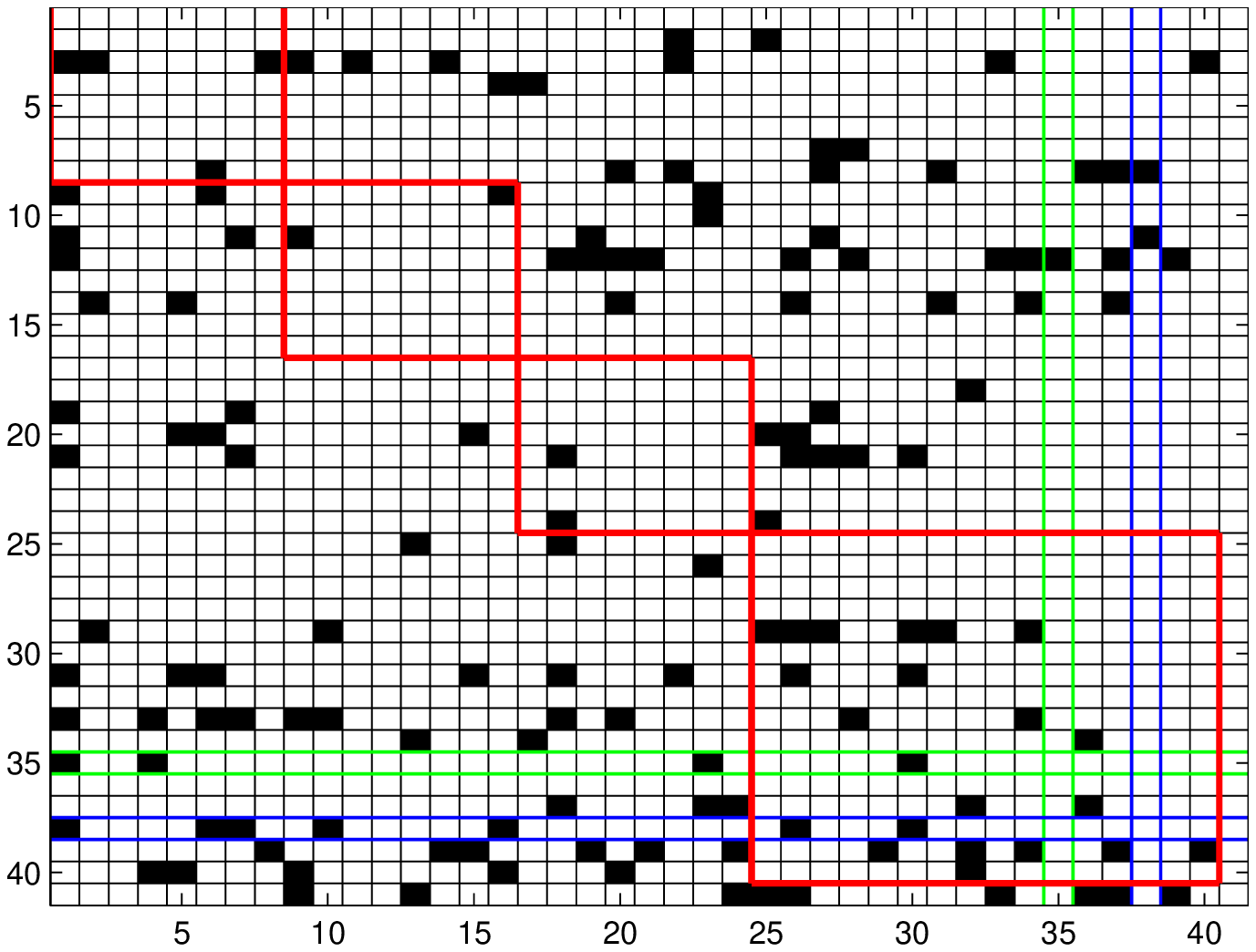}\\
OR-SGBN (WHOLE): NC & OR-SGBN (WHOLE): MCI\\
\end{tabular}
\caption{\em Visualization of connectivities for MRI-II. The four red boxes correspond to the frontal, parietal, occipital and temporal (including subcortical regions) lobes of the brain. The green row (Row 35) and column (Col 35) correspond to the left hippocampus while the blue ones (Row 38 and Col 38) correspond to the right hippocampus. }\label{fig:vis_connectivity}
\end{center}
\vspace{-3mm}
\end{figure}
\end{centering}

The structures of the brain networks recovered from NC and MCI groups are displayed in Fig.~\ref{fig:vis_connectivity} by using H-SGBN (BCD) and OR-SGBN (WHOLE), respectively. The network structure is obtained by thresholding the edge weights $\boldsymbol \Theta$  with a cutoff value of $0.01$~\cite{Huang-TPAMI-2012}. Each row $i$ represents the effective connections (dark dots) starting from the node $i$, and each column $j$ represents the effective connections ending at the node $j$. Note that, due to the different optimization problems involved, the same parameter $\lambda$ leads to different sparsity levels for H-SGBN and OR-SGBN. However, for a given method, different $\lambda$ values do not change the major structures of the resulting networks.  

In Fig.~\ref{fig:vis_connectivity}, it is noticed that H-SGBN (BCD) usually generates more connections in the upper triangle of the graphs even when we randomly permute the nodes. We suspect that this is caused by the column-wise optimization. The parameters ${{\boldsymbol \theta}_i}$ (corresponding to the columns in the graph) optimized at the early stage tend to be made more sparse than those optimized later in order to satisfy the DAG constraint. This phenomenon is not observed in OR-SGBN (WHOLE) that is used to initialize the discriminative learning.

Let us focus on OR-SGBN. Compared with H-SGBN, OR-SGBN has an additional bias node corresponding to the last row and column in Fig.~\ref{fig:vis_connectivity}. Visualizing ${\boldsymbol \Theta}$ can provide rich information for medical analysis. Here we just list a few observations as examples. With the same $\lambda$, OR-SGBN produces $183$ edges for NC, and $145$ edges for MCI. Such loss of connectivity also happens at the temporal lobe ($24\%$ loss) for MCI.  The temporal lobe (and some subcortical structures) is known to play a very important role in the progression of AD. The loss of connectivity in this region has been well-documented in wide AD-related studies~\cite{Supekar-PLOS-2008,Wang-HBM-2007,Huang-TPAMI-2012}. In Fig.~\ref{fig:vis_connectivity}, we also observe an increase of connectivity (the left bottom corner in the figure) between the frontal and the temporal lobe in MCI. Some study~\cite{Gould-Neurology-2006} mentioned that the frontal lobe may have connectivity increase at the early stage of AD as a compensation of cognitive functions for the patients. Moreover, significant directionality changes are also found for the left (node 35) and the right (node 38) hippocampi, an important structure among the earliest ones affected by AD. Both hippocampi have reduced incoming connectivity (communications dominated by other nodes) but increased outgoing connectivity (communications dominated by themselves) in MCI. Please note that the above observations could be influenced by the factors such as the limited number of data, the degree of disease progression and the imaging modality used in this study. More reliable medical analysis should be validated on larger data sets and worth further exploration, which is, however, beyond the scope of this paper.

\begin{centering}
\begin{figure}[!h!t]
\begin{center}
\begin{tabular}{cc}
\hspace{-4mm}\includegraphics[width=0.24\textwidth]{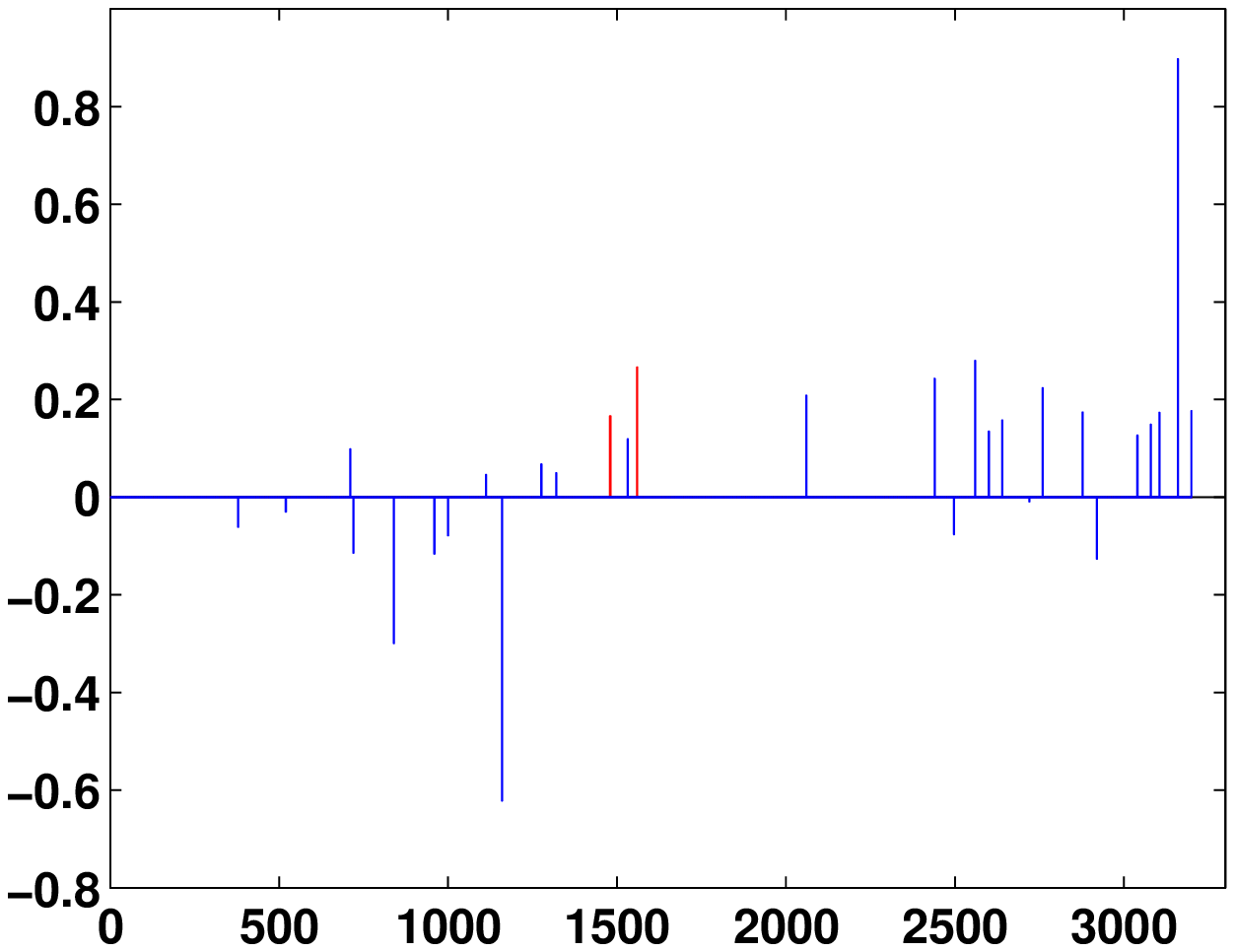}
&\hspace{-4mm}\includegraphics[width=0.24\textwidth]{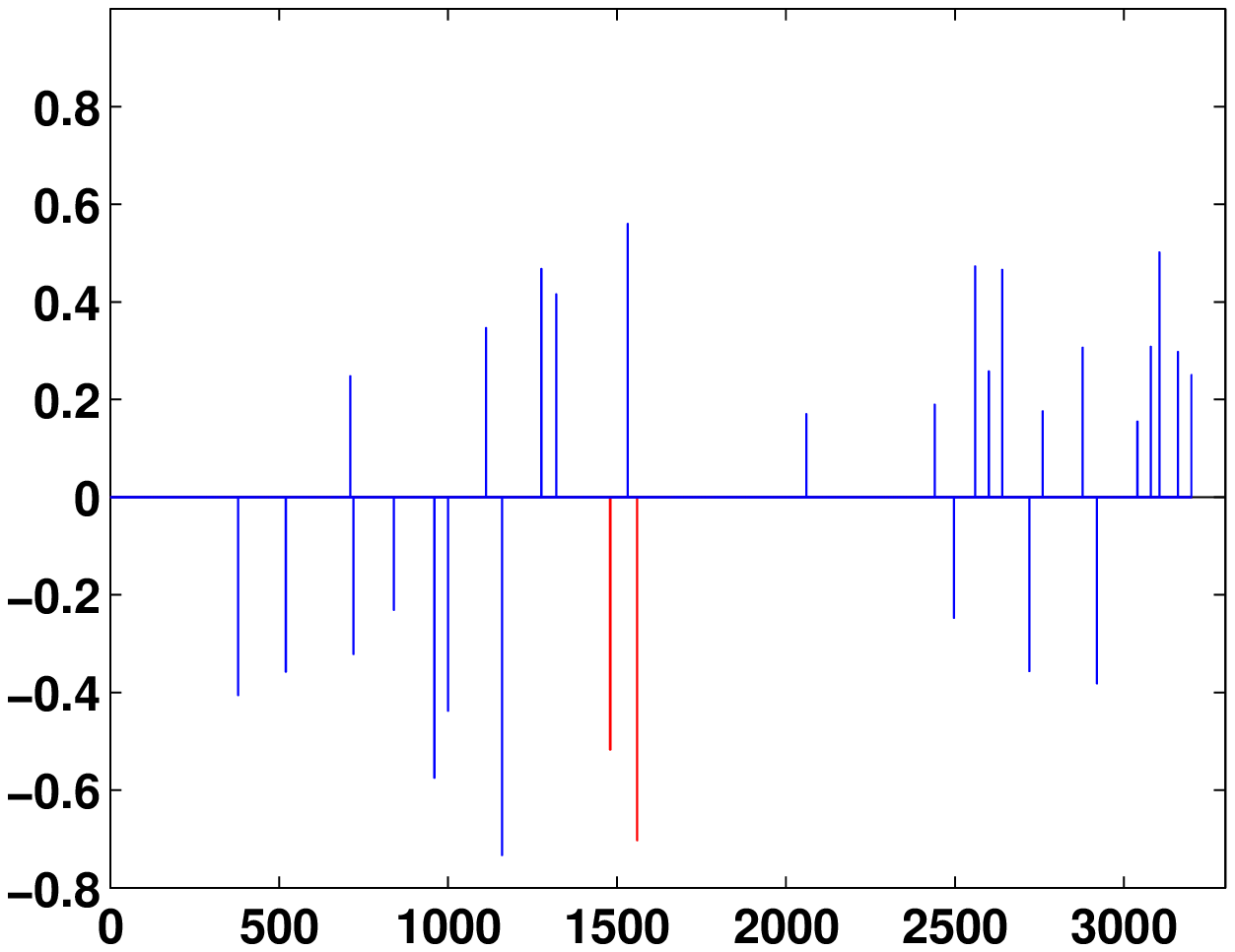}\\
(a) KL-SGBN& (b) MM-SGBN 
\end{tabular}
\caption{\em An example: change of edge weights learned by KL-SGBN and MM-SGBN}\label{fig:example}
\end{center}
\end{figure}
\end{centering}

To illustrate the difference of edge weights learned by KL-SGBN and MM-SGBN, an example of 30 edge weight changes (from the initial OR-SGBN) learned by these two methods is given in Fig.~\ref{fig:example}, where the SGBN networks from the two classes are vectorized and concatenated as $x$-axis. As shown, the signs of weight changes are quite similar in both methods. The most significant difference is that, MM-SGBN gives negative weight changes to the bias node of the left Amygdala and the right Parahippocampus (red lines in Fig.~\ref{fig:example}) while KL-SGBN gives them positive weight changes. The adjustment of edge weights leads to $10\%$ increase of test accuracy for MM-SGBN in this example.

\section{Conclusion}\label{sec:conclusion}
In this paper, we focus on the discriminative learning of Bayesian network for continuous variables, especially for neuroimaging data. Two discriminative learning frameworks are proposed to achieve this goal, i.e., KL-SGBN improves the performance of SVM classifiers based on SGBN-induced features, and MM-SGBN explicitly optimizes an SGBN-based criterion for classification. We demonstrate how to utilize Fisher-kernel to bridge the generative methods of SGBN and the discriminative classifiers of SVM, and how to embed the max-margin criterion into SGBN for discriminative learning. The optimization problems are analyzed in details, and the advantages and disadvantages of the proposed methods are discussed. Moreover, a new DAG constraint is proposed to ensure the validity of the graph with theoretical guarantee and validation on the benchmark data. We apply the proposed methods to modeling brain effective connectivity for early AD prediction. Significant improvements have been observed in the discriminative power of both the SGBN models and the associated SVM classifiers, without sacrificing much representation power.



\ifCLASSOPTIONcaptionsoff
  \newpage
\fi

\bibliographystyle{IEEEtran}
\bibliography{DL-SGBN}

\begin{thebibliography}{10}
\providecommand{\url}[1]{#1}
\csname url@rmstyle\endcsname
\providecommand{\newblock}{\relax}
\providecommand{\bibinfo}[2]{#2}
\providecommand\BIBentrySTDinterwordspacing{\spaceskip=0pt\relax}
\providecommand\BIBentryALTinterwordstretchfactor{4}
\providecommand\BIBentryALTinterwordspacing{\spaceskip=\fontdimen2\font plus
\BIBentryALTinterwordstretchfactor\fontdimen3\font minus
  \fontdimen4\font\relax}
\providecommand\BIBforeignlanguage[2]{{%
\expandafter\ifx\csname l@#1\endcsname\relax
\typeout{** WARNING: IEEEtran.bst: No hyphenation pattern has been}%
\typeout{** loaded for the language `#1'. Using the pattern for}%
\typeout{** the default language instead.}%
\else
\language=\csname l@#1\endcsname
\fi
#2}}

\bibitem{Verma-UAI-1991}
T.~Verma and J.~Pearl, ``Equivalence and synthesis of causal models,''
  \emph{Uncertainty in Artificial Intelligence}, vol.~6, pp. 255--268, 1991.

\bibitem{Spirtes-SSCR-1991}
P.~Spirtes and C.~Glymour, ``An algorithm for fast recovery of sparse causal
  graphs,'' \emph{Social Science Computer Review}, vol.~9, no.~1, pp. 62--72,
  1991.

\bibitem{Fast-thesis-2010}
A.~Fast, \emph{Learning the structure of Bayesian networks with constraint
  satisfaction.}\hskip 1em plus 0.5em minus 0.4em\relax Ph.D thesis, University
  of Massachusetts Amherst, 2010.

\bibitem{Scutari-arXiv-2014}
M.~Scutari, ``Bayesian network constraint-based structure learning algorithms:
  Parallel and optimised implementations in the bnlearn r package,''
  \emph{CoRR}, vol. abs/1406.7648, 2014.

\bibitem{Friedman-ML-2003}
N.~Friedman and D.~Koller, ``Being bayesian about network structure - bayesian
  approach to structure discovery in bayesian networks,'' \emph{Machine
  Learning}, vol.~50, no. 1-2, pp. 95--125, 2003.

\bibitem{Koivisto-JMLR-2004}
M.~Koivisto and K.~Sood, ``Exact bayesian structure discovery in bayesian
  networks,'' \emph{Journal of Machine Learning Research}, vol.~5, pp.
  549--573, 2004.

\bibitem{Geiger-arXiv-2013}
D.~Geiger and D.~Heckerman, ``Learning gaussian networks,'' \emph{CoRR}, vol.
  abs/1302.6808, 2013.

\bibitem{Suzuki-UAI-1993}
J.~Suzuki, ``A construction of bayesian networks from databases based on an mdl
  principle,'' in \emph{UAI}, 1993, pp. 266--273.

\bibitem{Acid-JAIR-2003}
S.~Acid and L.~Campos, ``Searching for bayesian network structures in the space
  of restricted acyclic partially directed graphs,'' \emph{Journal of
  Artificial Intelligence Research}, vol.~18, pp. 445--490, 2003.

\bibitem{Tsamardinos-ML-2006}
I.~Tsamardinos, L.~Brown, and C.~Aliferis, ``The max-min hill-climbing bayesian
  network structure learning algorithm,'' \emph{Machine Learning}, vol.~65,
  no.~1, pp. 31--78, 2006.

\bibitem{Jose-DMKD-2011}
J.~L.~M. Jos�e A~G�amez and J.~M. Puerta., ``Learning bayesian networks by
  hill climbing: Efficient methods based on progressive restriction of the
  neighborhood,'' \emph{Data Mining and Knowledge Discovery}, vol.~22, no. 1-2,
  pp. 106--148, 2011.

\bibitem{Schmidt-AAAI-2007}
M.~Schmidt, A.~Niculescu-Mizil, and K.~Murphy, ``Learning graphical model
  structures using l1-regularization paths,'' in \emph{AAAI}, 2007.

\bibitem{Pellet-JMLR-2008}
J.~Pellet and A.~Elisseeff, ``Using markov blankets for causal structure
  learning,'' \emph{JMLR}, vol.~9, pp. 1295--1342, 2008.

\bibitem{Huang-TPAMI-2012}
S.~Huang, J.~Li, J.~Ye, A.~Fleisher, K.~Chen, T.~Wu, and E.~Reiman, ``A sparse
  structure learning algorithm for gaussian bayesian network identification
  from high-dimensional data,'' \emph{IEEE TPAMI}, vol.~35, no.~6, pp.
  1328--1342, 2013.

\bibitem{Xiang-NIPS-2013}
J.~Xiang and S.~Kim, ``A* lasso for learning a sparse bayesian network
  structure for continuous variables,'' in \emph{NIPS}, 2013, pp. 2418--2426.

\bibitem{Pernkopf-JMLR-2010}
F.~Pernkopf and J.~Bilmes, ``Efficient heuristics for discriminative structure
  learning of bayesian network classifiers,'' \emph{JMLR}, vol.~11, pp.
  2323--2360, 2010.

\bibitem{Pernkopf-TPAMI-2012}
F.~Pernkopf, M.~Wohlmayr, and S.~Tschiatschek, ``Maximum margin bayesian
  network classifiers,'' \emph{IEEE TPAMI}, vol.~34, no.~3, pp. 521--532, 2012.

\bibitem{Guo-UAI-2005}
Y.~Guo, D.~Wilkinson, and D.~Schuurmans, ``Maximum margin bayesian networks,''
  in \emph{UAI}, 2005.

\bibitem{Jaakkola-NIPS-1998}
T.~Jaakkola and D.~Haussler, ``Exploiting generative models in discriminative
  classifiers,'' in \emph{NIPS}, 1998.

\bibitem{Bullmore-NRN-2009}
E.~Bullmore and O.~Sporns, ``Complex brain networks: graph theoretical analysis
  of structural and functional systems,'' \emph{Nat Rev Neurosci}, vol.~10,
  no.~3, pp. 186--198, 2009.

\bibitem{Smith-Neuroimage-2011}
S.~Smith, K.~Miller, G.~Khorshidi, M.~Webster, C.~Beckmann, T.~Nichols,
  J.~Ramsey, and M.~Woolrich, ``Network modelling methods for fmri,''
  \emph{Neuroimage}, vol.~54, no.~2, pp. 875--891, 2011.

\bibitem{RuiLi-PLoSOne-2013}
R.~Li, J.~Yu, S.~Zhang, F.~Bao, P.~Wang, X.~Huang, and J.~Li, ``Bayesian
  network analysis reveals alterations to default mode network connectivity in
  individuals at risk for alzheimer's disease,'' \emph{PLoS One}, vol.~8,
  no.~12, p. e82104, 2013.

\bibitem{Li-Neuroradio-2012}
R.~Li, X.~Wu, K.~Chen, A.~Fleisher, E.~Reiman, and L.~Yao, ``Alterations of
  directional connectivity among resting-state networks in alzheimer disease,''
  \emph{Am J Neuroradiol}, 2012.

\bibitem{Li-Neuroradiology-2011}
X.~Li, D.~Coyle, L.~Maguire, D.~Watson, and T.~McGinnity, ``Gray matter
  concentration and effective connectivity changes in alzheimer�s disease: A
  longitudinal structural mri study,'' \emph{Neuroradiology}, vol.~53, no.~10,
  pp. 733--748, 2011.

\bibitem{Zhou-CVPR-2013}
L.~Zhou, L.~Wang, L.~Liu, P.~Ogunbona, and D.~Shen, ``Discriminative brain
  effective connectivity analysis for alzheimers disease: A kernel learning
  approach upon sparse gaussian bayesian network,'' in \emph{CVPR}, 2013.

\bibitem{Zhou-MICCAI-2014}
------, ``Max-margin based learning for discriminative bayesian network from
  neuroimaging data,'' in \emph{MICCAI}, 2014.

\bibitem{Kim-HBM-2007}
J.~Kim, W.~Zhu, L.~Chang, P.~Bentler, and T.~Ernst, ``Unified structural
  equation modeling approach for the analysis of multisubject, multivariate
  functional mri data,'' \emph{Human Brain Mapping}, vol.~28, pp. 85--93, 2007.

\bibitem{Friston-Neuroimage-2003}
K.~Friston, L.~Harrison, and W.~Penney, ``Dynamic causal modeling,''
  \emph{Neuroimage}, vol.~19, pp. 1273--1302, 2003.

\bibitem{Perina-TPAMI-2012}
A.~Perina, M.~Cristani, U.~Castellani, V.~Murino, and N.~Jojic, ``Free energy
  score spaces: Using generative information in discriminative classifiers,''
  \emph{IEEE Transactions on Pattern Analysis and Machine Intelligence},
  vol.~34, no.~7, pp. 1249--1262, 2012.

\bibitem{Perronnin-cvpr-2007}
F.~Perronnin and C.~Dance, ``Fisher kernels on visual vocabularies for image
  categorization,'' in \emph{CVPR}, 2007.

\bibitem{Krapac-iccv-2011}
J.~Krapac, J.~Verbeek, and F.~Jurie, ``Modeling spatial layout with fisher
  vectors for image categorization,'' in \emph{ICCV}, 2011.

\bibitem{Sydorov-CVPR-2014}
V.~Sydorov, M.~Sakurada, and C.~Lampert, ``Deep fisher kernels - end to end
  learning of the fisher kernel gmm parameters,'' in \emph{CVPR}, 2014.

\bibitem{Maaten-ICML-2011}
L.~Maaten, ``Learning discriminative fisher kernels,'' in \emph{ICML}, 2011,
  pp. 217--224.

\bibitem{Chapelle-ML-2002}
O.~Chapelle, V.~Vapnik, O.~Bousquet, and S.~Mukherjee, ``Choosing multiple
  parameters for support vector machines,'' \emph{Machine Learning}, vol.~46,
  no. 1-3, pp. 131--159, 2002.

\bibitem{Wang-TPAMI-2008}
L.~Wang, ``Feature selection with kernel class separability,'' \emph{IEEE
  TPAMI}, vol.~30, no.~9, pp. 1534--1546, 2008.

\bibitem{Liu-TSMCB-2012}
X.~Liu, L.~Wang, J.~Yin, E.~Zhu, and J.~Zhang, ``An efficient approach to
  integrating radius information into multiple kernel learning,'' \emph{IEEE.
  TSMC-B}, 2012.

\bibitem{Bishop-2007}
C.~Bishop, \emph{Pattern Recognition and Machine Learning}.\hskip 1em plus
  0.5em minus 0.4em\relax Springer, 2007.

\bibitem{Pernkopf-ICML-2012}
R.~Peharz and F.~Pernkopf, ``Exact maximum margin structure learning of
  bayesian networks,'' in \emph{ICML}, 2012.

\bibitem{Zou-JASA-2006}
H.~Zou, ``The adaptive lasso and its oracle properties,'' \emph{Journal of the
  American Statistical Association}, vol. 101, no. 476, pp. 1418--1429, 2006.

\bibitem{Shojaie-Biometrika-2010}
A.~Shojaie and G.~Michailidis, ``Penalized likelihood methods for estimation of
  sparse high dimensional directed acyclic graphs,'' \emph{Biometrika},
  vol.~97, no.~3, pp. 519--538, 2010.

\bibitem{Fu-JASA-2013}
F.~Fu and Q.~Zhou, ``Learning sparse causal gaussian networks with experimental
  intervention: Regularization and coordinate descent,'' \emph{Journal of the
  American Statistical Association,}, vol. 108, no. 501, pp. 288--300, 2013.

\bibitem{ADNI}
ADNI, http://www.adni-info.org.

\bibitem{Kabani-Neuroimage-1998}
N.~Kabani, J.~MacDonald, C.~Holmes, and A.~Evans, ``A 3d atlas of the human
  brain,'' \emph{Neuroimage}, vol.~7, pp. S7--S17, 1998.

\bibitem{Tijms-cereb-2012}
B.~Tijms, P.~Seri�s, D.~Willshaw, and S.~Lawrie, ``Similarity-based
  extraction of individual networks from gray matter mri scans,'' \emph{Cereb
  Cortex}, vol.~22, no.~7, pp. 1530--1541, 2012.

\bibitem{BNR}
Http://www.cs.huji.ac.il/site/labs/compbio/Repository/.

\bibitem{Mackey-2003}
D.~Mackey, \emph{Information Theory, Inference, and Learning Algorithms}.\hskip
  1em plus 0.5em minus 0.4em\relax Cambridge University Press, 2003.

\bibitem{Margaritis-NIPS-1999}
D.~Margaritis and S.~Thrun, ``Bayesian network induction via local
  neighborhoods,'' in \emph{Proceedings of NIPS}, 1999.

\bibitem{Tsamardinos-IWAIS-2003}
I.~Tsamardinos and C.~Aliferis, ``Towards principled feature selection:
  Relevancy, filters and wrappers,'' in \emph{International Workshop on
  Artificial Intelligence and Statistics}, 2003.

\bibitem{Spirtes-1993}
P.~Spirtes, C.~Glymour, and R.~Scheines, \emph{Causation, Prediction, and
  Search}.\hskip 1em plus 0.5em minus 0.4em\relax Springer-Verlag, 1993.

\bibitem{Supekar-PLOS-2008}
K.~Supekar, V.~Menon, D.~Rubin, M.~Musen, and M.~Greicius, ``Network analysis
  of intrinsic functional brain connectivity in alzheimer's disease,''
  \emph{PLoS Computational Biology}, vol.~4, no.~6, pp. 1--11, 2008.

\bibitem{Wang-HBM-2007}
K.~Wang, M.~Liang, L.~Wang, L.~Tian, X.~Zhang, K.~Li, and T.~Jiang, ``Altered
  functional connectivity in early alzheimer's disease: a resting-state fmri
  study,'' \emph{Human Brain Mapping}, vol.~28, no.~10, pp. 967--978, 2007.

\bibitem{Gould-Neurology-2006}
R.~Gould, B.~Arroyo, R.~Brown, A.~Owen, E.~Bullmore, and R.~Howard, ``Brain
  mechanisms of successful compensation during learning in alzheimer disease,''
  \emph{Neurology}, vol.~67, no.~6, pp. 1011--1017, 2006.

\end{thebibliography}

%
%
%

\newpage
\onecolumn
\section*{Appendix}
\noindent \textbf{Proposition 1.} Given a sparse Gaussian Bayesian Network parameterized by ${\boldsymbol \Theta}$ and its associated directed graph $\mathcal G$ with $m$ nodes,  the graph $\mathcal G$ is DAG if and only if there exist some $o_i$ ($i=1,\cdots,m$) and ${\boldsymbol \Upsilon} \in {\mathbb R}^{m \times m}$, such that for arbitary $\Delta > 0$, the following constraints are satisfied:
\begin{align}
& o_j - o_i \geq \frac{\Delta}{m} - {\boldsymbol \Upsilon}_{ij},~~ \forall i, j \in \{1,\cdots,m\},~~i\neq j \tag{1a}\\
& {\boldsymbol \Upsilon}_{ij} \geq 0, \tag{1b} \\
&{\boldsymbol \Upsilon}_{ij}\times {\boldsymbol \Theta}_{ij} = 0, \tag{1c}\\
&\Delta \geq o_i \geq 0. \tag{1d}
\end{align}

\noindent \textbf{Proof.} As is known, a Bayesian network is equivalent to a topological ordering (Chapter 8, Section 8.1 on Page 362 in~\cite{Bishop-2007}). Therefore, we prove Proposition 1 by showing that i) Eqn.~(1a $\sim$ 1d) lead to a topological ordering (the necessary condition), and ii) a topological ordering from a DAG can meet the requirements in Eqn.~(1a $\sim$ 1d) (sufficient condition).

\begin{centering}
\begin{figure}[ht]
\begin{center}
\includegraphics[width=0.5\textwidth]{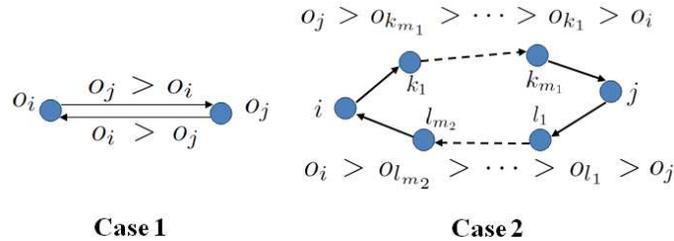}
\caption{\em Explanation of our ordering based DAG constraint.}\label{fig:DAG-Contradition}
\end{center}
\vspace{-2mm}
\end{figure}
\end{centering}

First, we prove the necessary condition by contradiction (Fig.~\ref{fig:DAG-Contradition}). We consider three cases for two nodes $j$ and $i$. Case 1) the nodes $j$ and $i$ are directly connected. If there is an edge from node $i$ to node $j$, the parameter ${\boldsymbol \Theta}_{ij}$ is then non-zero, and thus ${\boldsymbol \Upsilon}_{ij}$ must be zero. According to Eqn.~(1a), we have $o_j > o_i$. If at the same time, there is an edge from node $j$ to node $i$, similarly we have $o_i > o_j$, which contradicts with $o_j > o_i$, and therefore is impossible. Case 2: the nodes $j$ and $i$ are not directly linked but connected by a path. Suppose there is a directed path $P1$ from node $i$ to node $j$, where $P1$ is composed of nodes $k_1, k_2, \cdots, k_{m_1}$ in order. Following the above proof, we can have $o_j > o_{k_{m_1}}>\cdots>o_{k_1}>o_i$. If at the same time another directed path $P2$ links node $j$ to node $i$, where $P2$ is composed of nodes $l_1, l_2, \cdots, l_{m_2}$ in order, similarly we have $o_i > o_{l_{m_2}}>\cdots>o_{l_1}>o_j$, making the contradiction.
Case 3) If there is no edge between node $i$ and node $j$, by definition ${\boldsymbol \Theta}_{ij} = 0$. It is straightforward to see Eqn.~(1b) and Eqn.~(1c) hold for any arbitrary non-negative ${\boldsymbol \Upsilon}_{ij}$. Moreover, for any $o_i$ and $o_j$ satisfying Eqn.~(1d), we can show that as long as ${\boldsymbol \Upsilon}_{ij} \geq (\frac{1}{m}+1)\Delta$ (which is positive), Eqn.~(1a) will always hold. This is further explained as follows. By Eqn.~(1d), we have $-\Delta \leq o_j - o_i \leq \Delta$. For Eqn.~(1a) to be always held, we need some ${\boldsymbol \Upsilon}_{ij}$ such that $o_j - o_i \geq -\Delta \geq \frac{\Delta}{m} - {\boldsymbol \Upsilon}_{ij}$, which requires ${\boldsymbol \Upsilon}_{ij} \geq (\frac{1}{m}+1)\Delta$. Therefore, there exist a set of $o_i$ and ${\boldsymbol \Upsilon}$ valid for Eqn.~(1a $\sim$ 1d) when no edge links node $i$ and node $j$. 
In sum, Eqn.~(1a $\sim$ 1d) show a topological ordering, that is, if node $j$ comes after node $i$ (that is, $o_j > o_i$) in the ordering, there can not be a link from node $j$ to node $i$, which guarantees the acyclicity.

Now let us consider the sufficient condition. if ${\mathcal G}$ is a DAG, we can obtain some topological ordering $(1, 2, \cdots, m)$ from it. Let $\tilde{o}_i$ be the index of node $i$ in this ordering. Setting $o_i = (\tilde{o}_i - 1)\frac{\Delta}{m}$  ($\forall i \in \{1,\cdots,m\}$), we have $\min(o_i) = (1-1)\frac{\Delta}{m}=0$ and $\max(o_i)=(m-1)\frac{\Delta}{m} \leq \Delta$. If node $j$ comes after node $i$, we have $o_j - o_i \geq \frac{\Delta}{m} \geq \frac{\Delta}{m} - {\boldsymbol \Upsilon}_{ij}$. If node $j$ comes before node $i$, we can always set ${\boldsymbol \Upsilon}_{ij}$ sufficiently large to satisfy   Eqn.~(1a $\sim$ 1d). Therefore, from a DAG, we can always construct a set of ordering variables that satisfy Eqn.~(1a $\sim$ 1d).

Combining the proofs above, Eqn.~(1a $\sim$ 1d) are the sufficient and necessary condition for a directed graph ${\mathcal G}$ to be DAG. \hfill$\square$
\\
\\
\\

\noindent \textbf{Proposition 2.} The optimization problem in Eqn.~(2) (i.e., Eqn.~(4.2) in the paper) is iteratively solved by alternate optimizations of  (i) ${\mathbf o}$ and ${\boldsymbol \Upsilon}$ with ${\boldsymbol \Theta}$ fixed, and (ii) ${\boldsymbol \Theta}$ with ${\mathbf o}$ and ${\boldsymbol \Upsilon}$ fixed. This optimization converges and the output ${\boldsymbol \Theta}^{\star}$ is DAG when $\lambda_{dag} > \frac{2m(m-2)(n-1)^2 + m\lambda_1(2n-2-\lambda_1)}{\lambda_1(1+m)\Delta}$, where $m$ is the number of nodes and $n$ is the number of samples. 
\begin{align}
&\underset{{\boldsymbol \Theta}, {\mathbf o}, {\boldsymbol \Upsilon}}\min\sum_{i=1}^m\|{\mathbf x}_{:, i} - {\mathbf {PA}}_i^{\top}{\boldsymbol \theta}_i\|_2^2+\lambda_1\|{\boldsymbol \theta}_i\|_1+\lambda_{dag}{{\boldsymbol \epsilon}_i}^{\top}|{\boldsymbol \theta}_i| \tag{2}\\
&s.t. ~~  o_j - o_i \geq \frac{\Delta}{m} - {\boldsymbol \Upsilon}_{ij},\forall i, j \in \{1,\cdots,m\},~~i\neq j\nonumber\\
&~~~~~~~0 \leq o_i \leq \Delta, ~~ {\boldsymbol \Upsilon}_{ij} \geq 0 \nonumber
\end{align}
Here ${\mathbf o}$ and ${\boldsymbol \Upsilon}$ are the variables defined in the DAG constraint in Section 4.2, and ${\boldsymbol \Theta}$ is the model parameters of SGBN. The vector ${\boldsymbol \epsilon}_i$ denotes the $i$-th column of the matrix ${\boldsymbol \Upsilon}$, and $|{\boldsymbol \theta}_i|$ the component-wise absolute value of the $i$-th column of ${\boldsymbol \Theta}$. Other parameters are defined in Table 1 in the paper.\\

\noindent \textbf{Proof.}  In the following, we prove that:
\begin{enumerate}
\item The alternate optimization in Eqn.~(2)  converges.
\item The solution ${\boldsymbol \Theta}^{\star}$ of Eqn.~(2) is DAG when $\lambda_{dag}$ is sufficiently large.
\end{enumerate}
\vspace{5mm}

Let us denote $f({\boldsymbol \Theta}, {\mathbf o}, {\boldsymbol \Upsilon}) = \sum_{i=1}^m\|{\mathbf x}_{:, i} - {\mathbf {PA}}_i^{\top}{\boldsymbol \theta}_i\|_2^2+\lambda_1\|{\boldsymbol \theta}_i\|_1+\lambda_{dag}{{\boldsymbol \epsilon}_i}^{\top}|{\boldsymbol \theta}_i|$. \\

\textbf{First}, we prove Eqn.~(2)  converges by showing that
(i) $f({\boldsymbol \Theta}, {\mathbf o}, {\boldsymbol \Upsilon})$  is lower bounded; and 
(ii) $f({\boldsymbol \Theta}^{(t+1)}, {\mathbf o}^{(t+1)}, {\boldsymbol \Upsilon}^{(t+1)}) \leq f({\boldsymbol \Theta}^{(t)}, {\mathbf o}^{(t)}, {\boldsymbol \Upsilon}^{(t)})$, meaning that the function value will monotonically decrease with the iteration number $t$. \\

It is easy to see that $f({\boldsymbol \Theta}, {\mathbf o}, {\boldsymbol \Upsilon})$  is lower bounded by $0$, since each term in $f({\boldsymbol \Theta}, {\mathbf o}, {\boldsymbol \Upsilon})$ is non-negative. And the second point can be proven as follows. At the $t$-th iteration, we update ${\boldsymbol \Theta}$ by

\begin{align}
{\boldsymbol \Theta}^{(t+1)} &= \underset{\boldsymbol \Theta}{\arg} \min\sum_{i=1}^m\|{\mathbf x}_{:, i} - {\mathbf {PA}}_i^{\top}{\boldsymbol \theta}_i\|_2^2+\lambda_1\|{\boldsymbol \theta}_i\|_1+\lambda_{dag}{{\boldsymbol \epsilon}_i}^{(t) \top}|{\boldsymbol \theta}_i| \tag{3} \\
& =\underset{\boldsymbol \Theta}{\arg} \min f({\boldsymbol \Theta}, {\mathbf o}^{(t)}, {\boldsymbol \Upsilon}^{(t)}). \nonumber
\end{align}

It holds that $f({\boldsymbol \Theta}^{(t+1)}, {\mathbf o}^{(t)}, {\boldsymbol \Upsilon}^{(t)}) \leq f({\boldsymbol \Theta}^{(t)}, {\mathbf o}^{(t)}, {\boldsymbol \Upsilon}^{(t)})$. 
Also it is noted that  ${\boldsymbol \Theta}^{(t+1)}$ is an achievable global minimum of ${\boldsymbol \Theta}$ since $f({\boldsymbol \Theta}, {\mathbf o}^{(t)}, {\boldsymbol \Upsilon}^{(t)})$ is a convex function with respect to ${\boldsymbol \Theta}$. Similarly, we then update ${\mathbf o}$ and ${\boldsymbol \Upsilon}$ by
\begin{align}
\{{\mathbf o}^{(t+1)}, {\boldsymbol \Upsilon}^{(t+1)} \}
=&{\arg}~\underset{{\mathbf o}, {\boldsymbol \Upsilon}}\min f({\boldsymbol \Theta}^{(t+1)}, {\mathbf o}, {\boldsymbol \Upsilon})\tag{4}\\
&s.t. ~~  o_j - o_i \geq \frac{\Delta}{m} - {\boldsymbol \Upsilon}_{ij},\forall i, j \in \{1,\cdots,m\},~~i\neq j\nonumber\\
&~~~~~~~0 \leq o_i \leq \Delta, ~~ {\boldsymbol \Upsilon}_{ij} \geq 0. \nonumber
\end{align}
It holds that $f({\boldsymbol \Theta}^{(t+1)}, {\mathbf o}^{(t+1)}, {\boldsymbol \Upsilon}^{(t+1)}) \leq f({\boldsymbol \Theta}^{(t+1)}, {\mathbf o}^{(t)}, {\boldsymbol \Upsilon}^{(t)})$. Also, $f({\boldsymbol \Theta}^{(t+1)}, {\mathbf o}, {\boldsymbol \Upsilon})$ is a linear function with respect to ${\mathbf o}$ and ${\boldsymbol \Upsilon}$. Consequently we have 
\[
f({\boldsymbol \Theta}^{(t+1)}, {\mathbf o}^{(t+1)}, {\boldsymbol \Upsilon}^{(t+1)}) \leq f({\boldsymbol \Theta}^{(t+1)}, {\mathbf o}^{(t)}, {\boldsymbol \Upsilon}^{(t)}) \leq f({\boldsymbol \Theta}^{(t)}, {\mathbf o}^{(t)}, {\boldsymbol \Upsilon}^{(t)}).
\]
Therefore, the optimization problem in Eqn.~(2) is guaranteed to converge with the alternate optimization strategy, because the objective function is lower-bounded and monotonically decreases with the iteration numbers.\\

\textbf{Second}, we prove that when $\lambda_{dag} > \frac{2m(m-2)(n-1)^2 + m\lambda_1(2n-2-\lambda_1)}{\lambda_1(1+m)\Delta}$, the output ${\boldsymbol \Theta}^{\star}$ is guaranteed to be DAG. This could be proven by contradiction. Suppose that such a $\lambda_{dag}$ does not lead to a DAG, say, ${\boldsymbol \Upsilon}_{ji} \times {\boldsymbol \Theta}_{ji}^{\star} \neq 0$ for at least one pair of nodes $i$ and $j$, with ${\boldsymbol \Theta}_{ji}^{\star} \neq 0$ and ${\boldsymbol \Upsilon}_{ji} > 0$. Without loss of generality, we assume ${\boldsymbol \Upsilon}_{ji} \geq (\frac{1}{m}+1)\Delta$ (where $\Delta$ is an arbitary positive number), so the ordering constraints in Eqn.~(2) always hold regardless of the variables ${o}_i$ and ${o}_j$. This is because $o_i$ and $o_j$ are constrained by $0 \leq o_i \leq \Delta$ and $0 \leq o_j \leq \Delta$, and $o_j - o_i \geq -{\Delta}=\frac{1}{m}{\Delta} - (\frac{1}{m}+1)\Delta$. Based on the first-order optimality condition, ${\boldsymbol \Theta}_{ji}^{\star} \neq 0$ i.f.f. $2\left|\left({\mathbf x}_{:, i} - {\mathbf {PA}}_{i (\setminus j, :)}^{\top}{\boldsymbol \theta}_{i \setminus j}^{\star}\right)^{\top}{\mathbf x}_{:, j}\right| - (\lambda_1+\lambda_{dag}{\boldsymbol \Upsilon}_{ij}) > 0$. Here,  ${\mathbf {PA}}_{i (\setminus j, :)}$ denotes the elements in the matrix ${\mathbf {PA}}_{i}$ with the $j$-th row removed (i.e., parents of the node $i$ without the node $j$), and ${\boldsymbol \theta}_{i \setminus j}^{\star}$ denotes the elements in ${\boldsymbol \theta}_{i}^{\star}$ without ${\boldsymbol \Theta}_{ji}^{\star}$. However, it can be shown that,
\begin{align}
\left|\left({\mathbf x}_{:, i} - {\mathbf {PA}}_{i (\setminus j, :)}^{\top}{\boldsymbol \theta}_{i \setminus j}^{\star}\right)^{\top}{\mathbf x}_{:, j}\right|
&\leq \left|{\mathbf x}_{:, i}^{\top}{\mathbf x}_{:, j}\right| + \left|{\boldsymbol \theta}_{i \setminus j}^{\star~\top}{\mathbf {PA}}_{i (\setminus j, :)}{\mathbf x}_{:, j}\right|\tag{5}\\
&=\left|{\mathbf x}_{:, i}^{\top}{\mathbf x}_{:, j}\right| + \sum_{k=1, k\neq i, j}^{m} \left|{\boldsymbol \Theta}_{ki}^{\star}{\mathbf x}_{:, k}^{\top}{\mathbf x}_{:, j}\right|\nonumber\\
&\leq (n-1) + (m-2)(n-1)\max|{\boldsymbol \Theta}_{ki}^{\star}|\nonumber\\
&\leq (n-1) + \frac{(m-2)(n-1)^2}{\lambda_1}.\nonumber
\end{align}
The second last inequality holds  due to the normalization of features ${\mathbf x}_{:, i}$ (to zero mean and unit std). The last inequality holds because $\max|{\boldsymbol \Theta}_{ki}^{\star}| \leq \|{\boldsymbol \theta}_i^{\star}\|_1 \leq \frac{1}{\lambda_1}\left(\|{\mathbf x}_{:, i} - {\mathbf {PA}}_i^{\top}{\boldsymbol \theta}_i^{\star}\|_2^2+\lambda_1\|{\boldsymbol \theta}_i^{\star}\|_1+\lambda_{dag}{{\boldsymbol \epsilon}_i}^{\star\top}|{\boldsymbol \theta}_i^{\star}|\right) = \frac{1}{\lambda_1}f({\boldsymbol \Theta}^{\star}, {\mathbf o}^{\star}, {\boldsymbol \Upsilon}^{\star}) \leq \frac{1}{\lambda_1}f({\mathbf 0}, {\mathbf o}^{\star}, {\boldsymbol \Upsilon}^{\star})
 = \frac{1}{\lambda_1}{\mathbf x}_{:, i}^{\top}{\mathbf x}_{:, i} = \frac{n-1}{\lambda_1}$.  With the given $\lambda_{dag}$, Eqn.~(5) results in
\[
2\left|\left({\mathbf x}_{:, i} - {\mathbf {PA}}_{i (\setminus j, :)}^{\top}{\boldsymbol \theta}_{i \setminus j}^{\star}\right)^{\top}{\mathbf x}_{:, j}\right| - (\lambda_1+\lambda_{dag}{\boldsymbol \Upsilon}_{ij}) < 0,
\]
which contradicts the above first-order optimality condition  with ${\boldsymbol \Theta}_{ji}^{\star} \neq 0$. Therefore, when $\lambda_{dag}$ is sufficiently large, the output ${\boldsymbol \Theta}^{\star}$ is guaranteed to be DAG. \\

Summing up the proofs above, the alternate optimization of Eqn.~(2) converges and the output ${\boldsymbol \Theta}^{\star}$ is guaranteed to be DAG when $\lambda_{dag}$ is sufficiently large.  \hfill $\square$\\\\
\end{document}